\pdfoutput=1

\documentclass[11pt]{article}

\usepackage[]{acl}

\usepackage{times}
\usepackage{latexsym}
\usepackage[T1]{fontenc}
\usepackage[utf8]{inputenc}
\usepackage{microtype}
\usepackage{amsmath}
\usepackage{graphicx}
\usepackage{multicol}
\usepackage{adjustbox}
\usepackage{booktabs}
\usepackage{multirow}
\usepackage{float}
\usepackage{cleveref}

\usepackage{xspace}
\newcommand{\ie}{\emph{i.e.}\xspace}
\newcommand{\eg}{\emph{e.g.}\xspace}

\author{Evgenia Ilia \\
  University of Amsterdam \\
  \texttt{e.ilia@uva.nl} \\\And
  Wilker Aziz \\
  University of Amsterdam \\
  \texttt{w.aziz@uva.nl} \\}

\title{Predict the Next Word: \textit{<Humans exhibit uncertainty in this task and language models \_\_\_\_\_>}}

\begin{document}

\maketitle

\begin{abstract}

Language models (LMs) are statistical models trained to assign probability to human-generated text. As such, it is reasonable to question whether they approximate linguistic variability exhibited by humans well. 
This form of statistical assessment is difficult to perform at the passage level, for it requires acceptability judgments (\ie, human evaluation) or a robust automated proxy (which is non-trivial). At the word level, however, given some context, 
samples from an LM can be assessed 
via exact matching against a prerecorded dataset of alternative single-word continuations of the available context.
We exploit 
this fact and evaluate the LM's ability to reproduce variability that humans (in particular, a population of English speakers)
exhibit in the `next word prediction' task. This can be seen as assessing a form of calibration, which, in the context of text classification, \citet{baan-etal-2022-stop} termed \emph{calibration to human uncertainty}. 
We assess GPT2, BLOOM and ChatGPT and find that they exhibit fairly low calibration to human uncertainty. 
We also verify the failure of expected calibration error (ECE) to reflect this, and as such, advise the community against relying on it in this setting. 

\end{abstract}

\section{Introduction}

\begin{figure}[ht]
    \includegraphics[width=7 cm]{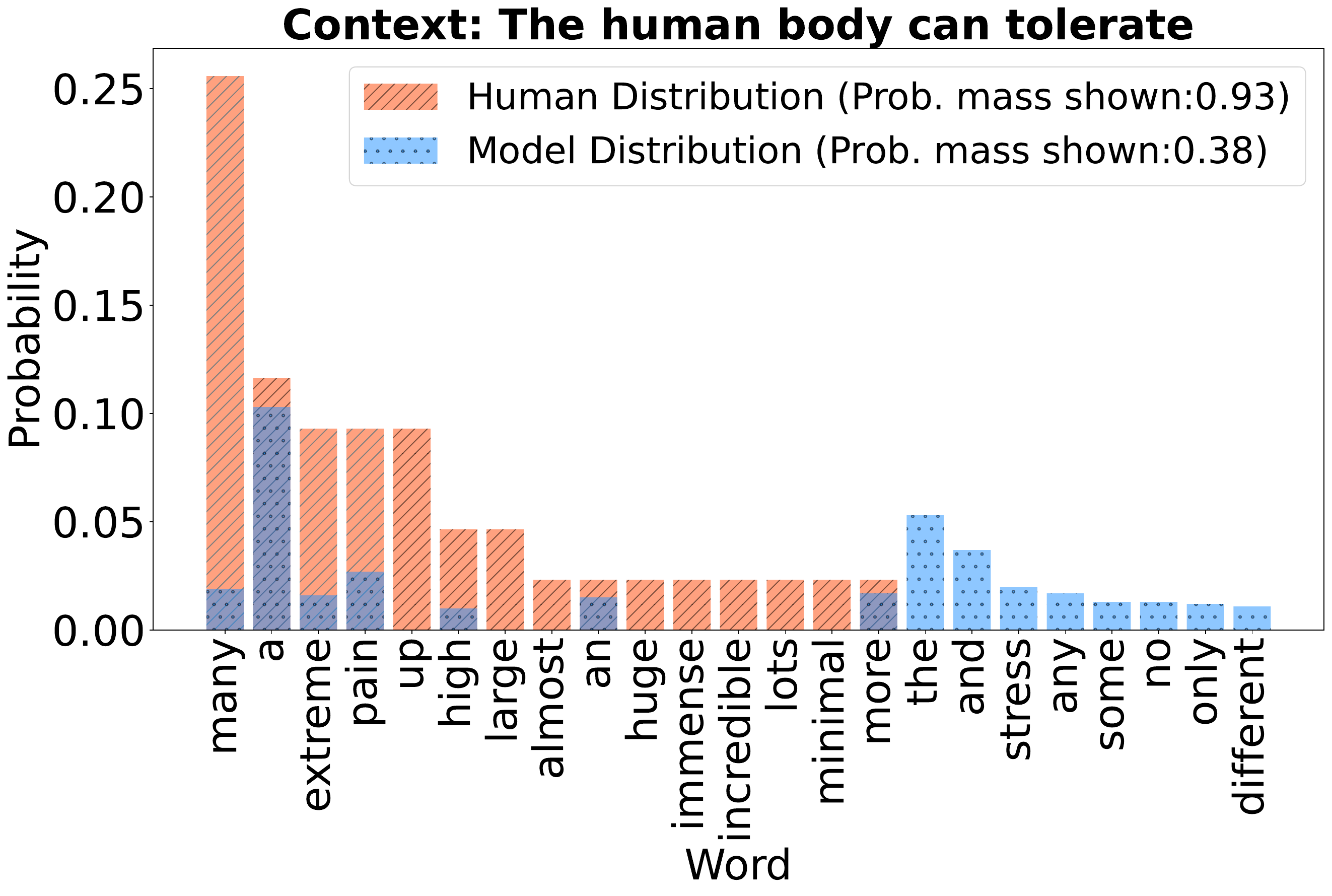}
     \includegraphics[width=7 cm]{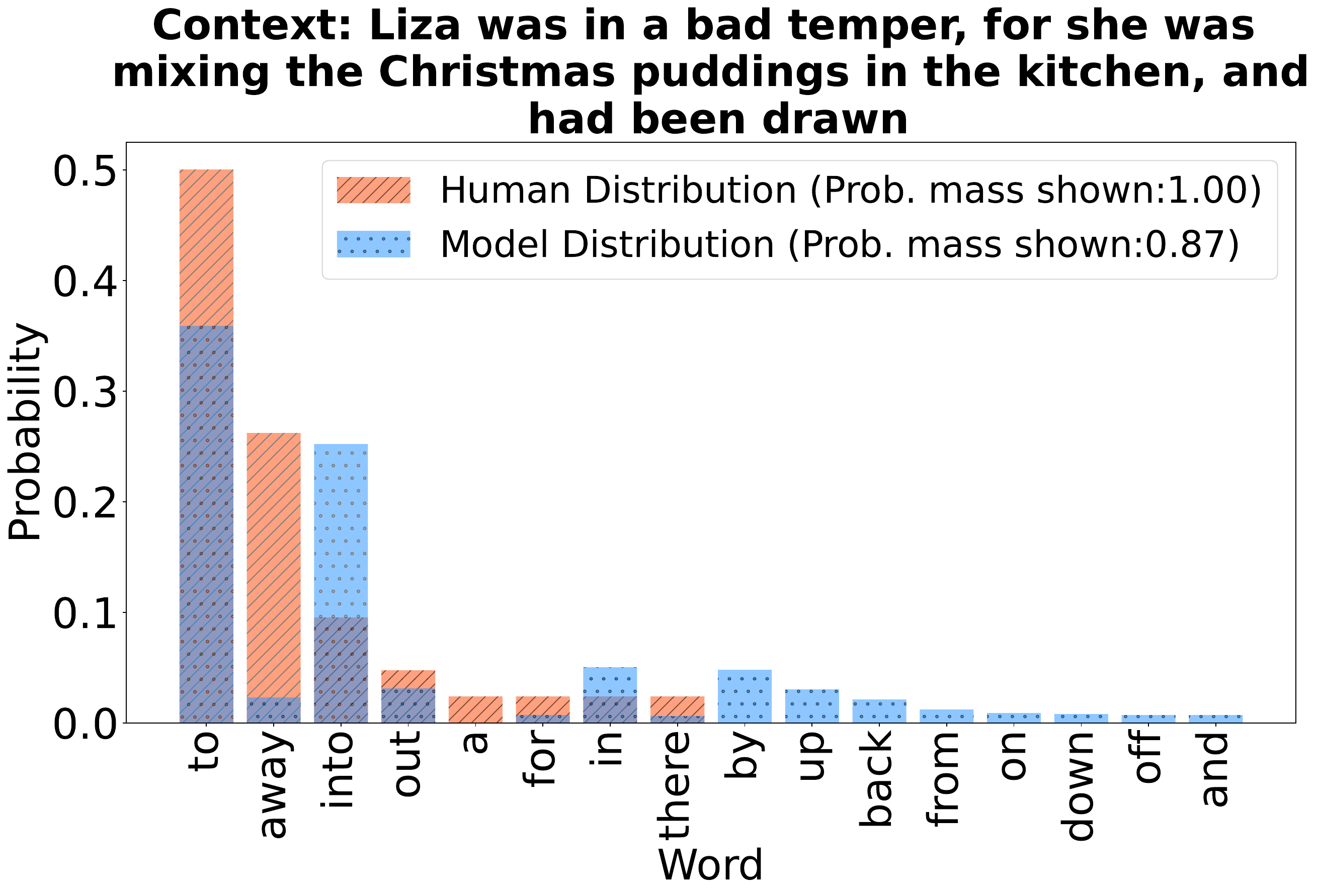}
     \caption{Estimated human and model distributions for contexts (15 most probable words of each distribution).}
     \label{fig:dists}
\end{figure}

Language models (LMs) are trained to assign probability to human-generated text. The typical LM treats a piece of text as a sequence of tokens whose joint probability it factorises autoregressively, with conditional token probabilities predicted from the available context by a neural network  \citep{mikolov2010recurrent,radford2019language, scao2022bloom}. 
An LM can be viewed as a representation of uncertainty about human linguistic production \citep{serrano2009modeling,takahashi-tanaka-ishii-2019-evaluating, meister-cotterell-2021-language,giulianelli2023comes}, specifically, one that reflects the production variability exhibited by the population(s) who generated the training data. 
Despite how plausible this variability is, 
LMs are not consistently exposed to it at the level of individual contexts (\ie, due to data sparsity, most contexts are unique)  
leading us to investigate their ability to predict it well. 

One way to appreciate plausible variability is to ask humans to perform \emph{next word prediction}: show multiple participants the same prefix of a passage and ask each of them to contribute a word that plausibly extends it. 
An LM that assigns probability to any next-word candidate similar to the proportion of the human population contributing it as the next word serves as a good proxy to the production variability of that human population---a  desideratum \citet{baan-etal-2022-stop} termed  \textit{calibration to human uncertainty}.\footnote{Such calibration might be assessed against any population of interest, \eg a specific target audience in a human-machine interaction setting (\eg \citet{williams2008generating}).} %
Studying different notions of calibration of text classifiers, \citet{baan-etal-2022-stop} show that the very popular expected calibration error \citep[ECE;][]{guo2017calibration} is flawed in the presence of data uncertainty (\eg, due to the task's inherent ambiguity \citep{plank-2022-problem}). 
As data uncertainty is hardly avoidable in language modelling, we must entertain the possibility that ECE is not a reliable tool to assess the predictive distributions of an LM, despite its widespread use \citep{kumar2019calibration,wang-etal-2020-inference,tian2023just}.\looseness=-1

To assess calibration to human uncertainty, we compare the uncertainty exhibited by LMs to the uncertainty exhibited by humans in the next word prediction task (Figure \ref{fig:dists})---for which we use Provo Corpus \citep{luke2018provo}, a dataset (in English) with multiple human responses per available context.
We analyse three pretrained LMs of different sizes and training objectives (\ie, GPT2 \citep{radford2019language}, BLOOM \citep{scao2022bloom} and ChatGPT \citep{Introducing_ChatGPT}) and find that they exhibit low calibration to human uncertainty.
We verify ECE's unreliability in this setting and advise the community against relying on it as a meaningful notion of calibration of generative models.

\section{Background}

Given context, an autoregressive LM predicts a conditional probability distribution (cpd) over the model’s vocabulary of known tokens (\textit{i.e.,} subword units). Hence, at this level, an LM can be regarded as a probabilistic multi-class classifier. This  motivates research \citep{muller2019does,kumar2019calibration,wang-etal-2020-inference} assessing the extent to which probabilities predicted by  LMs are interpretable as `rate of correctness', a property referred to as calibration \citep{niculescu2005predicting,naeini2015obtaining,guo2017calibration}.

A multi-class classifier is said to be \emph{confidence-calibrated} if its probabilities predict the classifier's accuracy, specifically, if $(100 \times q)$\% of its predictions made with probability (close to) $q$ are judged to be correct.  
The ECE estimator \citep{guo2017calibration} is the average absolute difference between average confidence and frequency of correctness across confidence bins.\footnote{Correctness is determined by comparing the mode of the predicted cpd to the target label (as pre-recorded in a dataset); the mode's probability is regarded as the classifier's confidence;  closeness to $q$ is determined via a binning scheme.}
\citet{baan-etal-2022-stop} uncovered a logical flaw in measuring ECE under data uncertainty---settings in which human disagreement is a plausible property of the task and hence not to be dismissed as error \citep{aroyo-etal-2019-crowdsourcing,plank-2022-problem}.\footnote{There are many variants of ECE in the literature \citep{kumar-trainable-2018,widmann2019calibration,gupta2021calibration,si-etal-2022-examining,dawkins-nejadgholi-2022-region}. Some variants, in particular, evaluate all probabilities of a cpd (not only the mode probability; \eg, class-wise \citep{vaicenavicius2019evaluating,kull2019beyond}, static and adaptive \citep{Nixon_2019_CVPR_Workshops}), these still assume no aleatoric uncertainty in the data generating process and, hence, remain inadequate tools for our setting. Besides, they are not common in language generation literature.}
They show this in theory and empirically, and propose to assess predicted probabilities against estimates of \emph{target probabilities}. The idea is to exploit multiple judgments per input to obtain the maximum likelihood estimate (MLE) of the target cpd and compare that to the model cpd at the instance level.

\section{Methodology}  \label{sec:methodology}
We compare  the uncertainty that LMs and humans exhibit in next word prediction. 
For that, we must represent their uncertainty over a shared space.

\paragraph{Human distributions.} 
Given some context $c$, we assume that human uncertainty is captured by a single underlying cpd and, hence, regard  human responses to the next word prediction task as i.i.d. draws from it. Then, given multiple responses, the MLE for this cpd assigns probability $p(w|c)$ to word $w$ given  $c$ equal to the relative frequency with which humans predict $w$ to follow $c$.

\paragraph{Model distributions.} 

LMs decompose sentences as sequences of subword units, rather than words. However, humans predict complete words, 
hence, we establish a process for re-expressing the model cpds over the space of complete words.\footnote{Though artificial, one could tokenise the human data and analyse cpds over subword units, we do that in Appendix \ref{appendix:token_exp}.
}
For a given context $c$, we sample unbiasedly complete words from the model and use an empirical estimate of their probabilities; a word $w$ drawn given $c$ is assigned probability $q(w|c)$ equal to its relative frequency in the sample. 
To generate complete words, we (i) sample a token sequence generally long enough to include a word boundary; (ii) merge subword units and slice the first complete word from each generation (using a basic tokeniser); and, finally, (iii) reject samples that failed to generate a full word.\footnote{In Appendix \ref{appendix:alt_model_dist}, we explore an  estimator that uses model probabilities, as it is biased and does not show advantages over MC estimation, we do not adopt it for our main analysis.} 
This procedure  samples potentially different segmentations of the same word(s) approximately marginalising out tokenisation ambiguity---which \citet{cao-rimell-2021-evaluate} show to be an important and unduly neglected aspect of  LM evaluation.

\section{Experiments}

\paragraph{Data.}
Provo Corpus \citep{luke2018provo} contains 55 passages (50 words long on average) in English from various sources \eg news, fiction, science. Each prefix sequence of all passages (2687 prefixes) is given as context to 40 humans, on average, who predict a one-word completion. We use this corpus to estimate target cpds.

\paragraph{Models.} For each context, we estimate cpds for different models. 
First, GPT2 Small \citep{radford2019language}, for which we use 1000 unbiased samples per context.\footnote{To obtain generations for GPT2-Small and Bloom-176B we used the Hugging Face API with arguments: do\_sample = True, num\_beams = 1, top\_k = 0/None (GPT2/Bloom), and temperature = 0.5, where relevant. For ChatGPT (i.e. gpt-3.5-turbo), the OpenAI API was used. Code and generations available from: \url{https://github.com/evgeniael/predict_next_word.git}.} To investigate whether a potential mismatch of training and test domain has an effect on our analysis, we fine-tune GPT2 on a subset of the original passages from Provo;  we call this setting GPT2$_{\text{FT}}$ (the complete experimental setup  is described in Appendix \ref{appendix:fine_tuning}). Additionally, we investigate the effect of temperature scaling (temperature = 0.5),\footnote{This biases the sampling procedure. While this often has a positive effect on ECE, there is no reason to expect a positive effect on calibration to human uncertainty.} and, to reduce computational costs, we opt for 40 generations per context in this analysis (a choice we motivate empirically in Appendix \ref{appendix:larger_models}).
To test the effect of scale on calibration to human uncertainty, we also analyse BLOOM-176B \cite{scao2022bloom}. Again, we opt for sampling 40 generations per context. Due to limited API access, we use a random subset of 669 Provo contexts. 
We are also interested in the effect of reinforcement learning from human feedback \citep[RLHF;][]{christiano2017deep, ibarz2018reward}, hence we  analyse ChatGPT \cite{Introducing_ChatGPT}. As before, we draw 40 samples per context and use a random subset of 500 Provo contexts. 
In one setting we prompt ChatGPT 40 independent times, in another setting (ChatGPT$_{\mathrm{D}}$) we prompt it once to generate a list with 40 options (prompt and additional details in Appendix \ref{appendix:larger_models}).
 For each context, we also have a `control cpd' formed by splitting the human annotation in two disjoint parts from which we estimate two cpds, one regarded as target, one regarded as an oracle model; this allows us to form an expectation about realistic levels of calibration.

\paragraph{Metrics.}
For each context, we compare a pair of cpds (a model vs the target for that context) 
in terms of their total variation distance (TVD).\footnote{$\mathrm{TVD}_c(p, q) = \frac{1}{2}\sum_{w} |p(w|c)-q(w|c)|$, where the sum is over the union of model- and human-generated words.} 
To study a whole dataset, we plot TVD's 
 distribution across contexts; for a numerical summary, following \citet{baan-etal-2022-stop}, we report \emph{expected TVD} (average TVD for all contexts) as a measure of calibration to human uncertainty.
Finally, we compute ECE by comparing the mode of each model cpd to the original corpus word and ECE variants that use as targets the human or oracle majority per context.

\section{Results}
\label{sec:measure_calibration}

\begin{table*}[t]
    \centering
    \footnotesize 
    \begin{tabular}{lllllllll}
        \toprule
        \multirow{1}{*}{Gold Label} & \multicolumn{7}{c}{ECE $\downarrow$} \\
        {} & {Human} & {Oracle$_2$} & {GPT2} & {GPT2$_{\text{F}}$} & {GPT2$_{\text{T}}$} & {Bloom} & {ChatGPT} & {ChatGPT$_{\text{D}}$}
        \\
        \midrule
        Original & 0.14 & 0.11  & 0.02 & 0.03 & 0.35 & 0.07 & 0.45 & 0.10\\
        Human Maj. & 0.60 & 0.57 & 0.20 & 0.22 & 0.13 & 0.09 & 0.37 &  0.08\\
        Oracle$_1$ Maj. & 0.19 & 0.32  & 0.19  & 0.19 & 0.15 & 0.07 & 0.37 & 0.08\\
        \midrule \midrule
        Avg TVD $\downarrow$& - & 0.42 & 0.64 & 0.66 & 0.61 & 0.61 & 0.76 & 0.82\\ \bottomrule
    \end{tabular}
\caption{ECE (the row indicates the target, the column indicates the system) and Expected TVD results. We resample the disjoint oracles  20 times and report the mean ECE (standard deviations < $0.1$).} 
\label{table:ece}
\end{table*}

\begin{figure}[t]
    \begin{center}
        \includegraphics[width=7.5 cm]{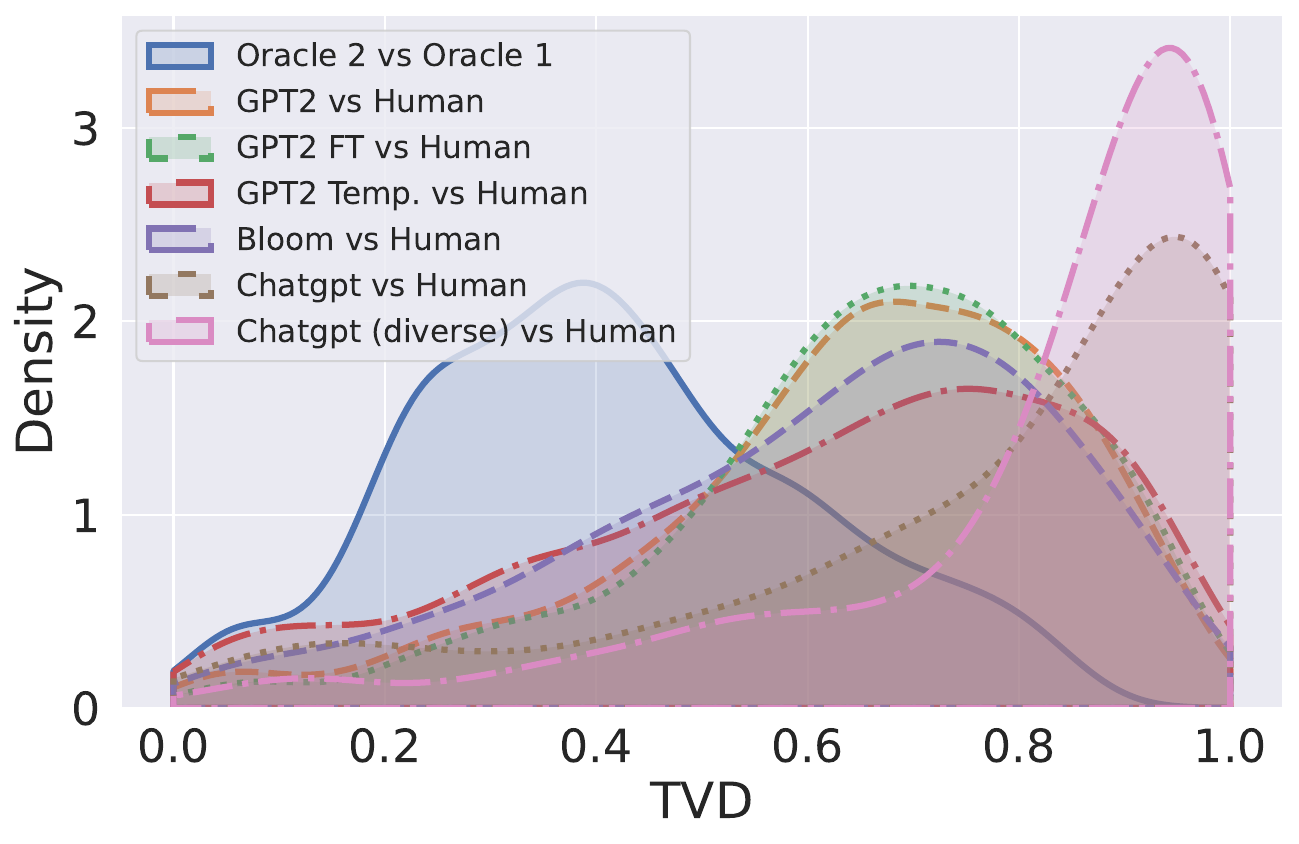}
        \caption{KDE plot of TVD values between a model and the estimated human target cpd, and between oracles.}  
        \label{fig:tvd_res}
    \end{center}

\end{figure}

\paragraph{Main findings.}Table \ref{table:ece} presents ECE and Expected TVD results. As predicted, ECE ranks most models as better calibrated than human oracles,  
confirming that  it cannot be trusted in this setting. Figure \ref{fig:tvd_res} illustrates kernel density estimate (KDE) plots of instance-level TVD values between our models' cpds and the target (human) cpds, along with the KDE plot of TVD values between two disjoint oracles. We observe how the distributions of all models are skewed towards higher TVD values, with ChatGPT performing the worst. 
The inability of models to reproduce variability cannot be attributed to population mismatch alone, as GPT2$_\text{FT}$ displays similar trends to GPT2, and it persists in larger models, while RLHF worsens the issue (for both sampling strategies). Lastly, we observe how temperature scaling does not meaningfully address the issue (regardless of its effect on ECE).  

\paragraph{What do TVD differences mean?}
We measure a difference of around 0.2 TVD units between GPT2's and oracles' means, but, we lack understanding of the practical significance of this difference. 
To gain some insight, we conduct a controlled experiment. We artificially improve $k$\% of the model's cpds by replacing them by an oracle estimate. We then measure TVD between this artificial improvement and a disjoint oracle allowing us to associate units of TVD with an interpretable rate of improvement (\ie, percentage of plausible cpds). We find that we need to replace about 60\% of GPT2's cpds to achieve TVDs that distribute similarly to human performance.\footnote{In Appendix \ref{appendix:improving_model}, we verify that our findings a robust to choices of $k$, random seed and sample size.} 

\paragraph{Why can't models reproduce human variability?}
For further insight, we analyse GPT2's inability to reliably reproduce human variability. In Figure \ref{fig:dists}, we visualise target cpds and GPT2's (for the top-15 highest probability words) for two contexts; Appendix \ref{appendix:visual_dists} lists a full passage. We choose the distributions of Figure \ref{fig:dists} to demonstrate some observations; (1) GPT2's cpd fails to align with the human one in samples where the outcome is barely constrained (true for the majority of the many instances we examined), and (2) when the outcome is fairly constrained, such as when completing a prepositional verb, GPT2 performs much better. 

\begin{figure}
    \begin{center}
        \includegraphics[width=7 cm]{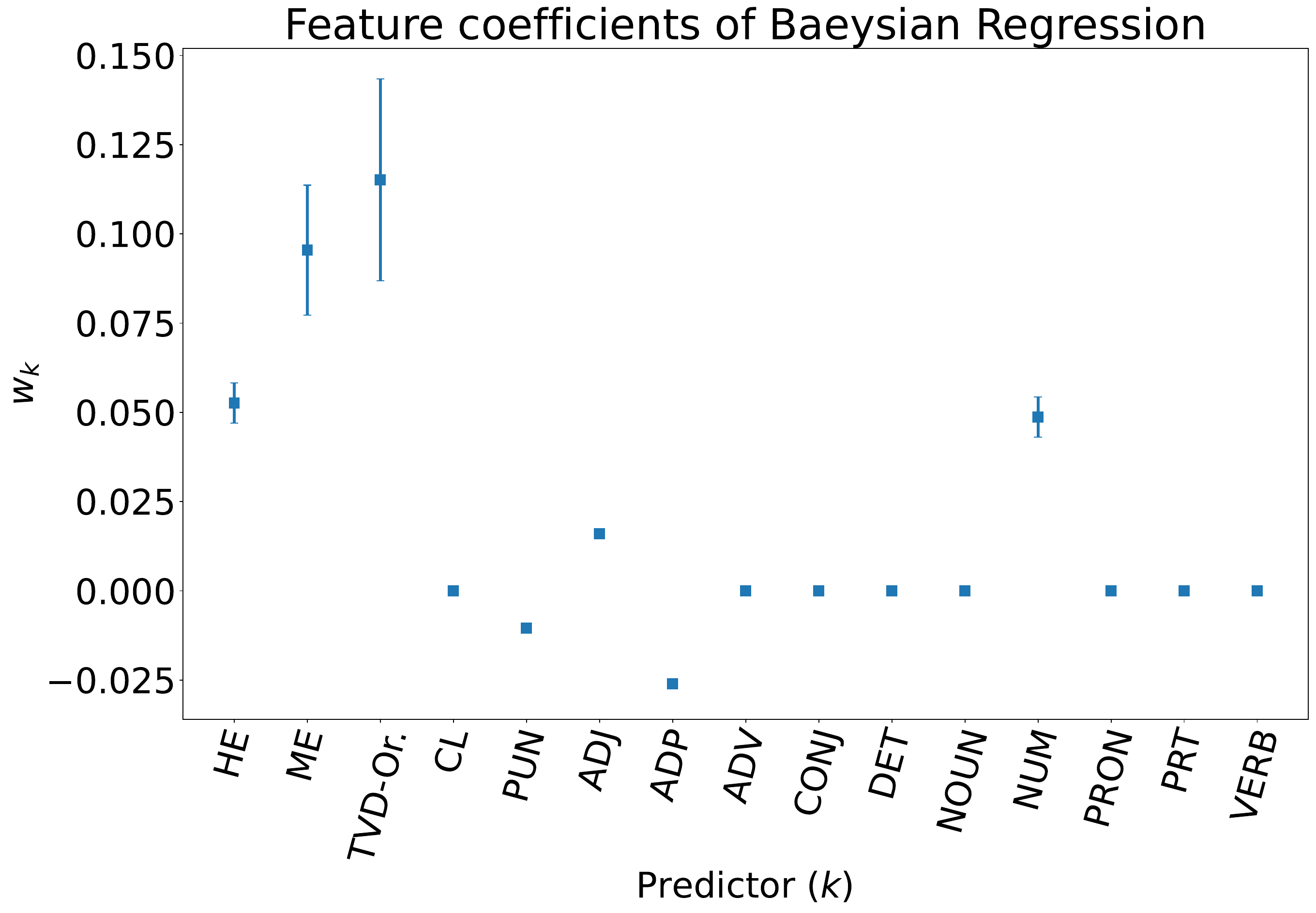}
        \caption{Regression coefficients and their credible intervals. Features, in order: Human entropy, Model Entropy, TVD between oracles, Context Length, Punctuation, and 10 universal POS tags.}
        \label{fig:regr_coef}
    \end{center}
\end{figure}

We attempt to quanitfy the effect of our observations. We perform Bayesian regression with automatic relevance determination \citep[ARD;][]{neal2012bayesian} using, for each context, TVD between GPT2 and the oracle cpd as the regression target, and predictors that are indicative of how constraining  a context is (TVD between oracles, entropy of target cpd), as well as context length and the entropy of the model cpd; with the former two being high precisely for contexts that admit more  plausible variability. 
We also add as predictor the POS-tag of the context's last word, according to a POS-tagger. Figure \ref{fig:regr_coef} presents the feature's coefficients and credible intervals. ARD ranked TVD between oracles as most important, confirming that GPT2 struggles precisely in those cases of higher plausible variability (discussion in Appendix \ref{appendix:predictors}).   

\paragraph{Beyond exact word matching.}
From our analysis, it is evident that models do not manage to reproduce human variability well at the surface word level. We investigate whether they manage to reproduce it on a more abstract level. We consider a (shallow) syntactic level, where models might produce words with parts-of-speech similar to humans; and a semantic level, where models might produce words that have similar meanings as humans. To measure this, we introduce syntactic TVD (TVD$_{syn}$) and semantic TVD (TVD$_{sem}$). 

We employ a POS-tagger on the concatenation of each context and human generation, so that we obtain the POS-tags of the human samples. Similarly, we obtain the POS-tags of the model generations. As in Section \ref{sec:methodology}, we obtain the human, model and oracle POS-tag distributions via their MLE estimates, so as to compute TVD$_{syn}$. 

\begin{figure}
    \begin{center}
        \includegraphics[width=7 cm]{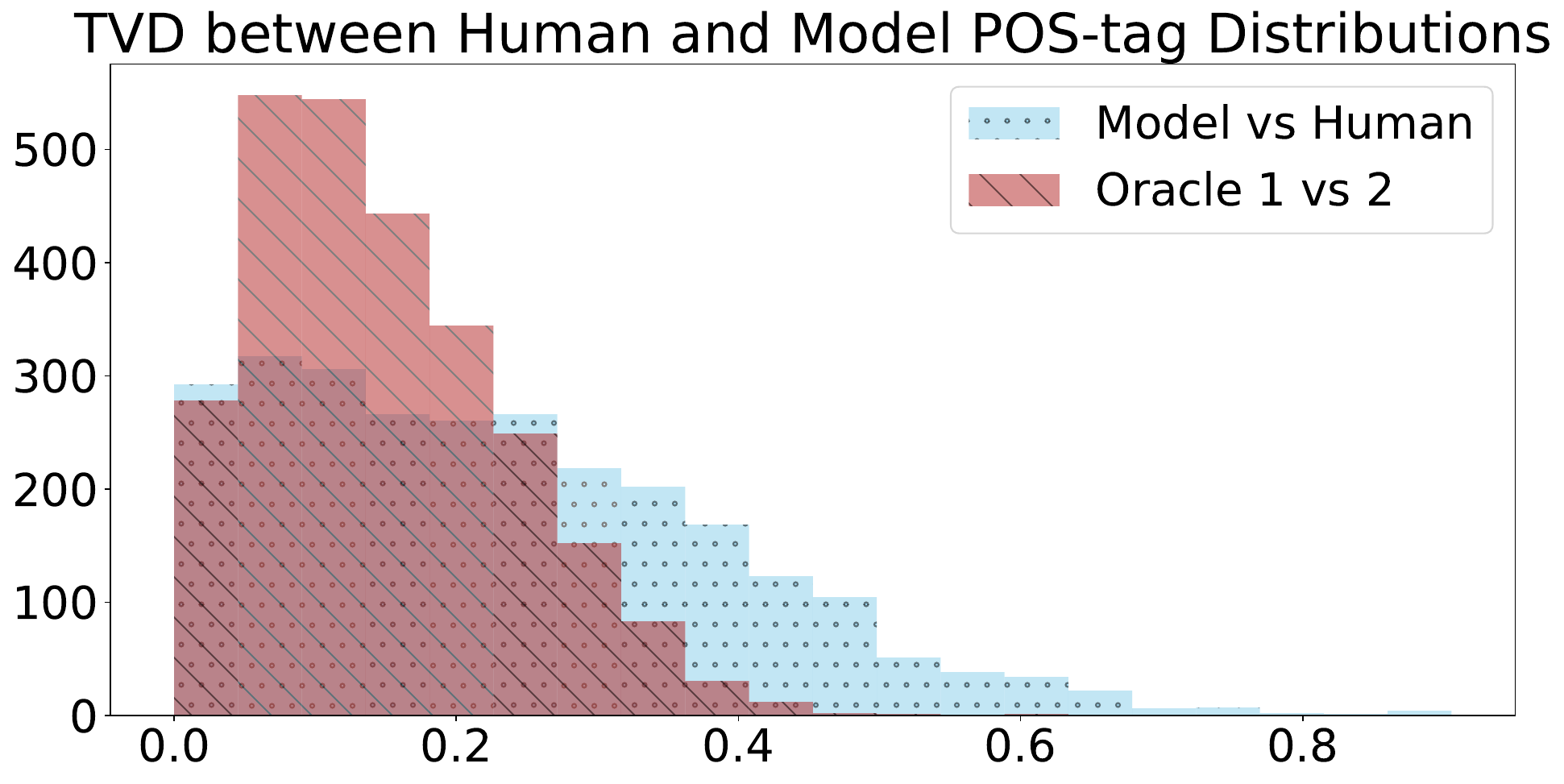}
        \includegraphics[width=7 cm]{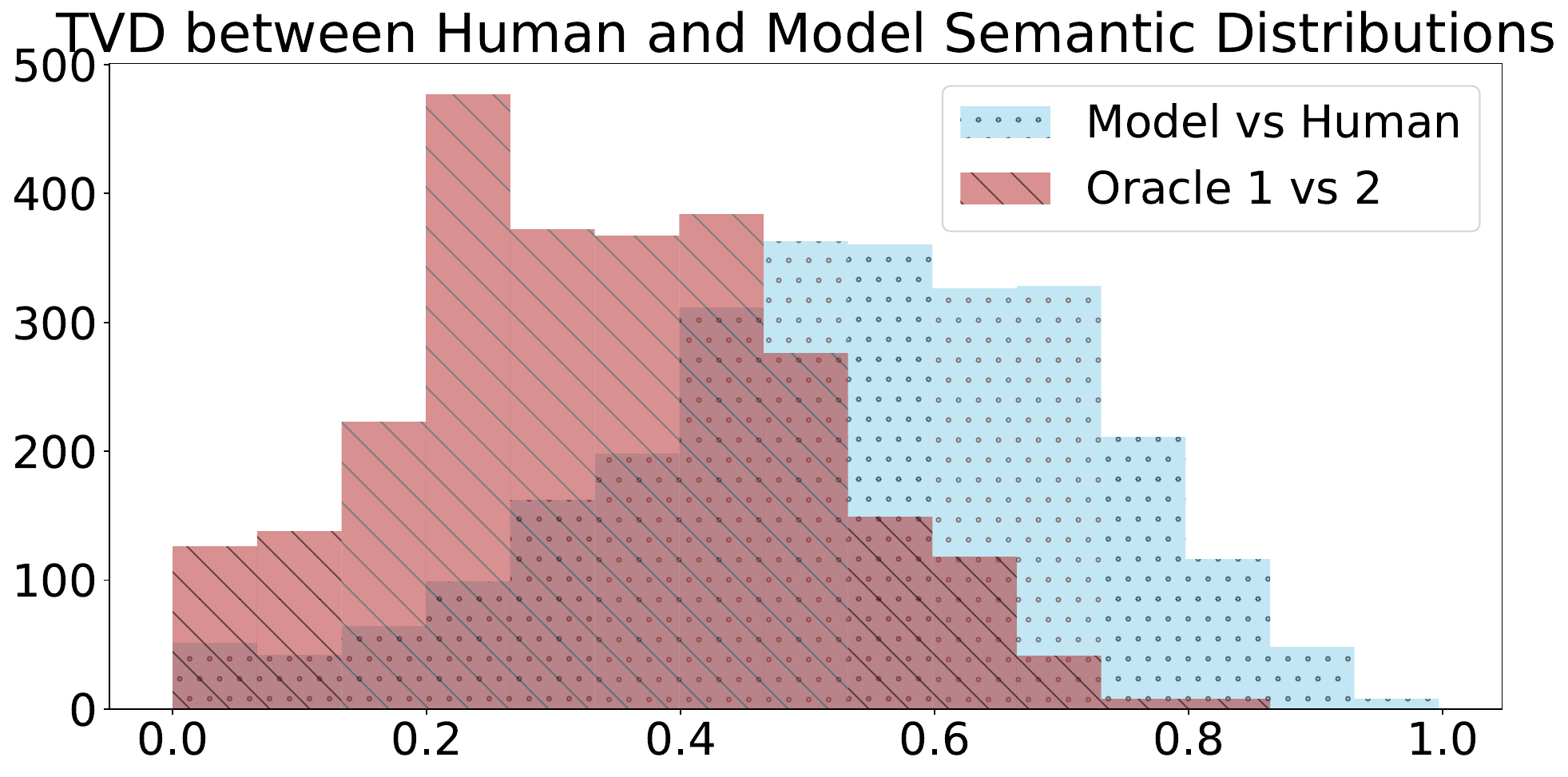}
        \caption{Histogram of TVD$_{syn}$ and TVD$_{sem}$ for all contexts}
        \label{fig:tvd_syntactic}
    \end{center}
\end{figure}

For the semantic analysis, we use clustering to identify words with similar meaning and repartition the support of the distributions. For each context, we create the joint set of human and model generations and cluster their word2vec embeddings using k-means. Words that do not have a word2vec embedding form a group on their own. 
Then, under each model, the probability of a word cluster is the sum of probabilities of the words in it. TVD$_{sem}$ is computed between two such distributions, for humans and the model and between oracles. Appendix \ref{appendix:semantic_anal} contains further details on the experimental setup. As POS tagging and word clustering are not free of errors, TVD$_{syn}$ and TVD$_{sem}$ may be under- or over-estimated in some cases.
Figure \ref{fig:tvd_syntactic} shows histograms for all contexts. We observe similar trends as in previous experiments.

\section{Related Work}

There has been work that exploits predictive distributions of LMs in various ways. \citet{lebrun2022evaluating} analyses such distributions and finds that they overestimate the probability of ill-formed sequences. 
Others investigate alternative training signals that minimise the distance between the data and model distributions \citep{ji2023tailoring, labeau-cohen-2019-experimenting, zhang2023mixce}. 
Our work exploits predictive distributions as an uncertainty representation of human linguistic production and study their calibration. 
Several works study how well-calibrated LMs are and how to alleviate miscalibration \citep{he2023preserving, lee2022adaptive, xiao2022uncertainty, ahuja2022calibration, chen2022close, kumar2019calibration, li2022calibration, xiao2021hallucination} --- the majority using ECE to substantiate their findings, whose inadequacy makes us believe that a new round of studies is needed to assess this matter; our work being an example.

There is a line of work that stresses the value of obtaining multiple human labels per input \citep{plank-2022-problem, basile2020s, grossmann2022beyond, prabhakaran2021releasing}, embracing data uncertainty in classification; \citet{baan-etal-2022-stop} propose calibration metrics that accommodate label variability in natural language inference \citep[NLI;][]{bowman-etal-2015-large}. In concurrent work, \citet{lee2023can} measure the calibration of LM-based classifiers to human uncertainty on ChaosNLI \citep{nie-etal-2020-learn}, also using \citeauthor{baan-etal-2022-stop}'s expected TVD. 

Other work further investigates uncertainty in an NLG setting. \citet{zhou2023navigating} and \citet{kadavath2022language} prompt LMs to output uncertainty linguistically. \citet{kuhn2023clam} prompt LMs to ask for clarifying questions when faced with ambiguous inputs. Similarly, \citet{cole2023selectively} sample repeatedly from LMs to assess whether they are able to answer ambiguous questions. \citet{giulianelli2023comes} analyse various NLG tasks, their variability, and the ability of LMs to capture it. Additionally, \citet{ kuhn2023semantic} introduce semantic entropy, which incorporates linguistic invariances such as meaning equivalence, while  \citet{santurkar2023whose} prompt LMs to assess whether they represent the political views of US Americans from different demographics. 
Finally, \citet{eisape2020cloze} analyse the miscalibration of LMs from a psycho-linguistic lens, and fine-tune an LSTM model using multiple labels. Our work is an addition to this line of work.

\section{Conclusion}
Our work joins a stream of work acknowledging and better incorporating data uncertainty into evaluation protocols \citep{baan-etal-2022-stop, giulianelli2023comes}. In particular, we find empirical evidence for ECE's unreliability in this setting and advise further research into calibration of LMs not to use it.
With a more appropriate tool, we analyse three modern pretrained LMs and find that they are not well calibrated to human uncertainty, unlike ECE might suggest. 
We believe that this inability stems from models not being consistently subjected to human production variability during training, and plan to investigate this further in future work. 

\section*{Limitations}
The assessment of calibration to human uncertainty we have conducted is only one aspect of a system's quality and is not meant to de-emphasise  the importance of any other sound form of evaluation, but rather to offer a complementary tool that supports an insightful set of observations about modern LMs. 
The computational costs of generating a large amount of continuations can be restrictive; as well as the cost of multiple annotations for each context. However, we believe that the benefits of obtaining such data and measuring uncertainty with more reliable methods, outweigh these costs. To foster research, we share the generations that supported this research. The high cost of obtaining data with multiple references per prompt results in another limitation: the limited availability of such labelled data.
The limited number of human annotations per context is another limitation which is hard to alleviate. 
We considered all human annotations to be draws from the same underlying distribution, which is an assumption we cannot verify easily (\eg we do not know if all participants had similar perspectives and backgrounds).
Lastly, we only studied models trained for English. For less resourced languages, data-scarcity is expected to have worse effects on LMs' calibration. Simultaneously, English has a relatively fixed word order and simple morphology. Other languages might exhibit even greater variability due to their own typological features. In turn, we might be required to annotate larger datasets or study the phenomenon at a different level of granularity.

\section*{Acknowledgements}
Evgenia Ilia and Wilker Aziz are supported by the EU’s Horizon Europe research and innovation programme (grant agreement No. 101070631, UTTER).

\bibliography{anthology,custom}

\appendix

\section*{Appendix}

\section{Method 2 - Biased Model Estimate}
\label{appendix:alt_model_dist}

We attempted constructing another estimator of the model distribution. Unlike the MC estimator in the main text, this estimator is  biased due to it overestimating the probability of words in the distribution support and underestimating ones not belonging to it. This estimator forces the model to assign non-zero probabilities to humans responses; in an attempt to see if the model will, in this case, be able to predict human variability better.

We construct the support of the distribution as words that are `likely' under the model. These include words generated with unbiased and nucleus sampling, the greedy word, as well as the original corpus word and human-answered words. For the words requiring sampling from the model, we follow a procedure similar to the unbiased estimator for ensuring sampled words are complete.

The probability for each word is computed by renormalising the joint probabilities the model assigns for the corresponding token sequences:
\begin{align}
   \log q(w|c) &= \log f(c,w) - \log f(c) \nonumber \\
   &- \mathrm{logsumexp}_k [\log f(c,k) - \log f(c)] ~,
\end{align}
where $f(.)$ is the joint probability of the tokenised sequence, as assigned by the neural model.

\begin{figure}
    \begin{center}
        \includegraphics[width=7.5 cm]{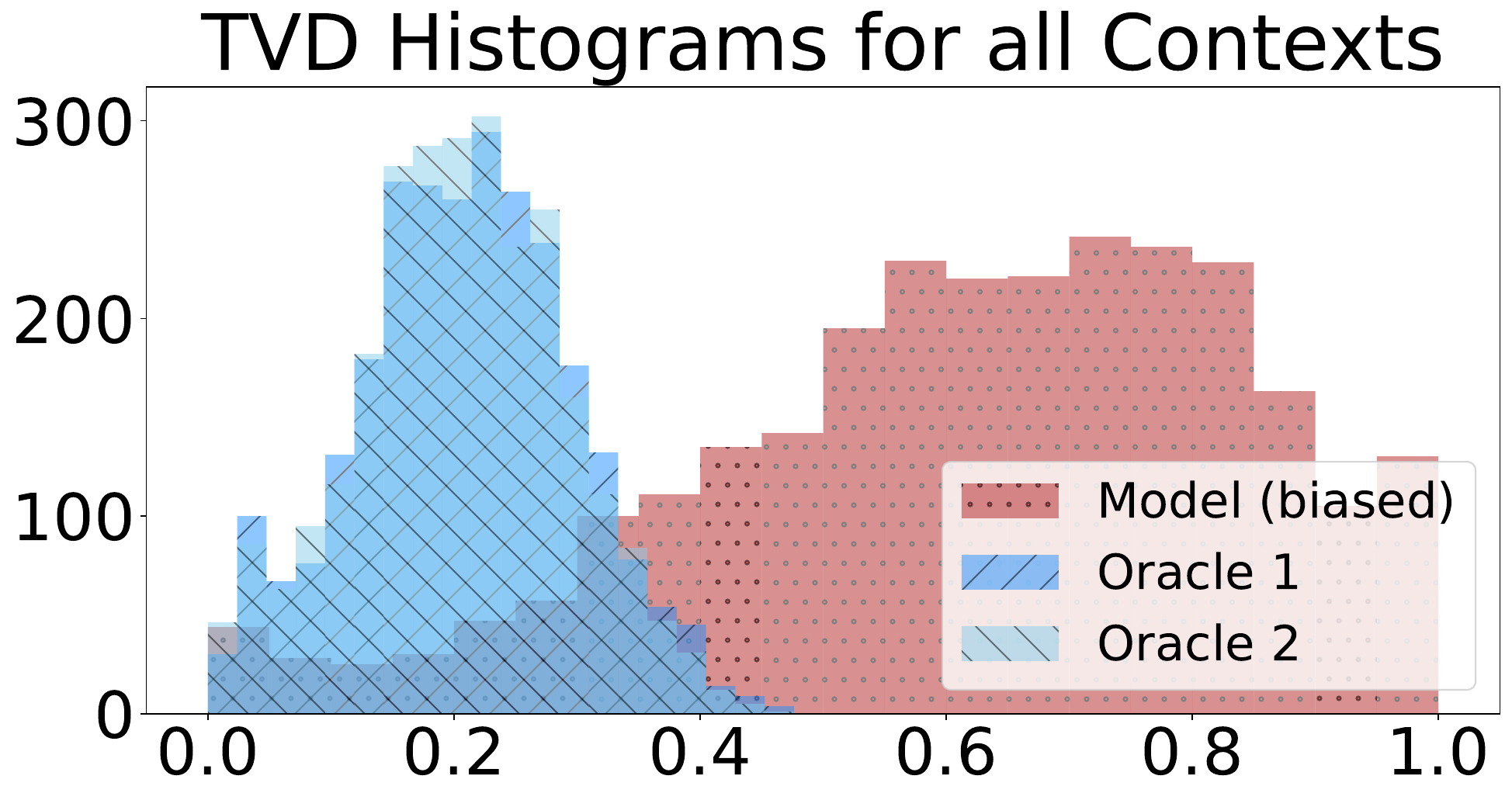}
        \caption{Histogram of TVD values for (biased) model and oracle distributions when compared to the full human distribution}
        \label{fig:tvd_biased}
    \end{center}

\end{figure}

We also evaluated the model's performance using such distributions. We use the same 1000 unbiased samples as before and an additional 100 nucleus samples for each of $p \in {0.7, 0.8, 0.9}$. Results for ECE and TVD are shown in Table \ref{table:ece_biased} and Figure \ref{fig:tvd_biased} respectively. We observe similar results with the unbiased model in terms of both ECE and TVD.

\begin{table}[ht]
    \centering
    \begin{tabular}{lllll}
        \toprule
        \multirow{2}{*}{Gold Label} & \multicolumn{3}{c}{ECE} \\
        {} & Model & Oracle 1 & Oracle 2 \\
        \midrule
        Corpus Word & 0.068 & 0.116  & 0.185 \\
        Human Majority & 0.138 & 0.563 & 0.458 \\
        \bottomrule
    \end{tabular}
\caption{ECE results for the (Biased) Model and Oracle Distributions when considering the Gold-Label to be the corpus word or the human majority}
\label{table:ece_biased}
\end{table}

\section{Predictors of TVD between model and oracle}
\label{appendix:predictors}

\begin{figure}
    \begin{center}
        \includegraphics[width=6.7 cm]{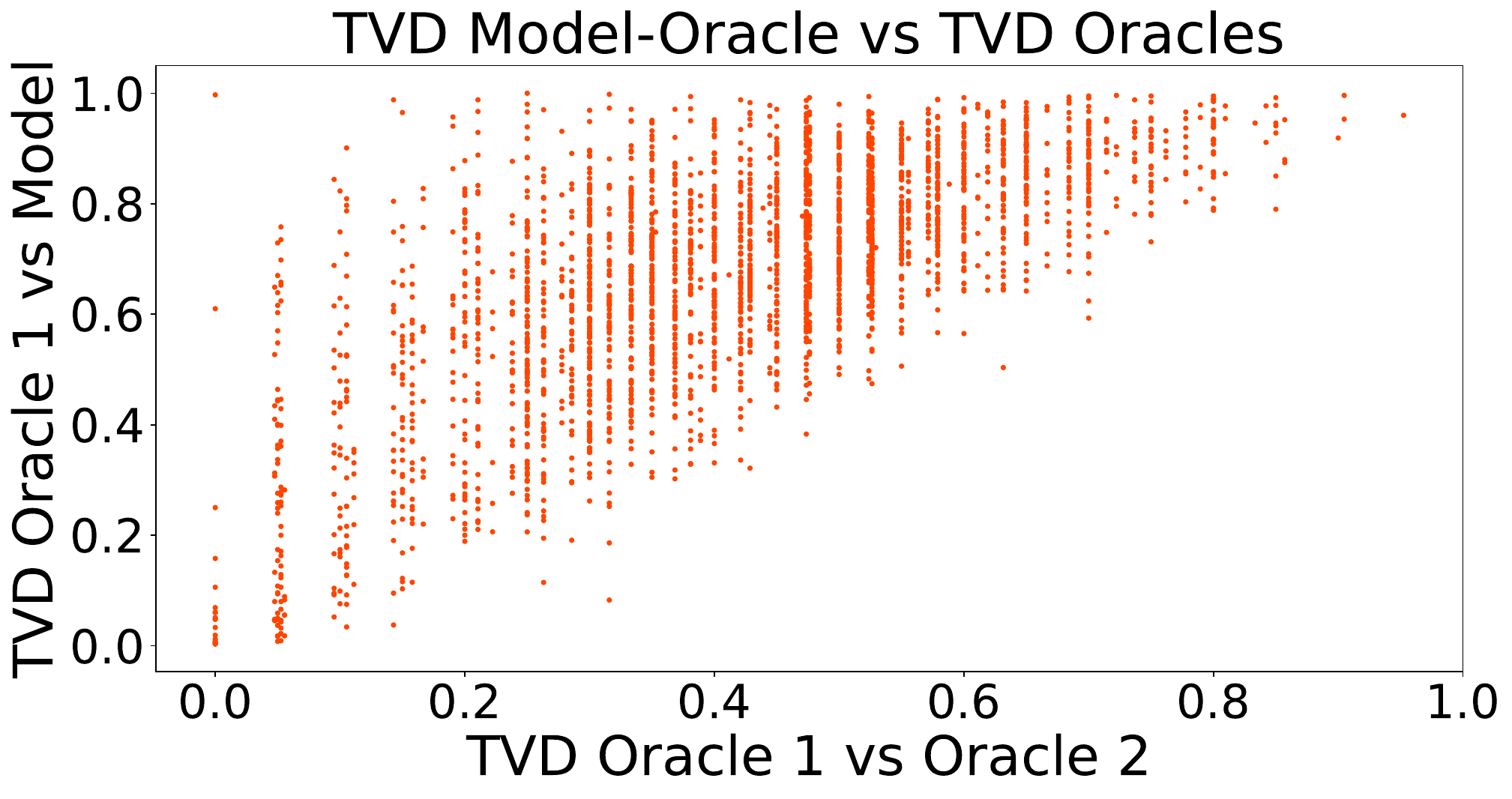}
        \caption{TVD values between oracles and TVD values between model and an oracle}
        \label{fig:predictor_tvd_oracles}
        \includegraphics[width=6.7 cm]{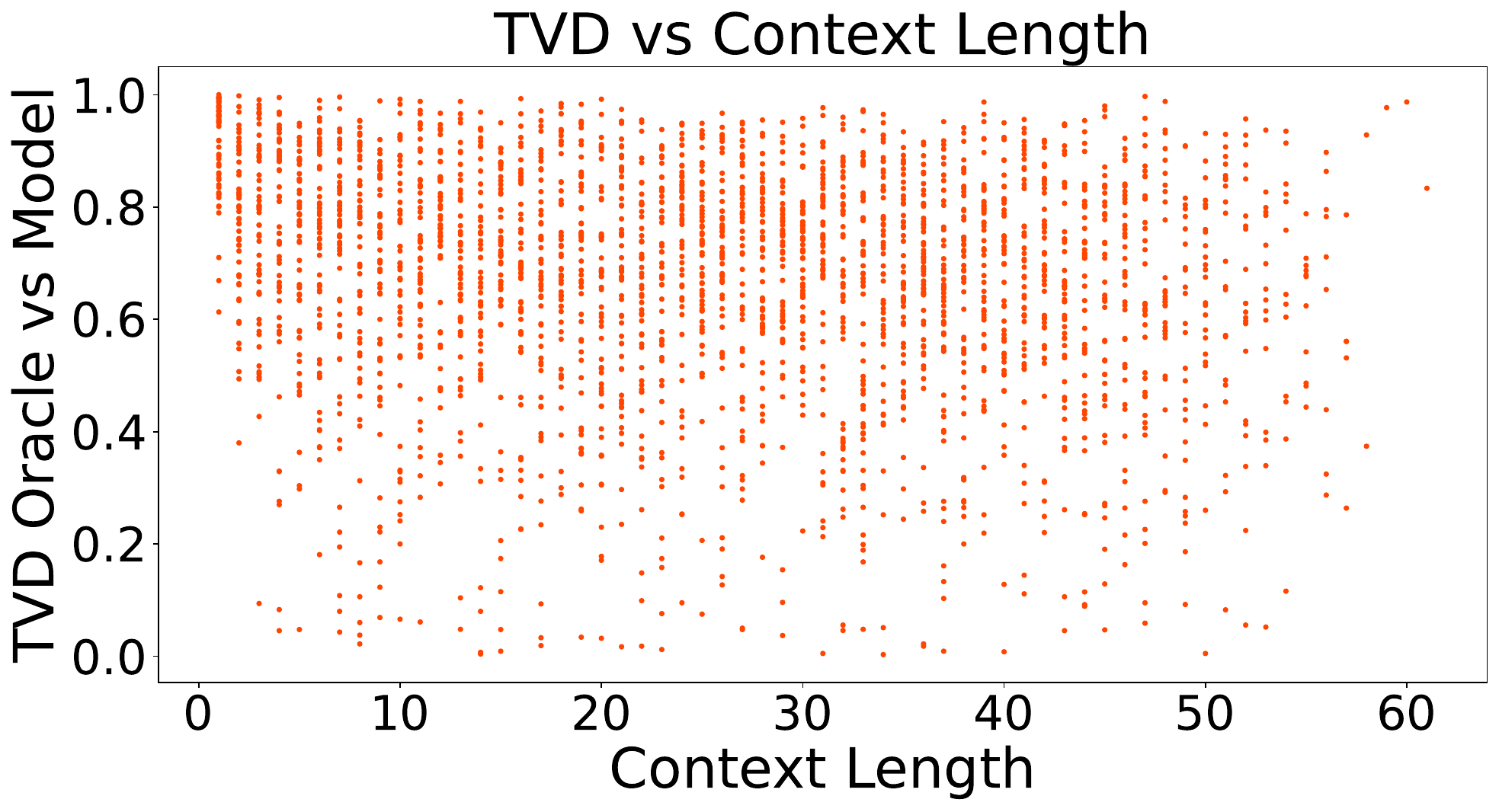}
        \caption{TVD values (between model and oracle) against Context Length}
        \label{fig:predictor_context_length}
        \includegraphics[width=6.7 cm]{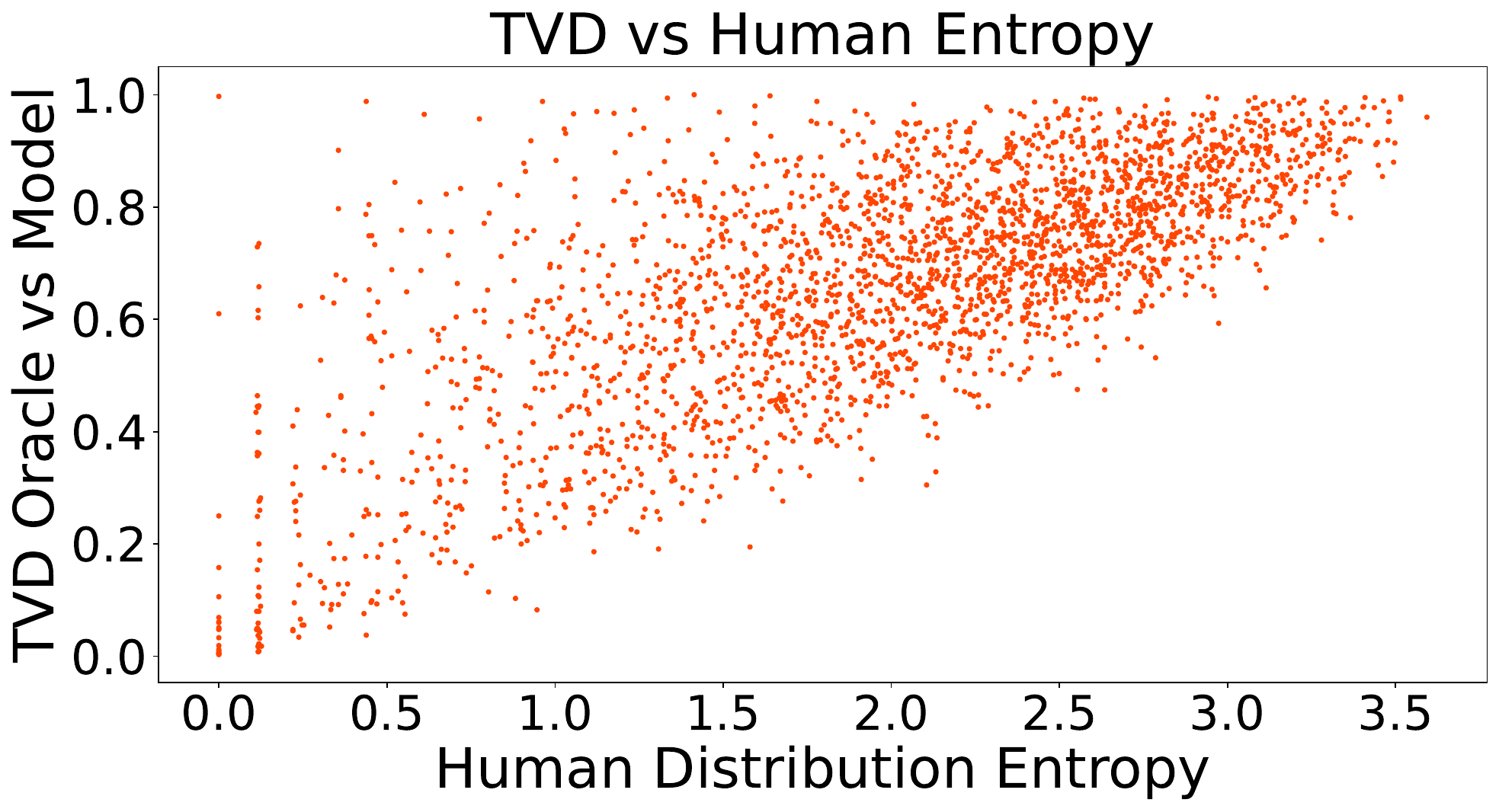}
        \caption{TVD values (between model and oracle) against Human Entropy}
        \label{fig:predictor_human_entropy}
        \includegraphics[width=6.7 cm]{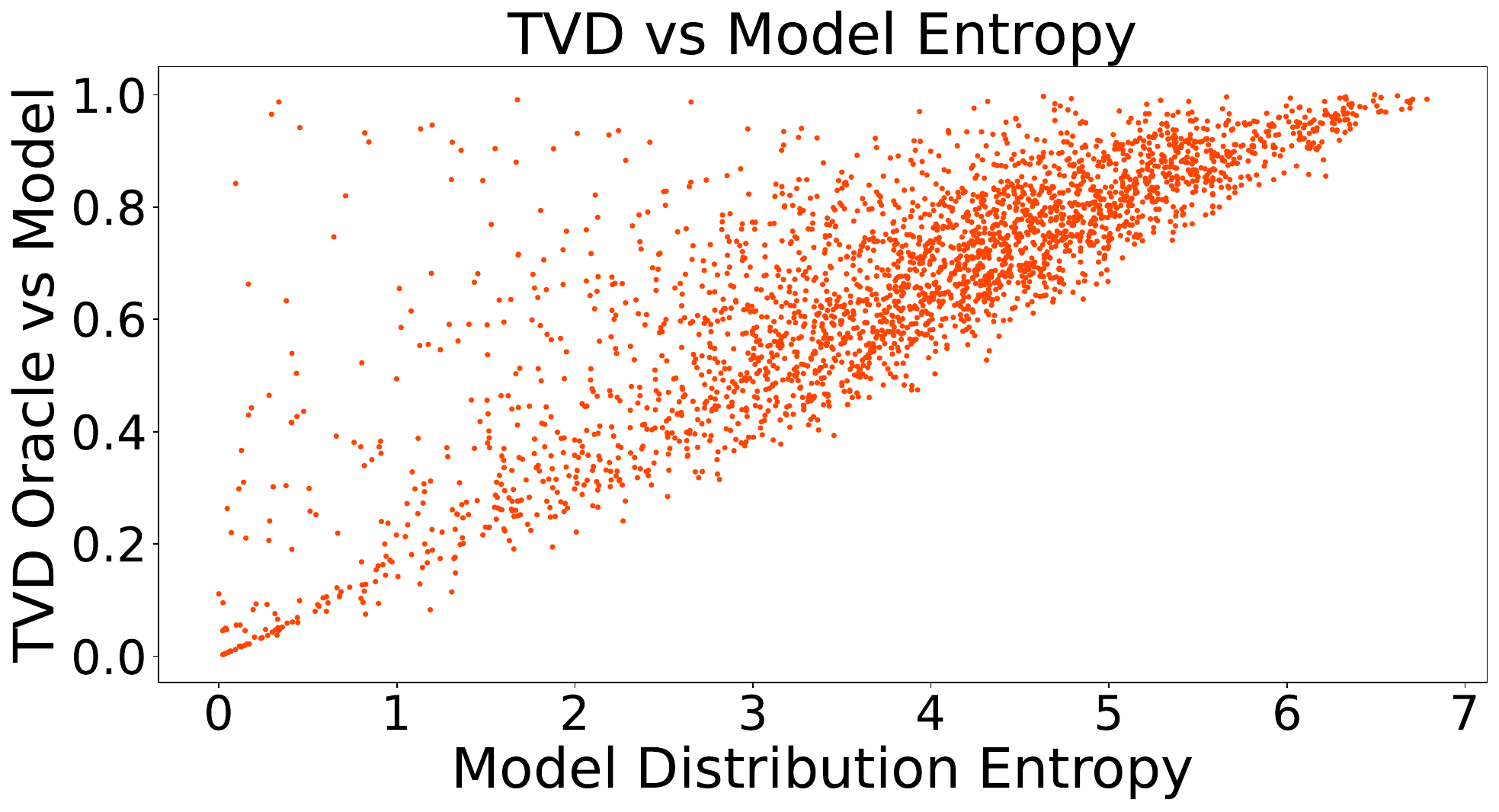}
        \caption{TVD values (between model and oracle) against Model Entropy}
        \label{fig:predictor_model_entropy}
        \includegraphics[width=6.7 cm]{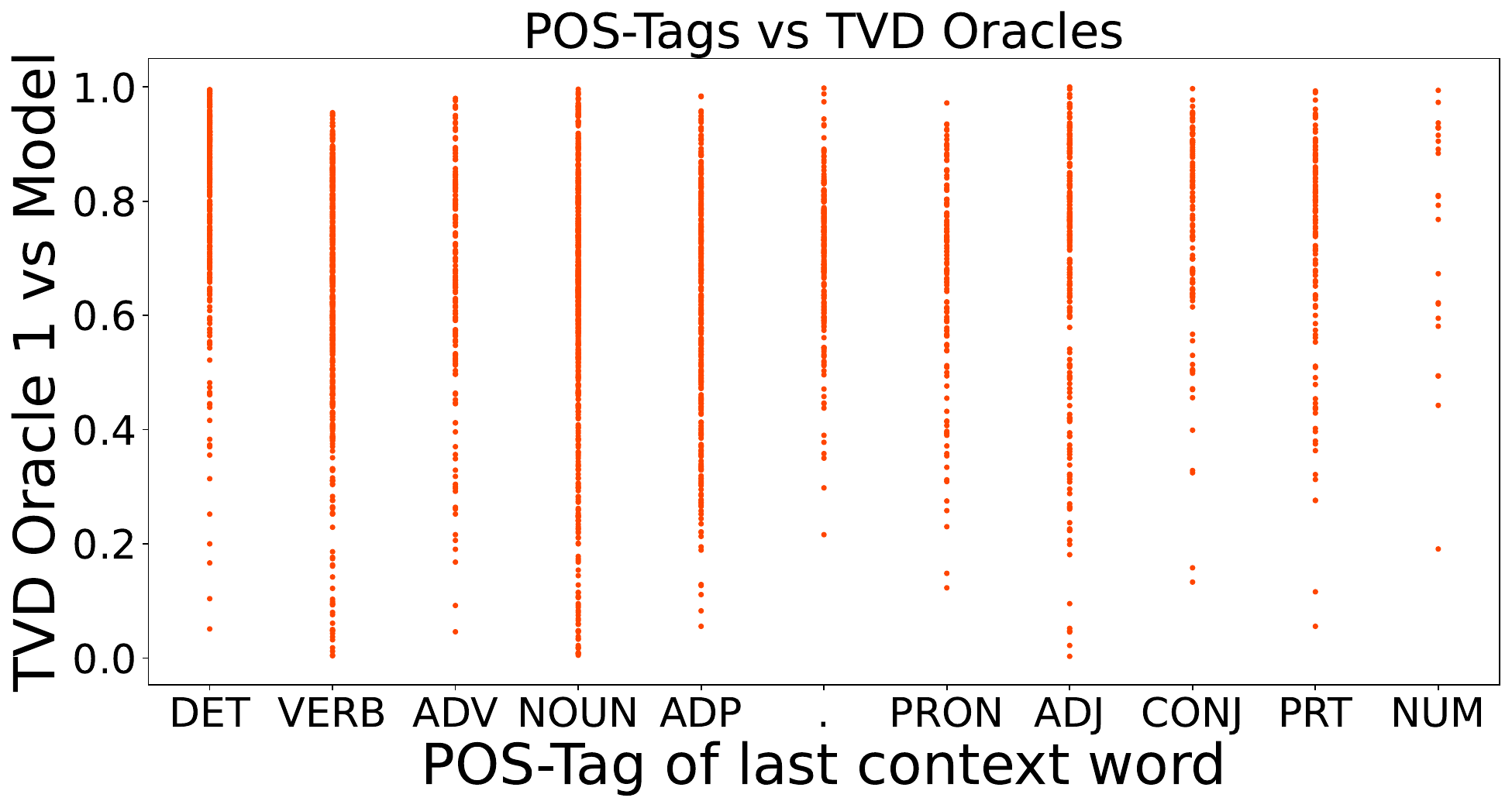}
        \caption{TVD values (between model and oracle) against Pos-tags of last context word}
        \label{fig:predictor_pos}
    \end{center}

\end{figure}

We plot the target variable, TVD between the human and the model cpds against different predictors of interest (Figure \ref{fig:predictor_tvd_oracles} - \ref{fig:predictor_pos}). 
One particular predictor, the TVD between Oracles (Figure \ref{fig:predictor_tvd_oracles}) is of interest, since it provides support for the claim made in Section \ref{sec:measure_calibration}; regarding GPT2's ability to predict variability well when the next word prediction task is less constrained.
The results seem to support this theory - in the very low disagreement range between humans (TVD < 0.15), the model seems to predict variability well - or better, the lack of it. We also investigate context length as a predictor of the model's ability to predict human variability (Figure \ref{fig:predictor_context_length}) - but surprisingly, we observe how the two seem to not be correlated. The plot with the human entropy and model entropy as the predictors, show a positive correlation (Figure \ref{fig:predictor_human_entropy} and \ref{fig:predictor_model_entropy} respectively). This seems to be reinforced by the ARD results. 
Regarding the POS-tag predictors, when the last context word is an adjective, this seems to be an indicator of models being worse at reproducing human variability. Since nouns commonly follow adjectives - this might imply that when models predict nouns, their predictions do not align well with human ones. This might stem from the fact that nouns are content words, and that might inherently allow for higher variability. For a similar reason, the numerical POS-tag (which again is commonly followed by nouns), appears to be a predictor of worse model performance. We observe how adpositions have a negative coefficient, meaning that when models predict words that follow prepositions or postpositions, their predictions align better with human ones. This might be related to the observation discussed in Section \ref{sec:measure_calibration} (when the outcome
is fairly constrained GPT2 performs much better). Punctuation also seems to exhibit a similar trend.
The results from the Bayesian regression with automatic feature determination are in Table \ref{table:bayesian_regr}, where each predictor and its coefficient are shown.

\begin{table}[t]
    \centering
    \begin{tabular}{lll}
        \toprule
        Predictor & Coefficient \\
        \midrule
        Human Entropy &  0.053 \\
        Model Entropy &  0.095 \\
        TVD between Oracles & 0.117 \\
        Context Length & 0 \\
        Punctuation & -0.010 \\
        Adjective & 0.016 \\
        Adposition & -0.026 \\
        Adverb & 0 \\
        Conjunction & 0 \\
        Determiner & 0 \\
        Noun & 0 \\
        Numerical & 0.049 \\
        Pronoun & 0 \\
        Particle & 0 \\
        Verb & 0 \\
        \bottomrule
    \end{tabular}
\caption{Bayesian Regression Predictors and Coefficients}
\label{table:bayesian_regr}
\end{table}

\section{Model Sampling Details}
\label{appendix:larger_models}

\subsection{Subsampling experiment}
\begin{figure}
    \begin{center}
        \includegraphics[width=7.5 cm]{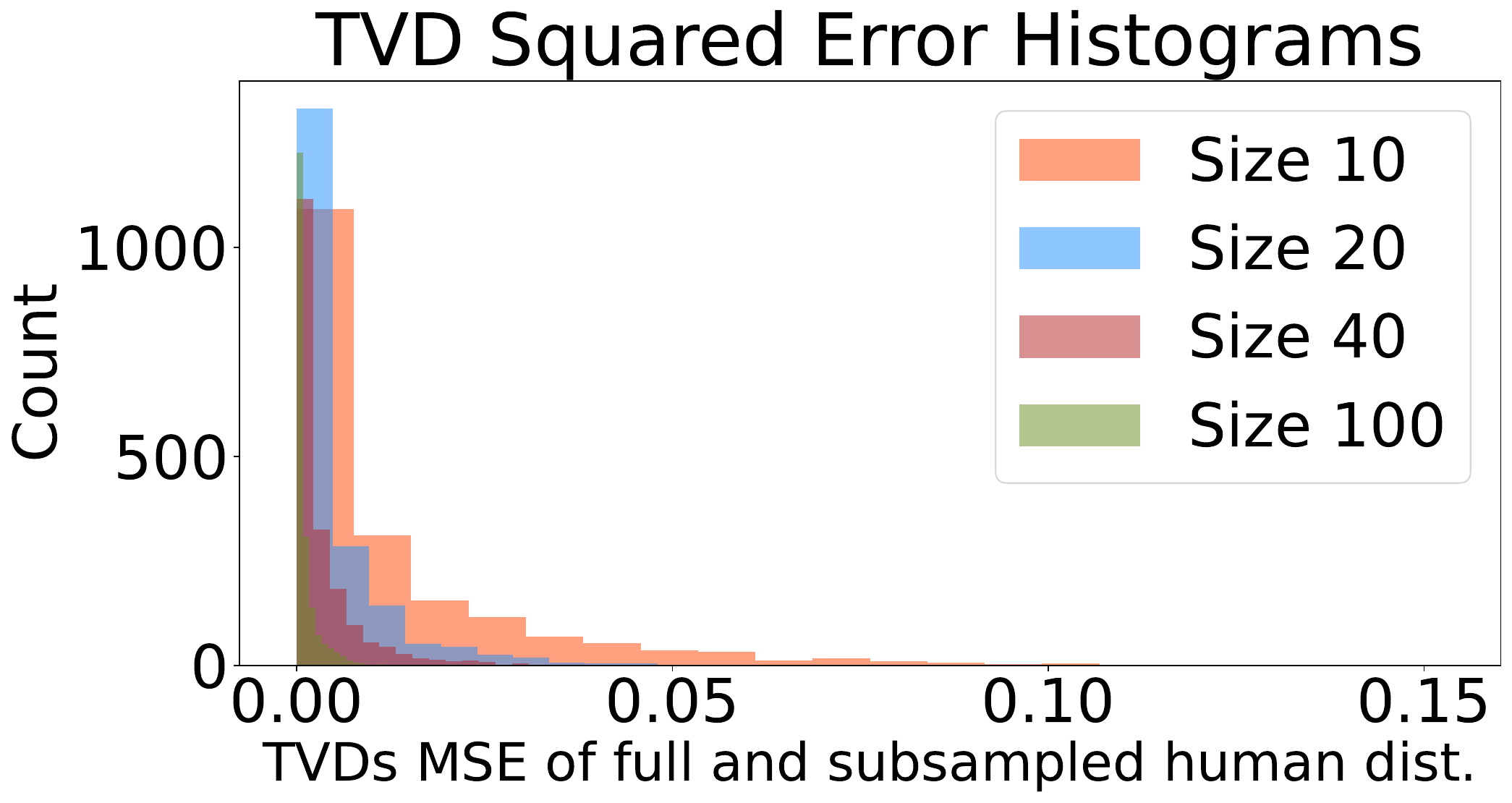}
        \caption{Histograms of MSE values between TVD values}
        \label{fig:mse}
    \end{center}
\end{figure}

Due to the high computational inference costs of large models, sampling 1000 ancestral generations for each context is infeasible. Hence, we opt for a lower number of samples - chosen on the basis of a subsampling experiment based on GPT-2. From the 1000 ancestral samples, we randomly selected subsamples of varying sizes (size = 10, 20, 40 and 100). For each of these, we re-computed the model distribution and computed the TVD values with an oracle. The Mean Squared Error between the TVD values of the subsampled distributions and the full-sampled distributions were computed and visualised through a histogram, as seen in Figure \ref{fig:mse}. We opted for a sample size of 40, since we considered it to be a good trade-off between computational costs and error.

\subsection{ChatGPT prompting}
Since ChatGPT is a conversational model - we prompt it to provide us with possible continuations to given contexts. We prompt it in two ways:

\small

\begin{enumerate}
  \item \texttt{You are ChatGPT, a large language model trained by OpenAI. I want you to answer which word is a plausible continuation to the context <CONTEXT>. I have no specific intent, I just want your guess. Return only the word and nothing else.}
  \item \texttt{You are ChatGPT, a large language model trained by OpenAI. I want you to answer which 40 words are plausible continuations to the context <CONTEXT>. I have no specific intent, I just want your guess. Return only the words and nothing else.}
\end{enumerate}
\normalsize

For the former, we request 40 generations and for the latter only one (for both, temp = 1); both ways returning eventually 40 continuations - which are ensured to be whole words. The first procedure imitates unbiased sampling more closely than the second - but due to the fact that minimal variability was observed, we implemented both methods. 

\subsection{Statistics of failed generations}
Rejecting samples that failed to generate a full word proved to be a quite rare occurrence and it mostly corresponded to producing the ‘end of sentence’ marker rather than failing to compute a full word. More specifically, for GPT2 generations, 0.05\% times we failed to produce a full word (1489 out of 2.7 million times). For Bloom, 0.2\% of times we failed to produce a word, (56 out of 27k generations), and for ChatGPT 0.04\% of times (7 out of  20k generations) - for the ‘unbiased’ sampling. ‘Diverse’ sampling did not necessarily ‘fail’ to generate any full words, but sometimes the model returned less than 40 words despite being prompted to return 40.

\subsection{TVD Differences}
\begin{figure}[h]
    \begin{center}
        \includegraphics[width=7 cm]{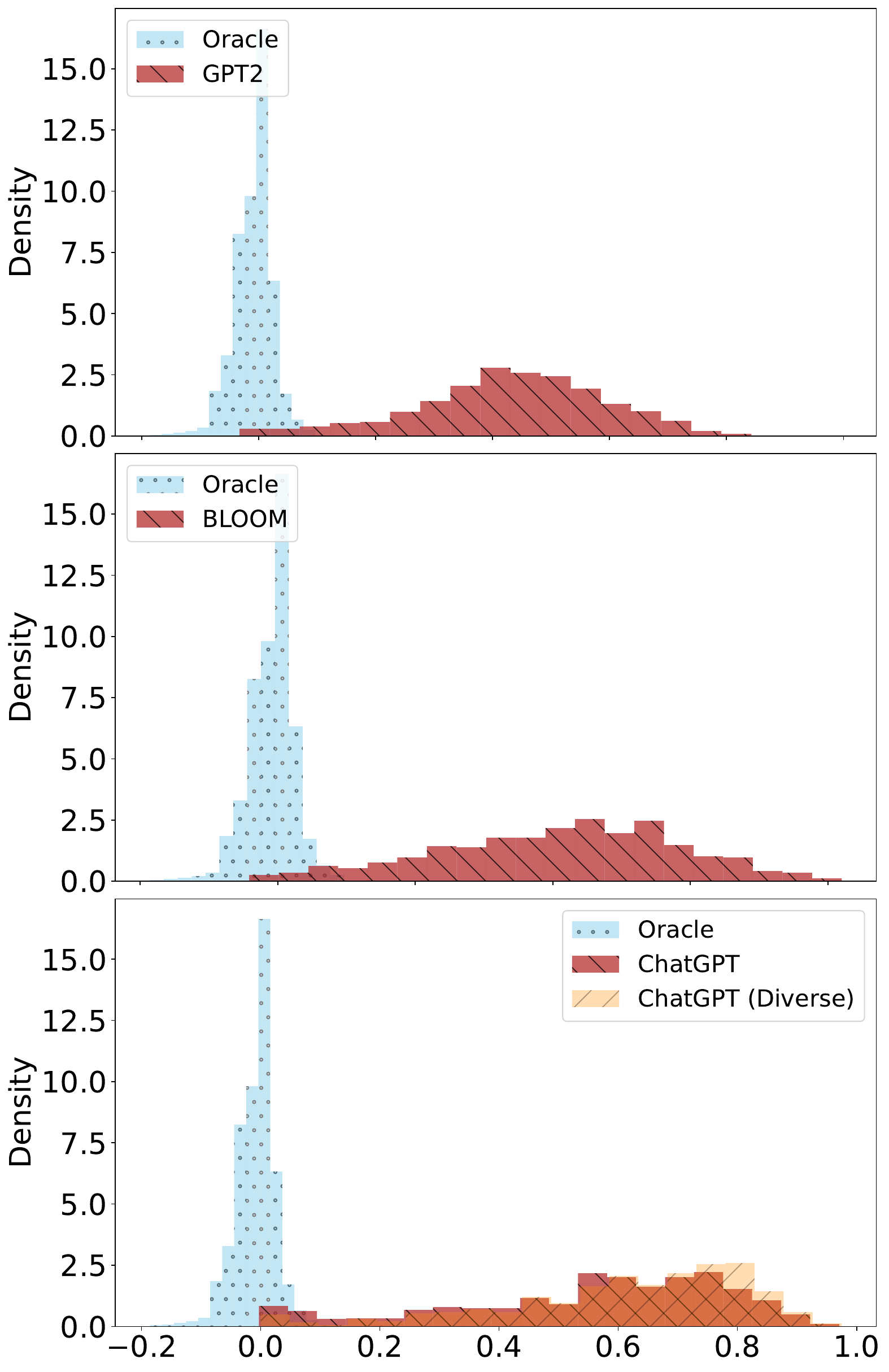}
        \caption{Histogram of TVD differences for model and oracle distributions when compared to the full human distribution. The vertical axis corresponds to density (normalizing counts so that the total histogram area equals 1).}
        \label{fig:tvd_res_cent}
    \end{center}

\end{figure}

We additionally visualise the histograms of the difference in TVD values between the model and the human distribution minus the oracle and human distributions (Figure \ref{fig:tvd_res_cent}). 

\subsection{Sampling Resources}
For both BLOOM and ChatGPT generations we used the Hugging Face and OpenAI API subscriptions respectively, for two months. Regarding GPT2, we run generations using 1 NVIDIA A100 GPU, each passage needing approximately 2 hours to compute 1000 generations for all contexts in the passage.

\section{Token-Level Experiment}
\label{appendix:token_exp}
\begin{figure}[h]
    \begin{center}
        \includegraphics[width=7.5 cm]{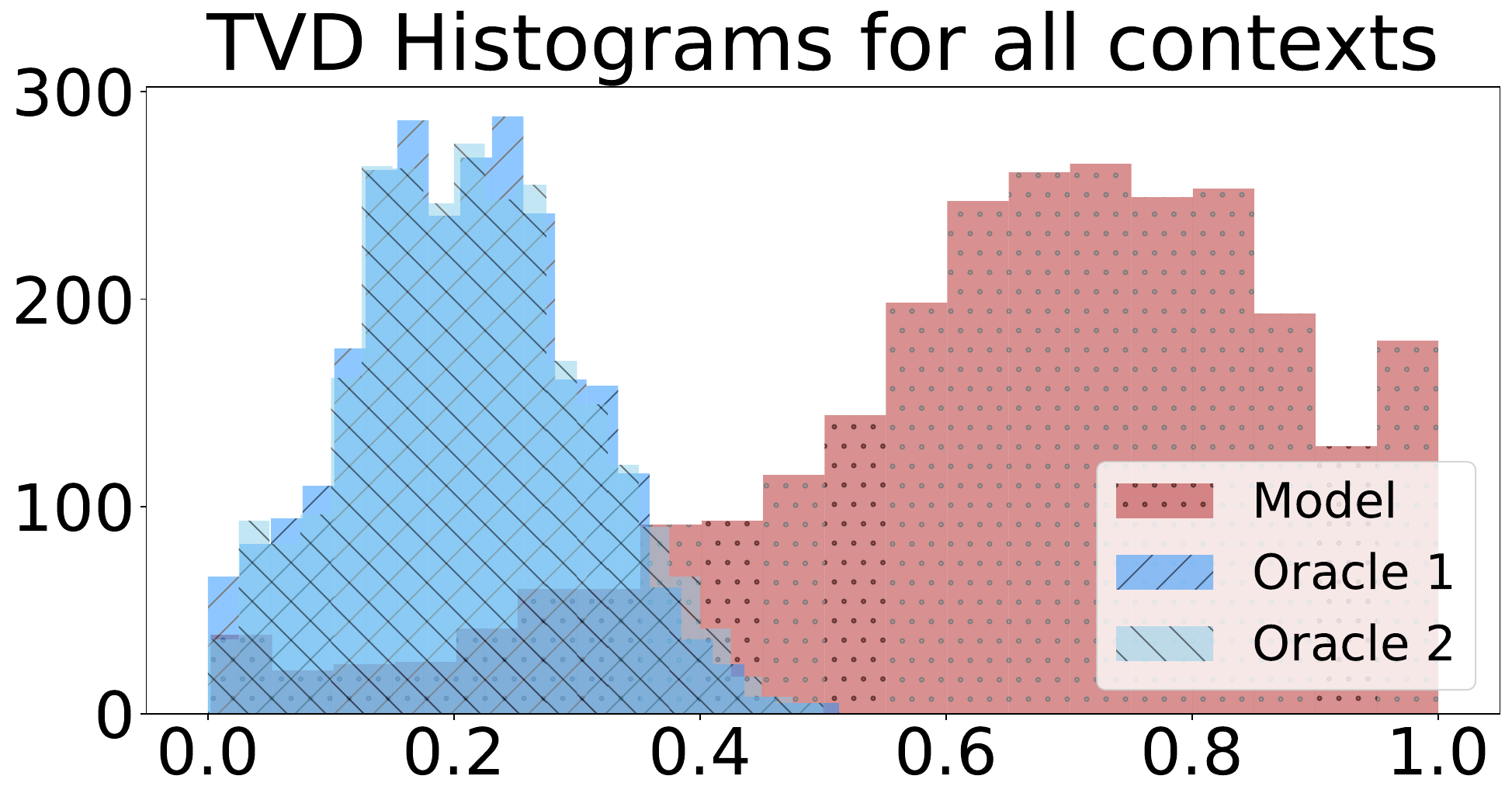}
        \caption{Histogram of TVD values for model and oracle distributions when compared to the full human distribution on a BPE-level analysis}
        \label{fig:tvd_bpes}
    \end{center}

\end{figure}

One could claim that by estimating next-word distributions instead of next-token ones, we introduce some level of bias towards the model - since they are trained on BPE tokens rather than words. Despite finding this artificial, we repeat a subset of the experiments on a token level: instead of finding a method to sample sequences of tokens that form complete words from the model, we tokenize human answers and create the target distribution of tokens. More specifically, we obtain from the model the distribution of next-tokens given a context. For the human distribution, we tokenize all human responses and take the first token of each one. We obtain the MLE of the human next-token distribution (and oracles) in a similar fashion to Section \ref{sec:methodology}. Then, we perform a similar analysis for ECE and TVD values. Results are similar to the word-level analysis (Table \ref{table:ece_bpe} and Figure \ref{fig:tvd_bpes}). We refrain from using token level analysis for calibration because it's not clear how to compare LMs with different tokenizers, whose vocabulary sizes differ.

\begin{table}[h]
    \centering
    \begin{tabular}{llll}
        \toprule
        \multirow{2}{*}{Gold Label} & \multicolumn{3}{c}{ECE} \\
        {} & Model & Oracle 1 & Oracle 2 \\
        \midrule
        Human Majority & 0.141 & 0.500 & 0.396 \\
        \bottomrule
    \end{tabular}
\caption{ECE results for the Biased Model and Oracle Distributions}
\label{table:ece_bpe}
\end{table}

\section{Improving Model Experiments}
\label{appendix:improving_model}

\begin{figure}
    \begin{center}
        \includegraphics[width=7.5 cm]{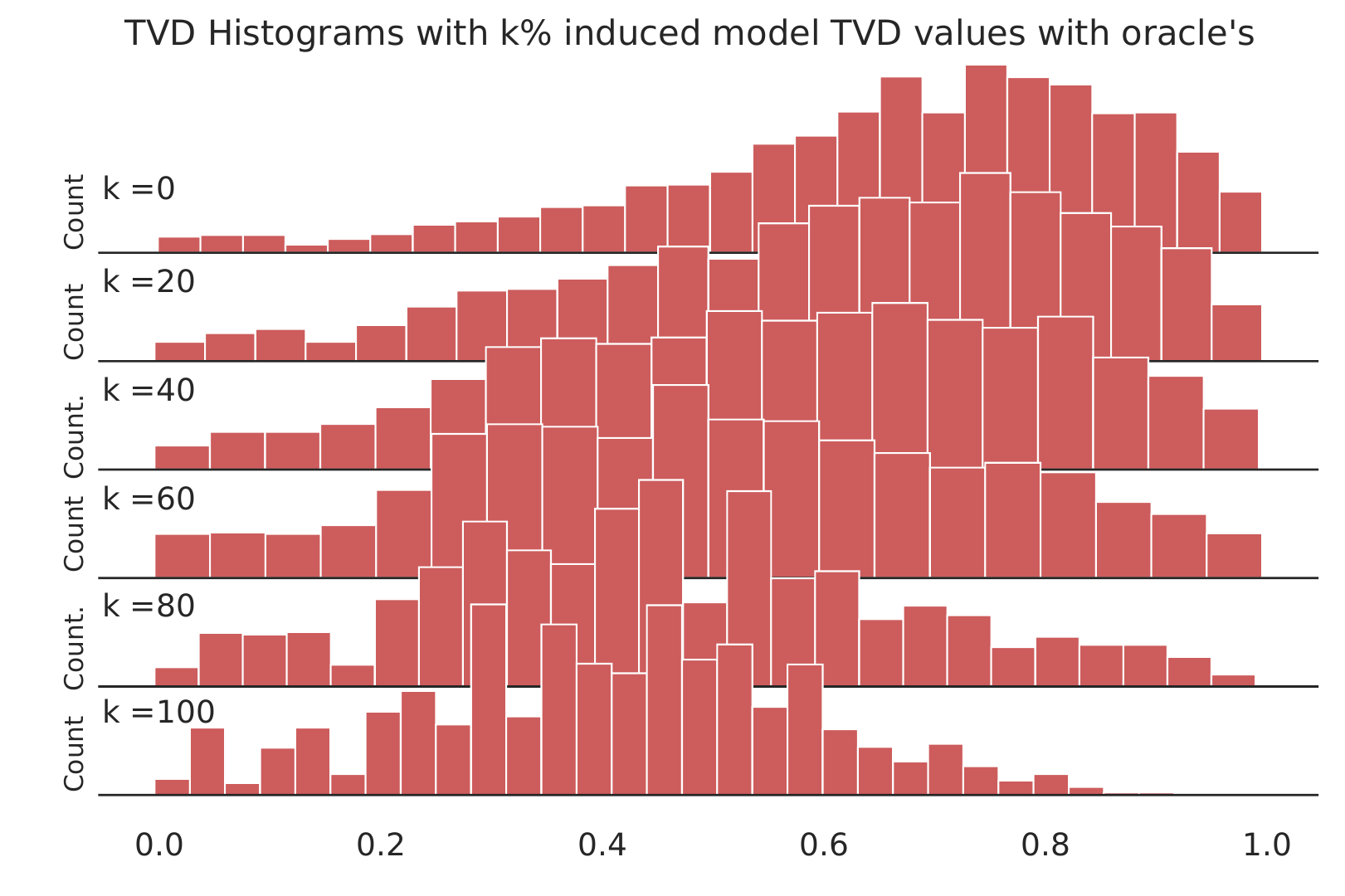}
        \caption{We artificially improve the Model-Oracle TVD histogram, by randomly replacing k\% of the TVD values with the respective TVD values between oracles.}
        \label{fig:improved_model}
    \end{center}
\end{figure}

\begin{figure}
\includegraphics[width=7.4 cm]{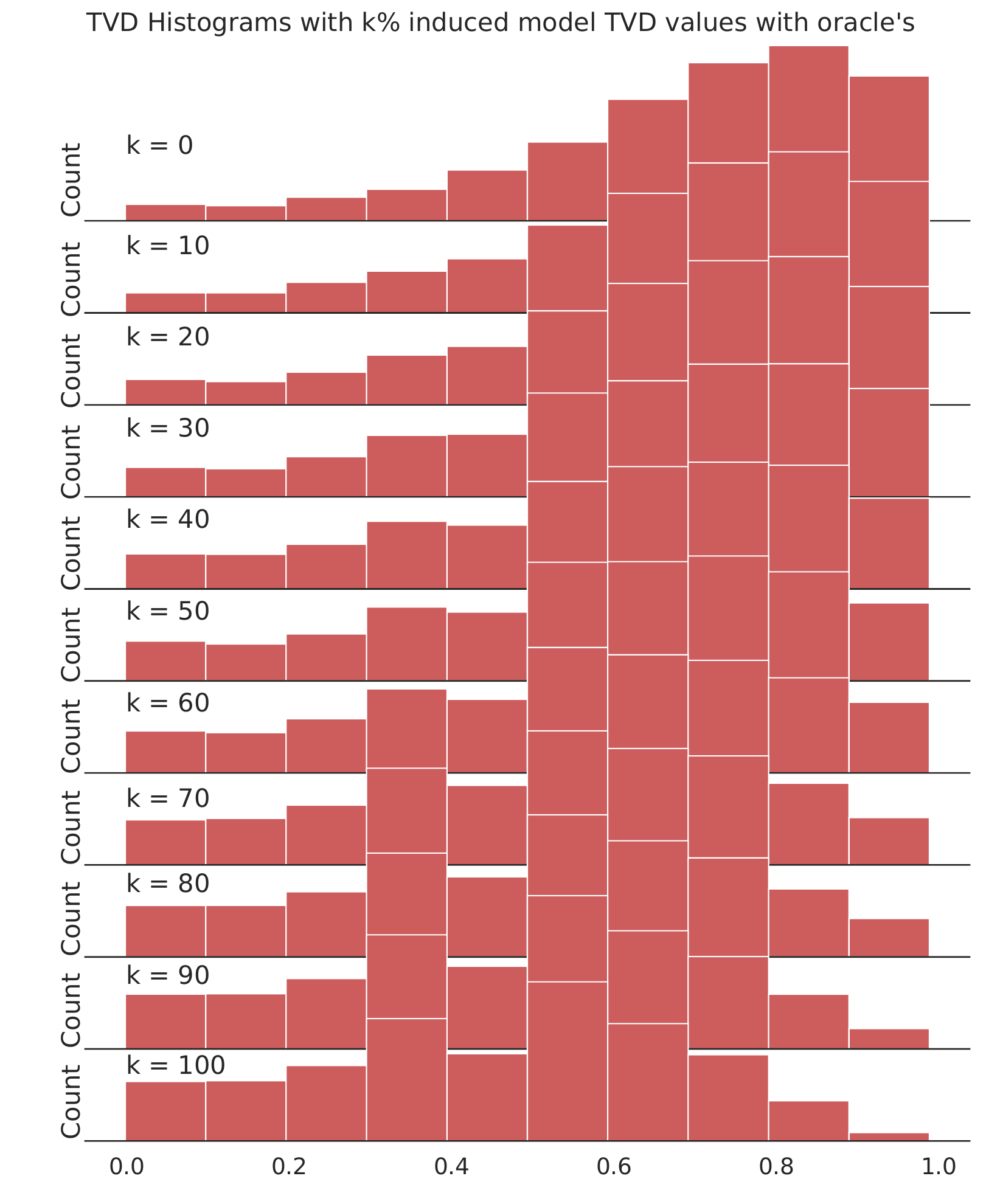}
\caption{Improving the Model-Oracle TVD histogram, by randomly replacing k\% of the TVD values with the respective TVD values between oracles, with an oracle size of 10, repeated on 10 seeds. k=0 corresponds to model performance and k = 100 to human performance.}
\label{fig:imp_oracle_size_10}
\end{figure}

\begin{figure}
\includegraphics[width=7.5 cm]{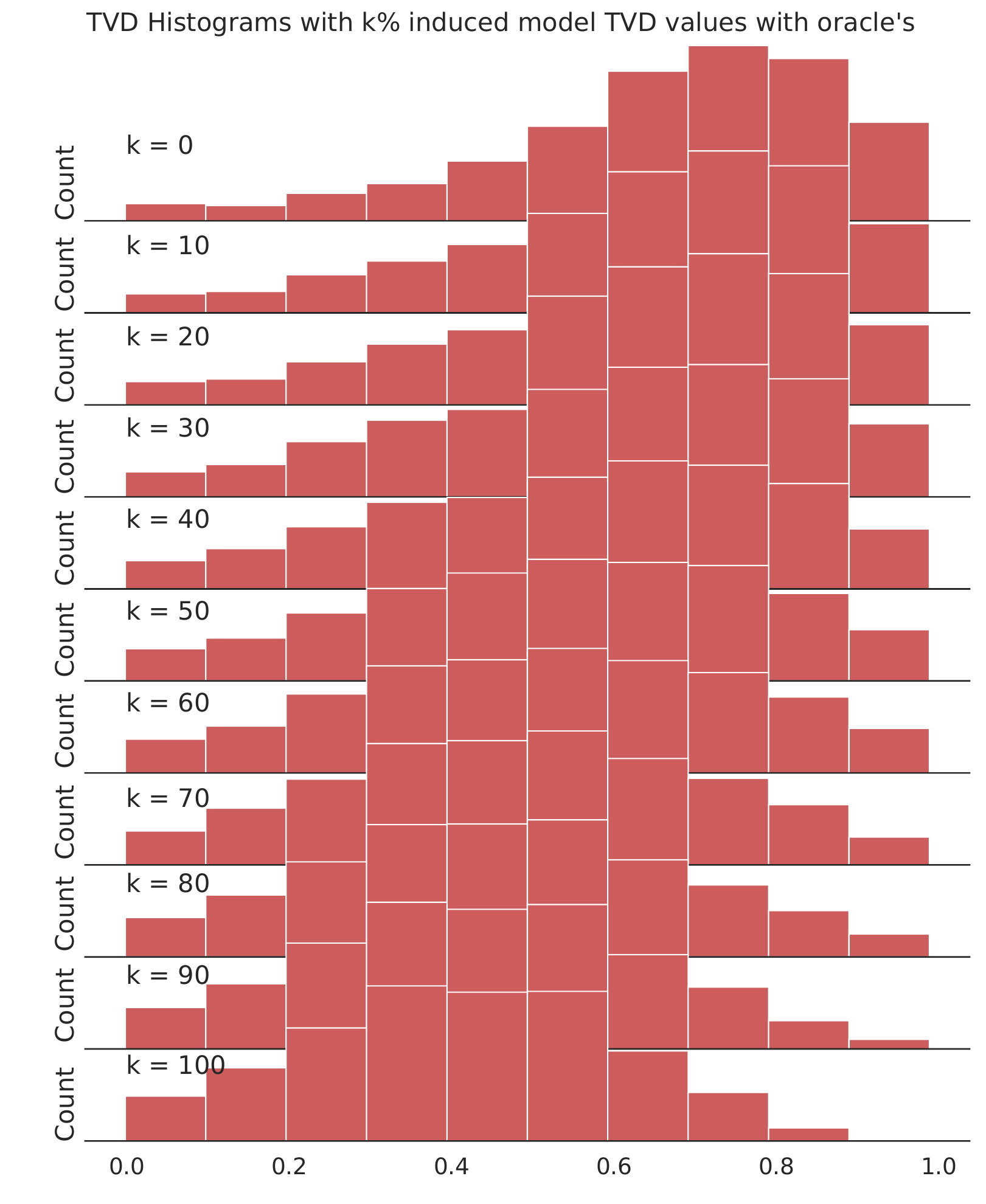}
\caption{Improving the Model-Oracle TVD histogram, by randomly replacing k\% of the TVD values with the respective TVD values between oracles, with an oracle size of 20, repeated on 10 seeds. k=0 corresponds to model performance and k = 100 to human performance.}
\label{fig:imp_oracle_size_20}
\end{figure}

We repeat the experiment where we artificially improve GPT2's performance (Section \ref{sec:measure_calibration}). This time, we create two types of disjoint oracles (by sampling from the human cpd without replacement) varying in size - a pair of size 20 and a pair of size 10. For each size, we sample 10 different pairs (using different seeds). For each pair, we compute the TVD value between them and the TVD value between an oracle and the model. As before, we randomly choose k\% of model-oracle TVD instances to be replaced by the respective oracle-oracle instances. The aggregated results for the 10 seeds can be found in Figures \ref{fig:imp_oracle_size_10} and \ref{fig:imp_oracle_size_20} for the oracles of size 10 and 20 respectively. Results are very similar as before, showing that results are robust to the oracle size and the sampled split itself.

\section{Out-Of-Distribution Effect Experiment}
\label{appendix:fine_tuning}
One could claim that we evaluate on a dataset, Provo Corpus, that does not necessarily originate from the distribution of the training dataset. To reinforce the validity of our results and establish that they are not just stemming from a domain mismatch of training and evaluation data, we complete experiments by fine-tuning on a subset of Provo Corpus. This way we, at least partly, remove the potential out-of-distribution effect - and re-evaluating calibration. Due to the Provo Corpus' limited size, the fine-tuning procedure has the following two aspects: 

(1) A k-fold cross validation split (k=4), using the first 40 passages (Paragraphs 1-40) of Provo Corpus to create the 4 equal splits - each 10 passages long. We iteratively train on 3 of the splits and evaluate on the last 15 passages of Provo Corpus (Paragraphs 41-55). The paragraphs from the unused split are used for the evaluation of uncertainty. Overall, we end up with 4 different models, each used to create model distributions for 10 paragraphs - which, in turn, are used to measure TVD values for all their contexts.

(2) We do not fine-tune on the whole model - we freeze all parameters except those of the last two layers of GPT2-Small, since our training dataset is very small. We train using the cross-entropy loss, the AdamW optimizer (epsilon = 1e-8), for 10 epochs, with a 5e-4 learning rate, a batch size of 5, using 0 as the seed value.

\begin{figure}[t]
\includegraphics[width=7.4 cm]{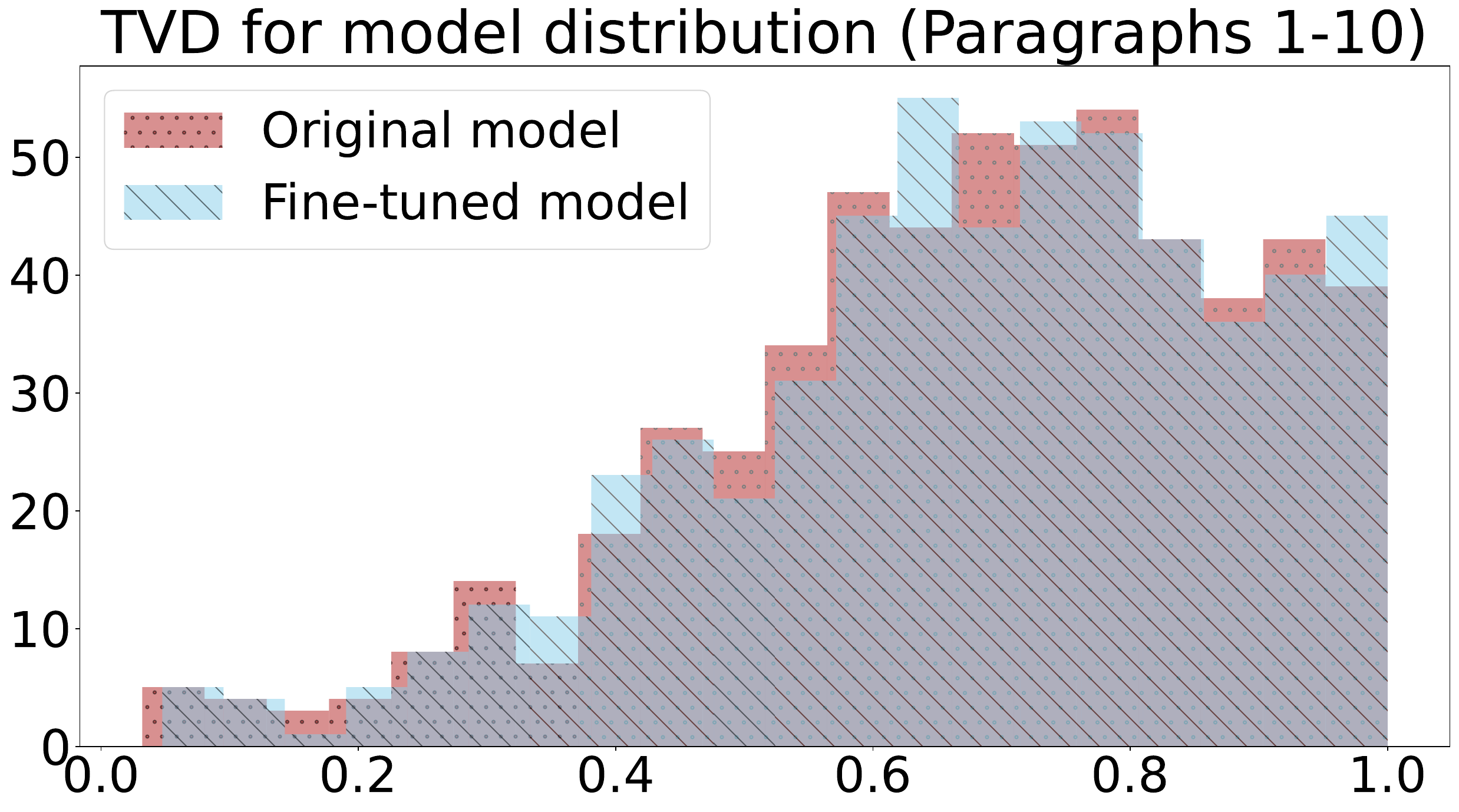}
\includegraphics[width=7.5 cm]{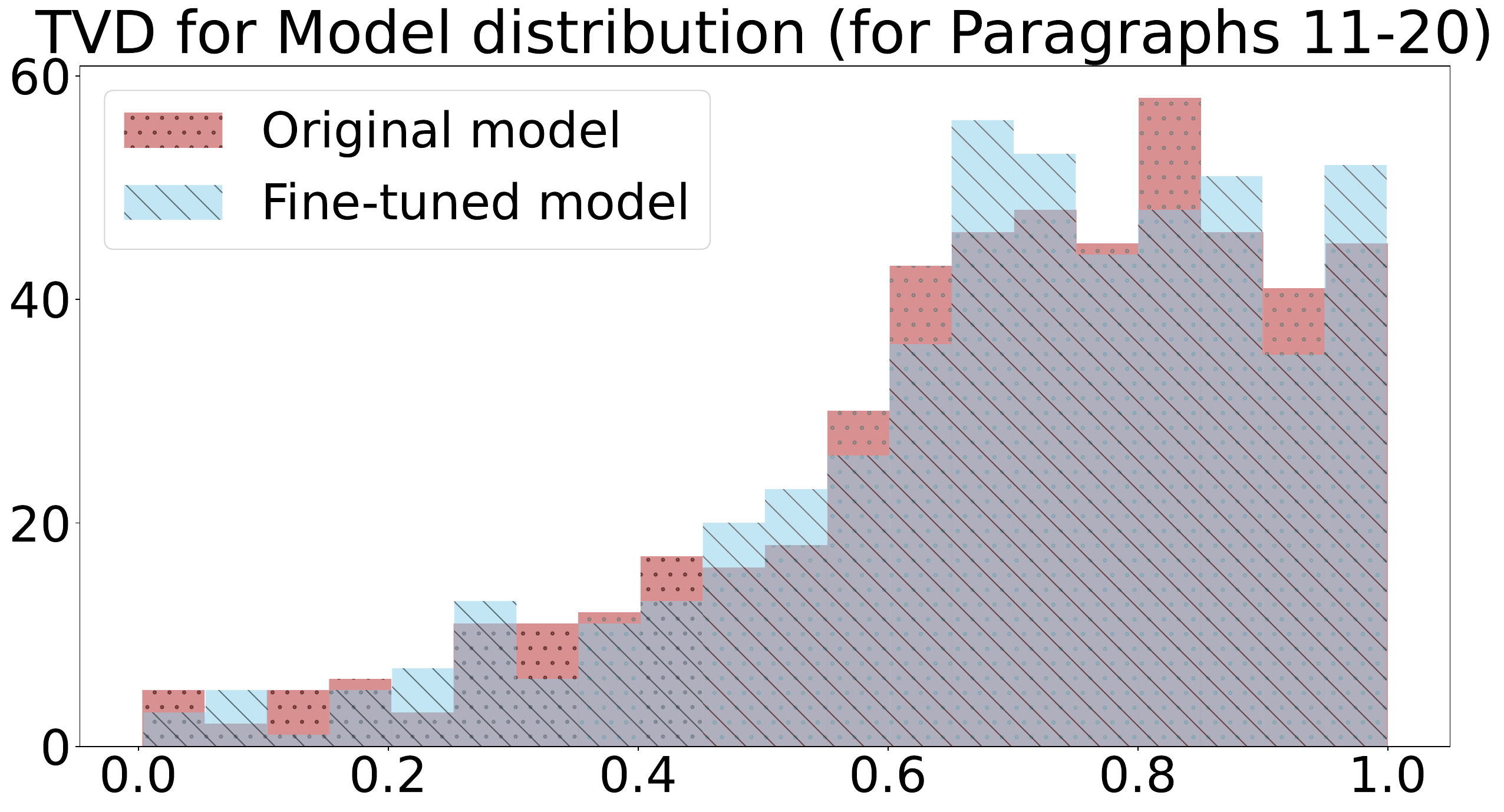}
\includegraphics[width=7.5 cm]{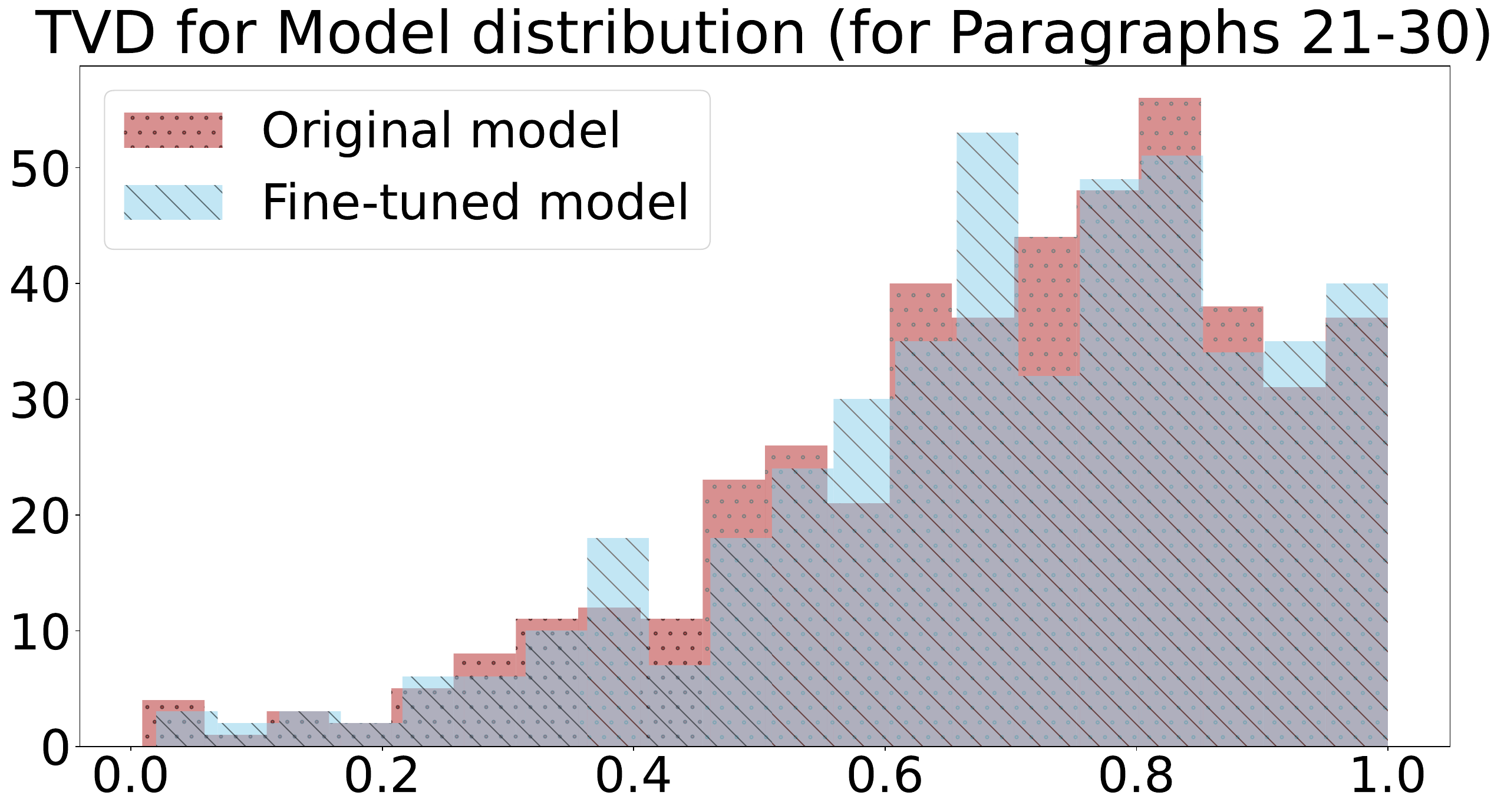}
\includegraphics[width=7.45 cm]{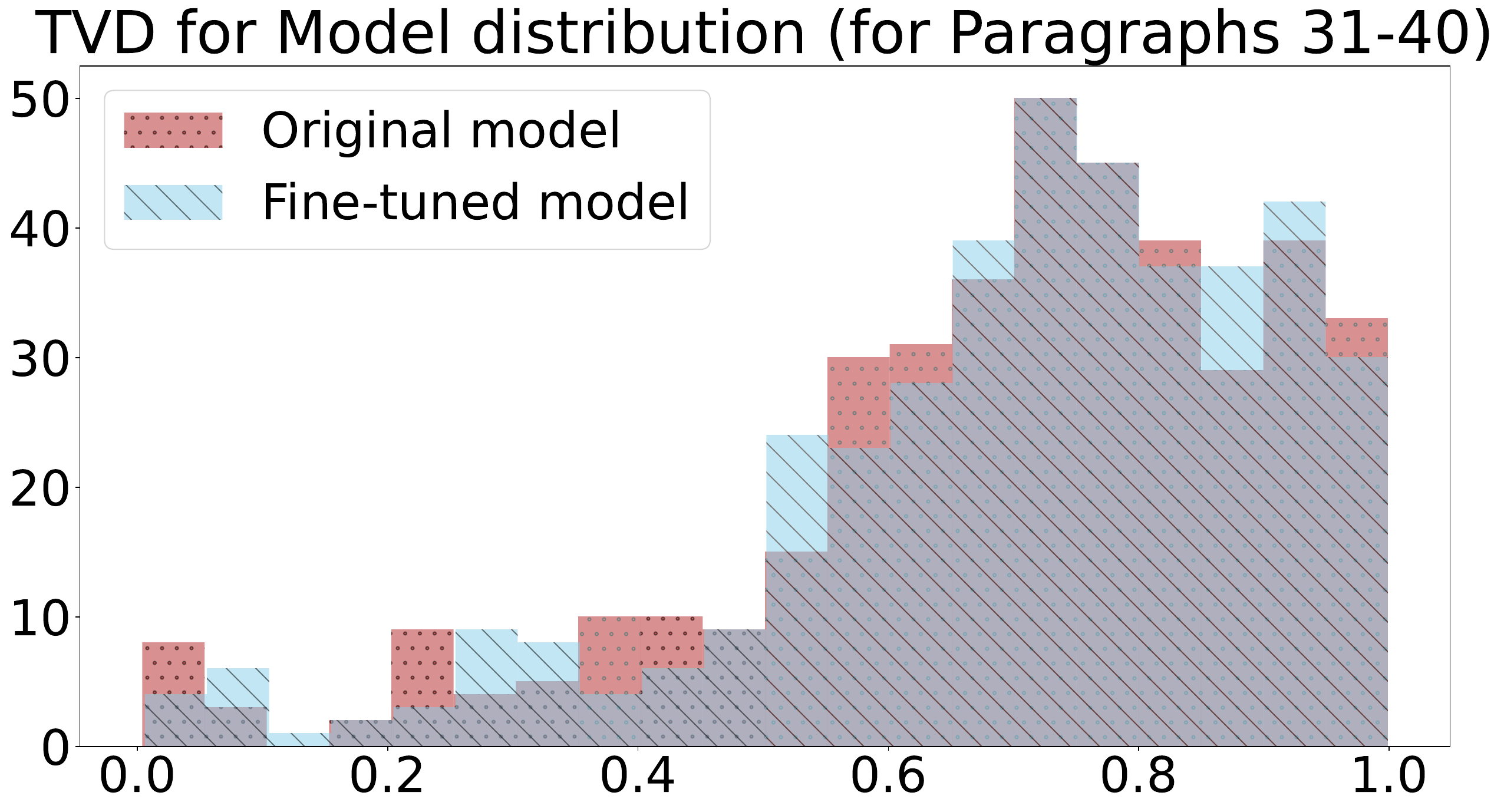}
\caption{TVD histograms for all contexts between models (original and fine-tuned) and humans}
\label{fig:fine_tuned}
\end{figure}

\begin{figure}[t]
\includegraphics[width=7 cm]{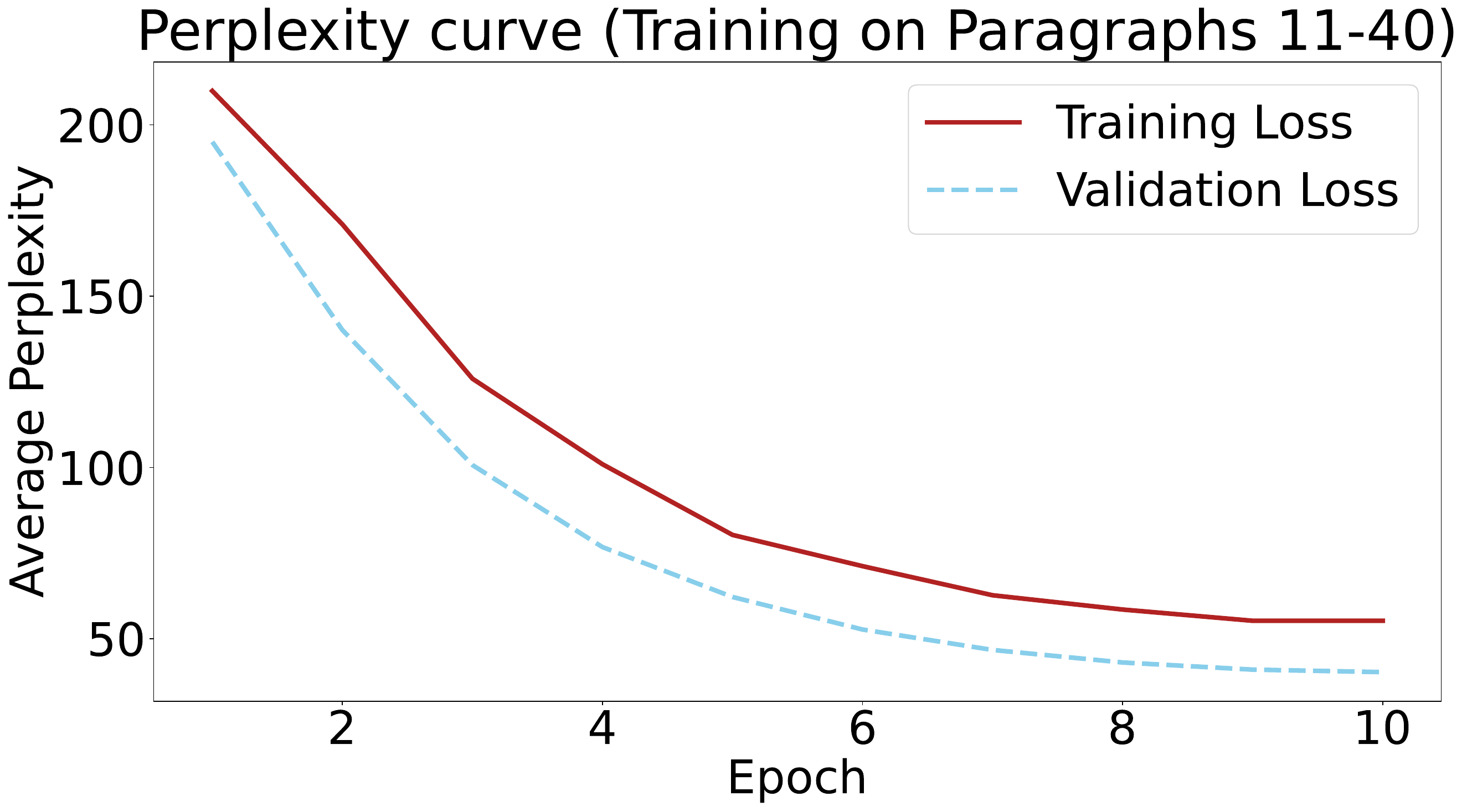}
\includegraphics[width=7 cm]{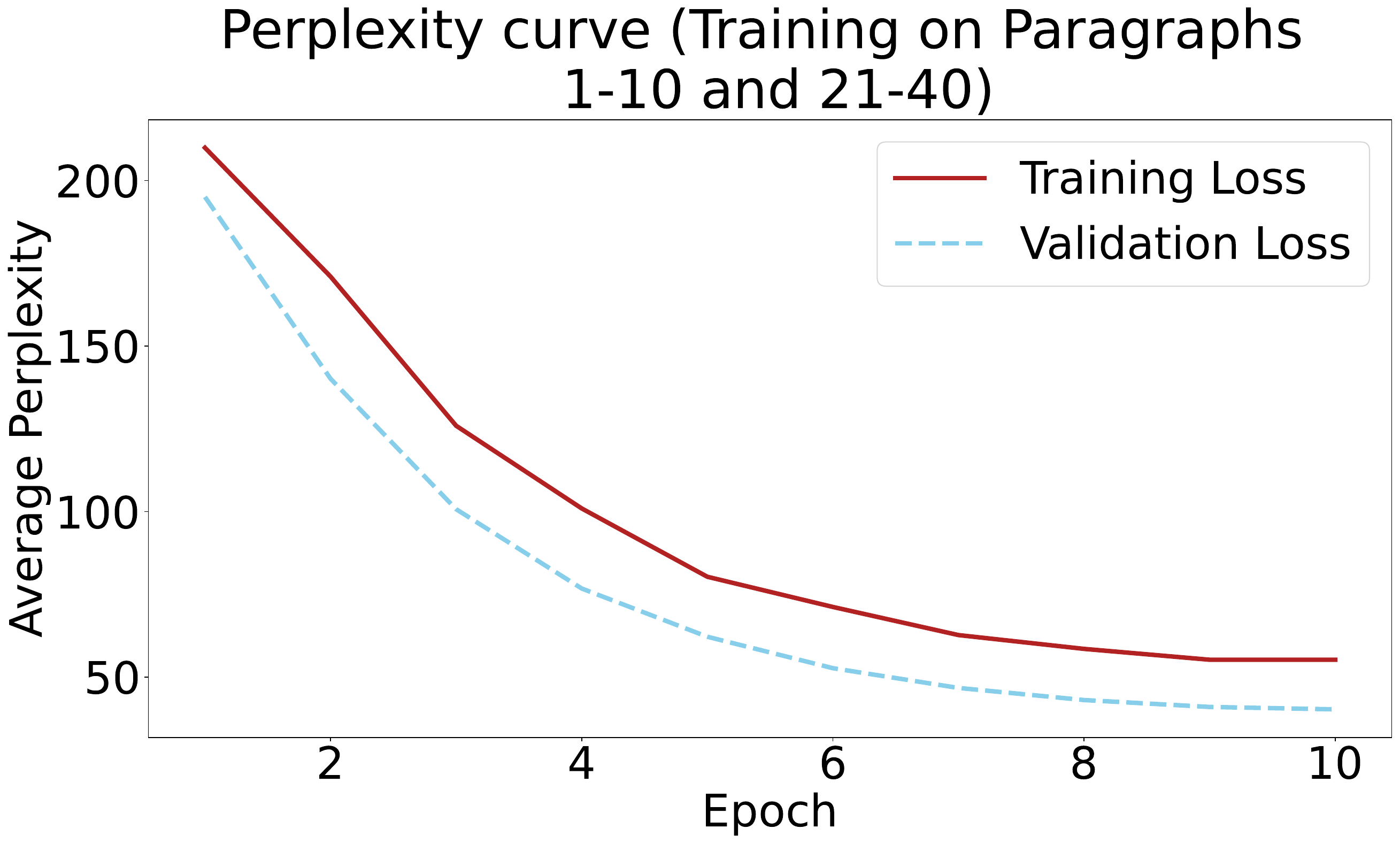}
\includegraphics[width=7 cm]{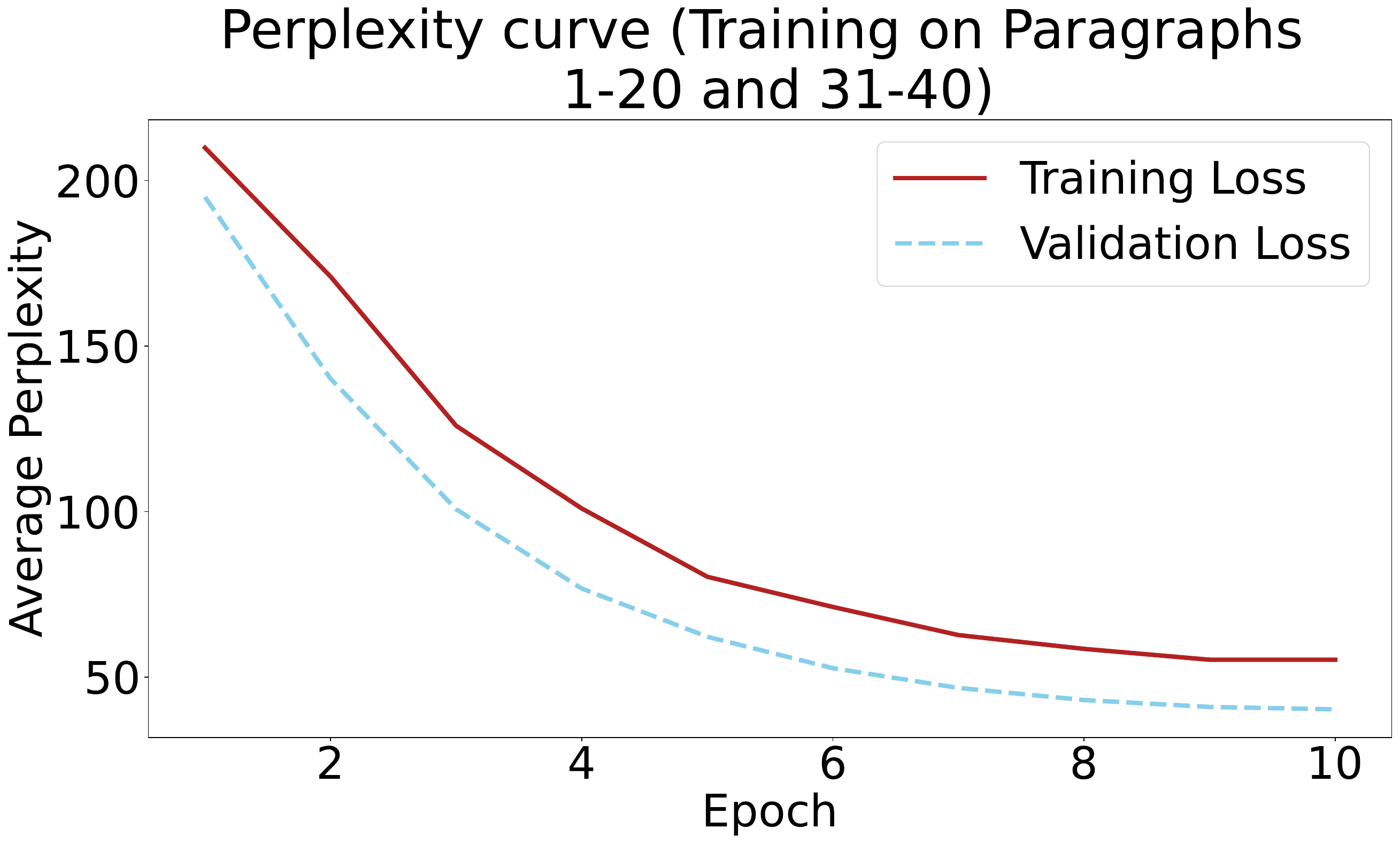}
\includegraphics[width=7 cm]{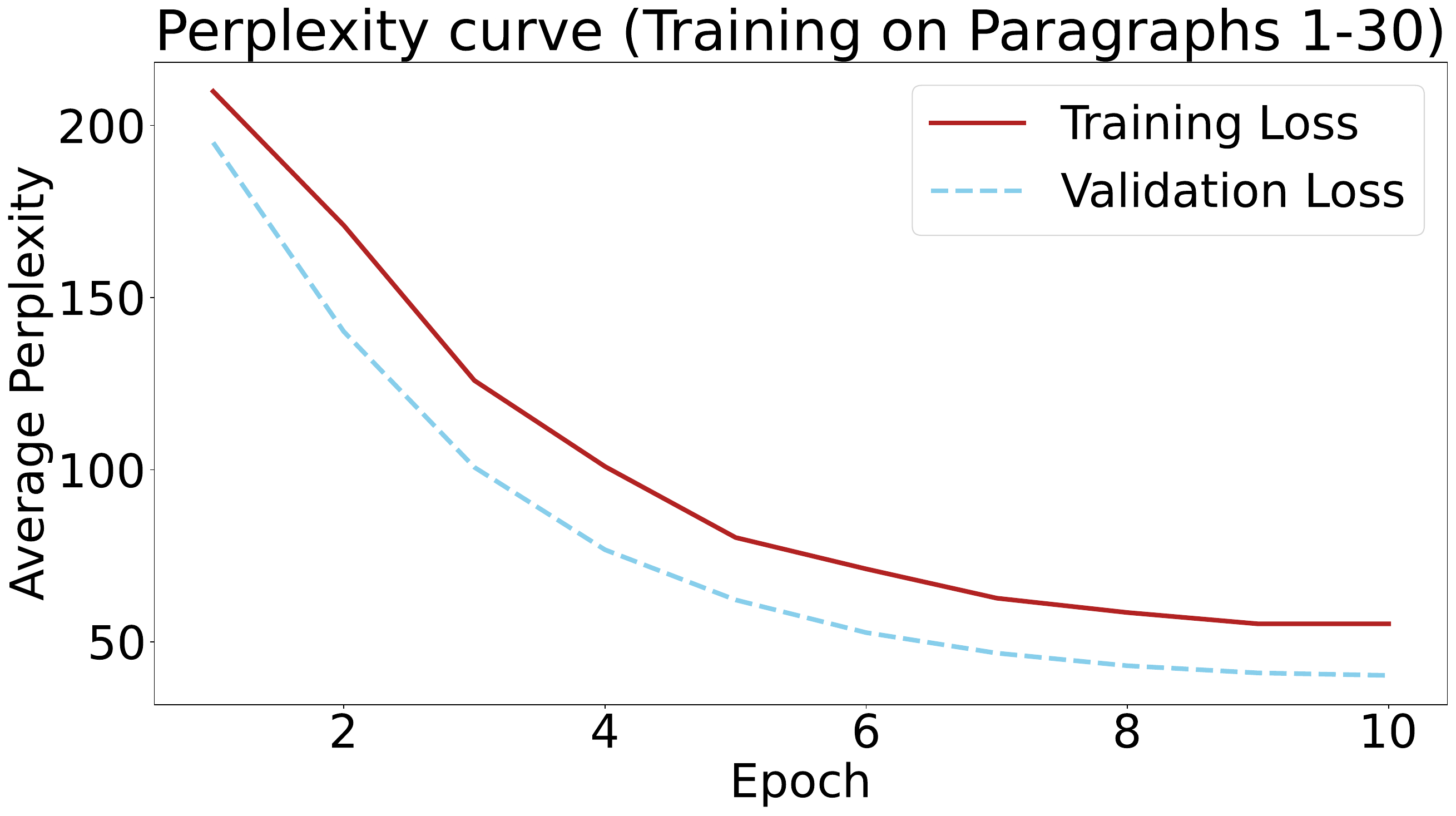}

\caption{Training and Validation loss during the fine tuning of our model on a subset of Provo Corpus}
\label{fig:loss_curves}
\end{figure}

The TVD results for the fine-tuned models', along with the respective perplexity curves during fine-tuning are in Figure \ref{fig:fine_tuned} and \ref{fig:loss_curves} respectively.

\section{Semantic \& Syntactic Analysis}
For TVD$_{syn}$, we use the default nltk POS-tagger using as arguments tagset='universal' on the concatenation of the context and each generation to obtain the POS-tag of the generation. We repeat this process for human and model generations. 

For TVD$_{sem}$, we cluster the set of human and model words using the Kmeans implementation from the sklearn library (using arguments n\_clusters, random\_state = 0, n\_init = 20, max\_iter = 400). The number of clusters was decided based on a selection of k using SSE 
(Within-Cluster-Sum of Squared Errors, i.e. Squared Error from each point to its predicted cluster center) --- incremental ks tested included k in range(2, k\_max, k\_max//3), where k\_max the number of words to be clustered. To obtain word feature representations, we use their respective word2vec embeddings ('word2vec-google-news-300' from the gensim library) --- scaled using the sklearn StandardScaler, after filtering out words without a word2vec representations. To obtain human, oracle and model distributions for each context, we assign for each cluster one element in the support (as well as one support element representing filtered-out words). The probability of the cluster elements is the summed probability of words assigned to the cluster (where probabilities are computed similarly to Section \ref{sec:methodology}).

\label{appendix:semantic_anal}

\section{Visual Analysis of Distributions}
\label{appendix:visual_dists}

We randomly choose one full passage (Paragraph 8) to illustrate further our conclusions. For all contexts, we provide the human and GPT2 distributions for the 15 most probable words of each cpd.

\newpage

\begin{figure}
    \includegraphics[width=7.8 cm]{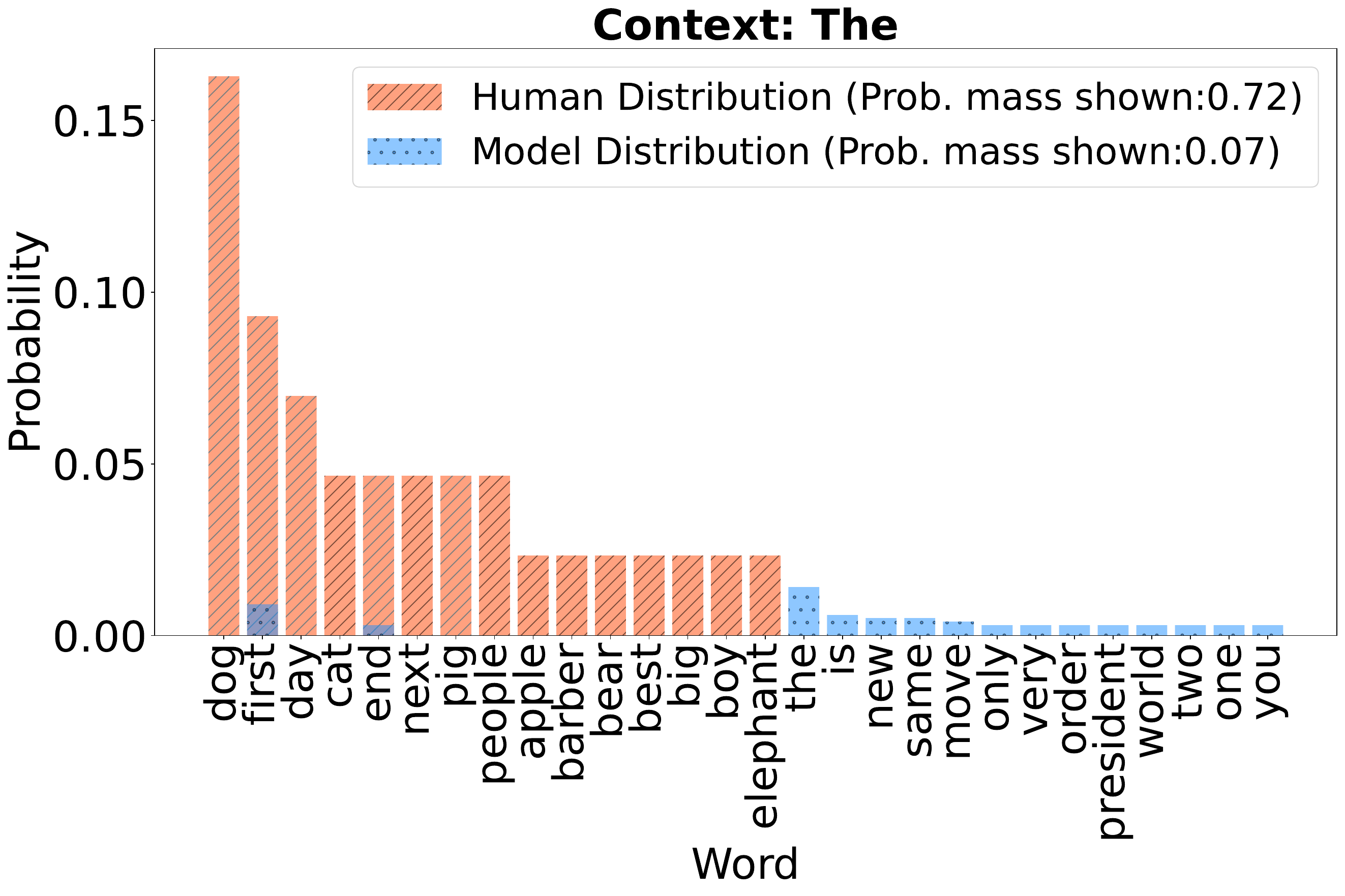}
    \includegraphics[width=7.8 cm]{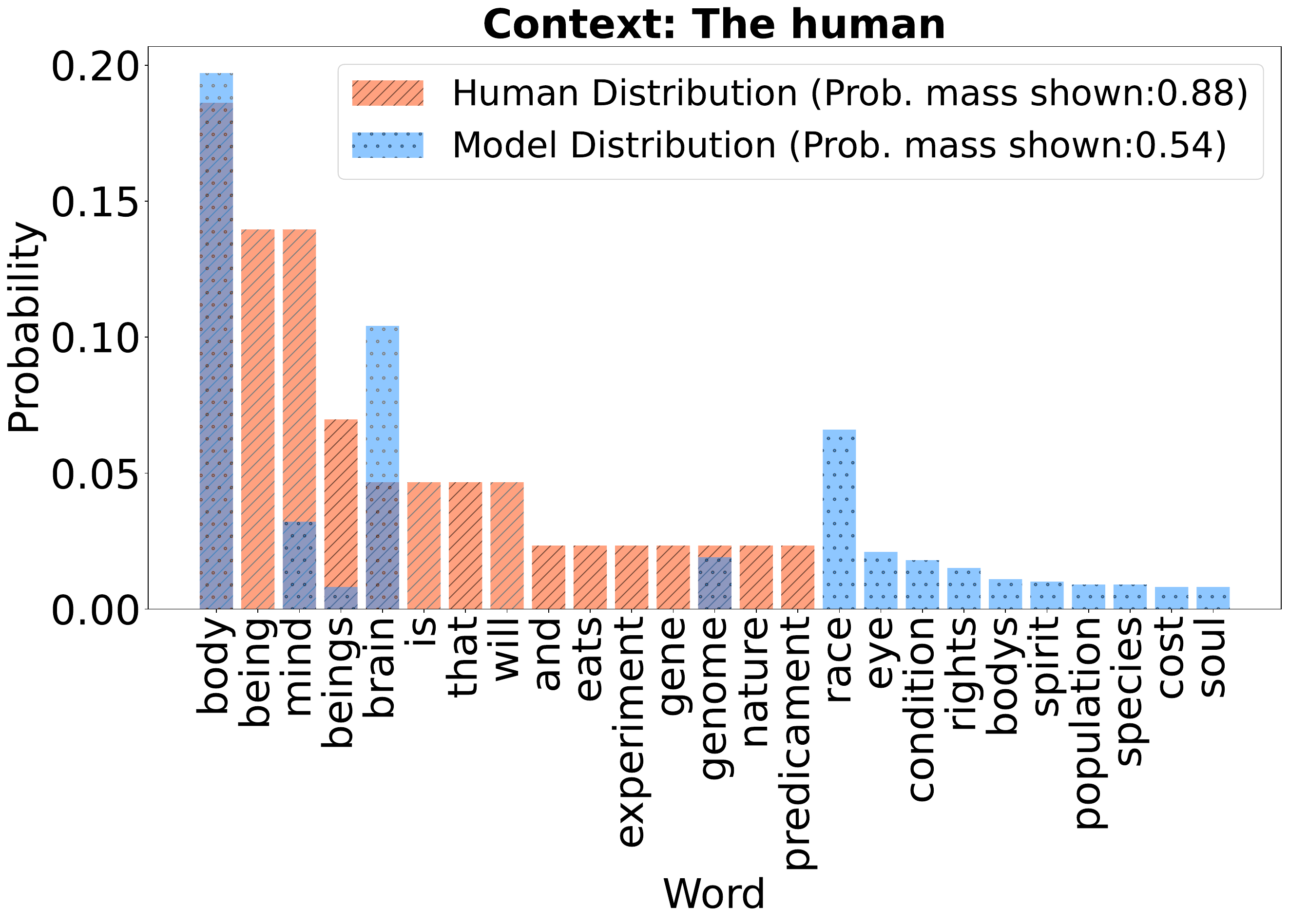} \includegraphics[width=7.8 cm]{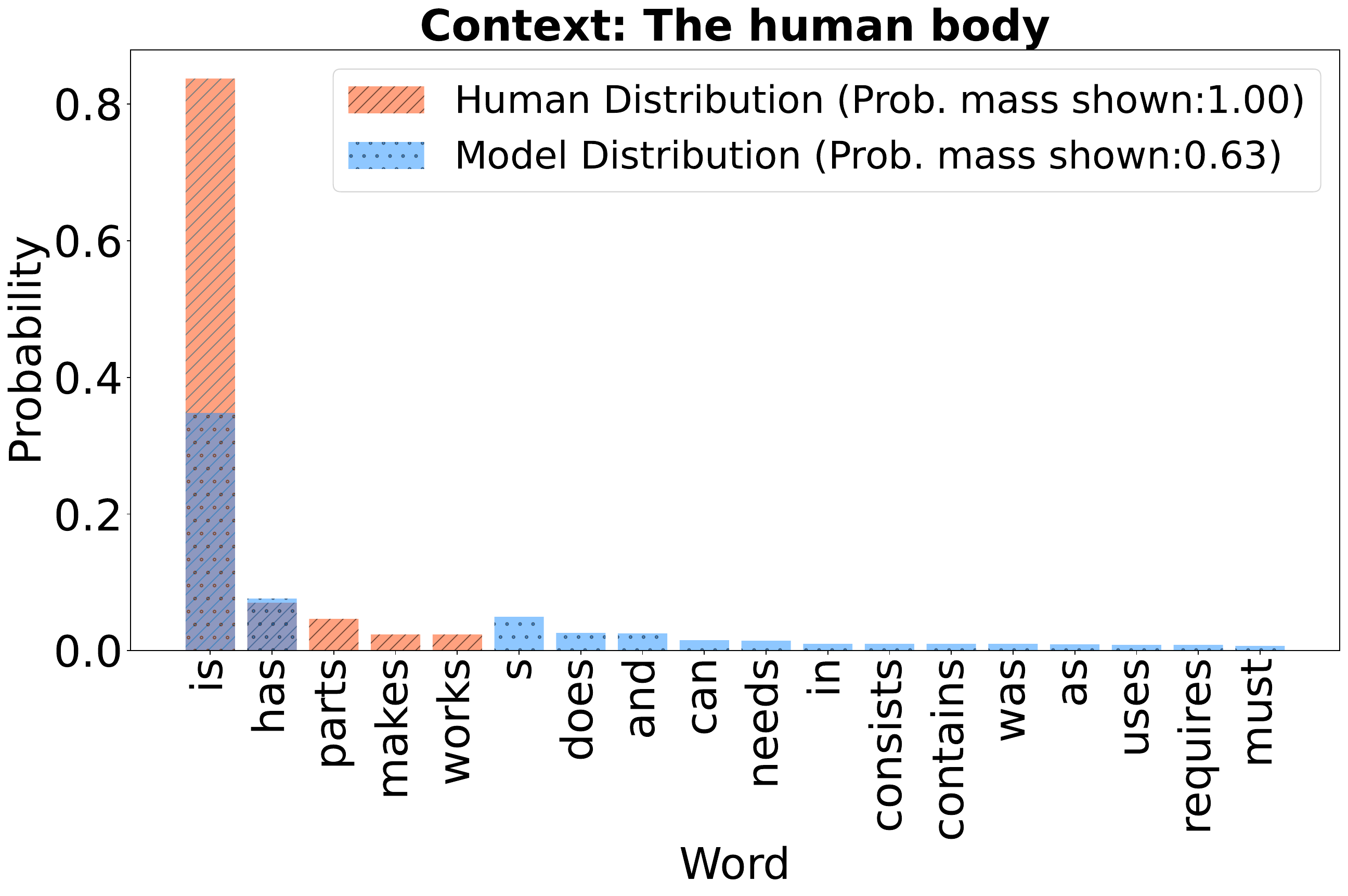}
    \includegraphics[width=7.8 cm]{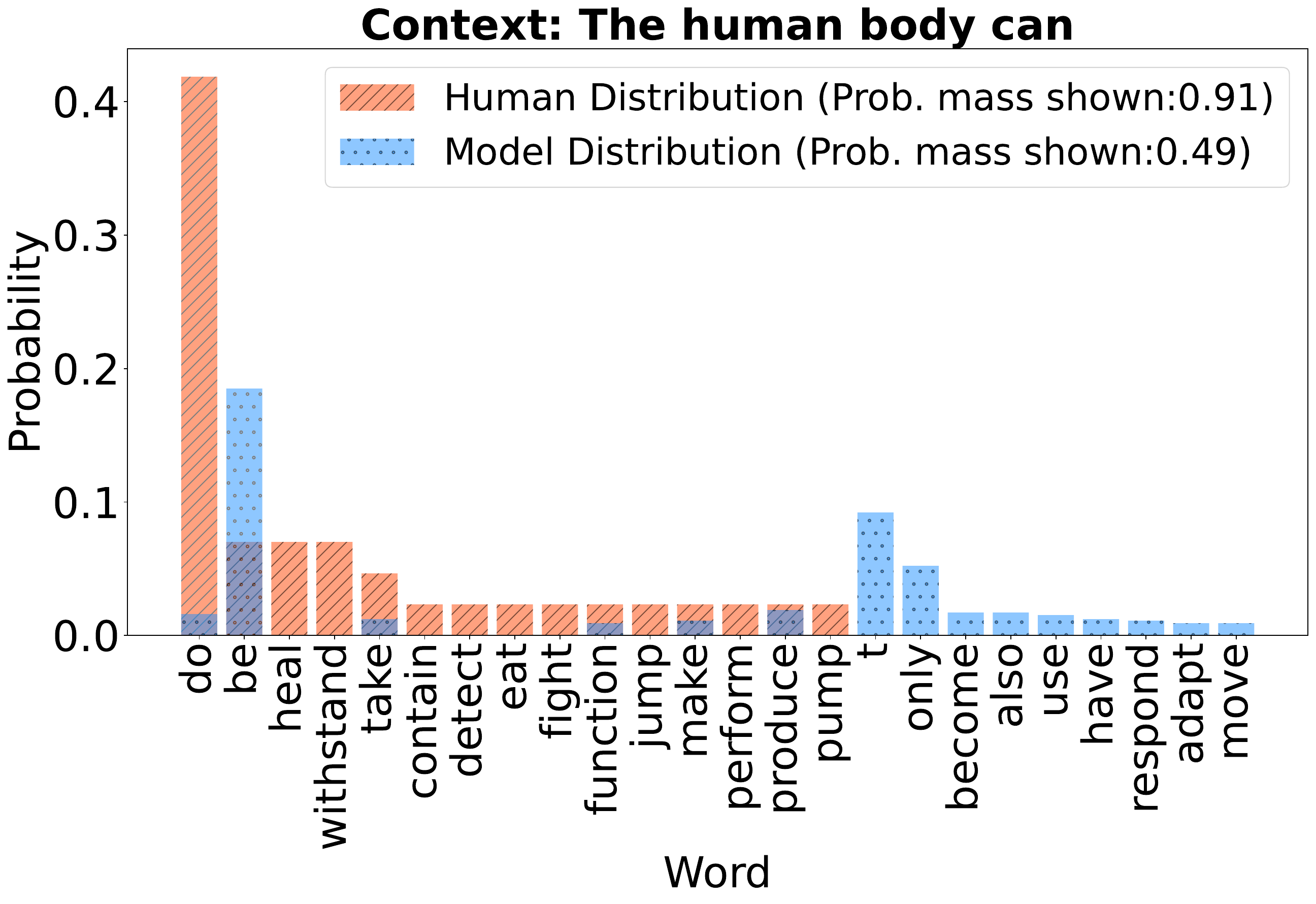}
\end{figure}

\begin{figure}
    \includegraphics[width=7.8 cm]{appendix/dists_6.pdf}
    \includegraphics[width=7.8 cm]{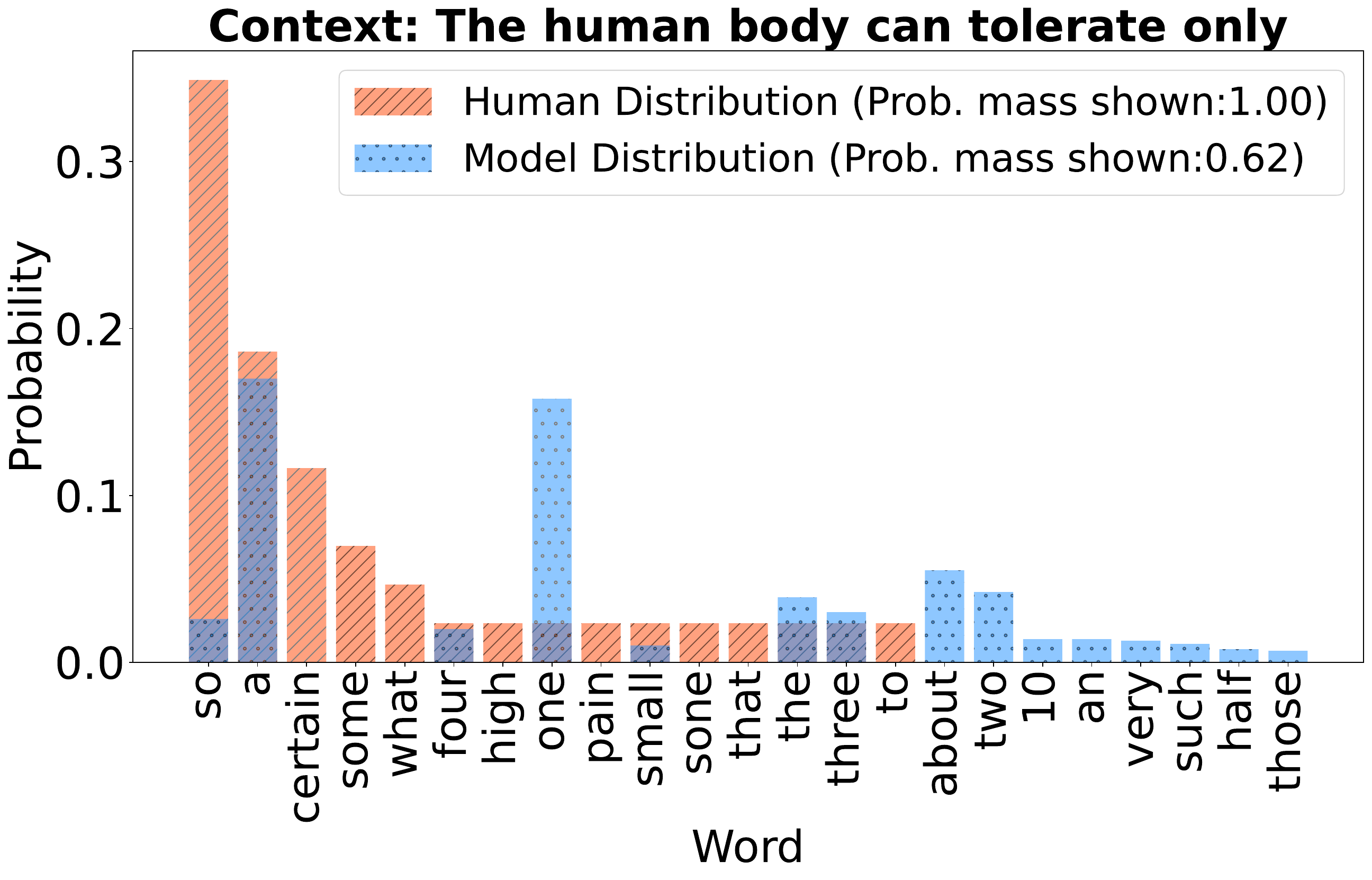} \includegraphics[width=7.8 cm]{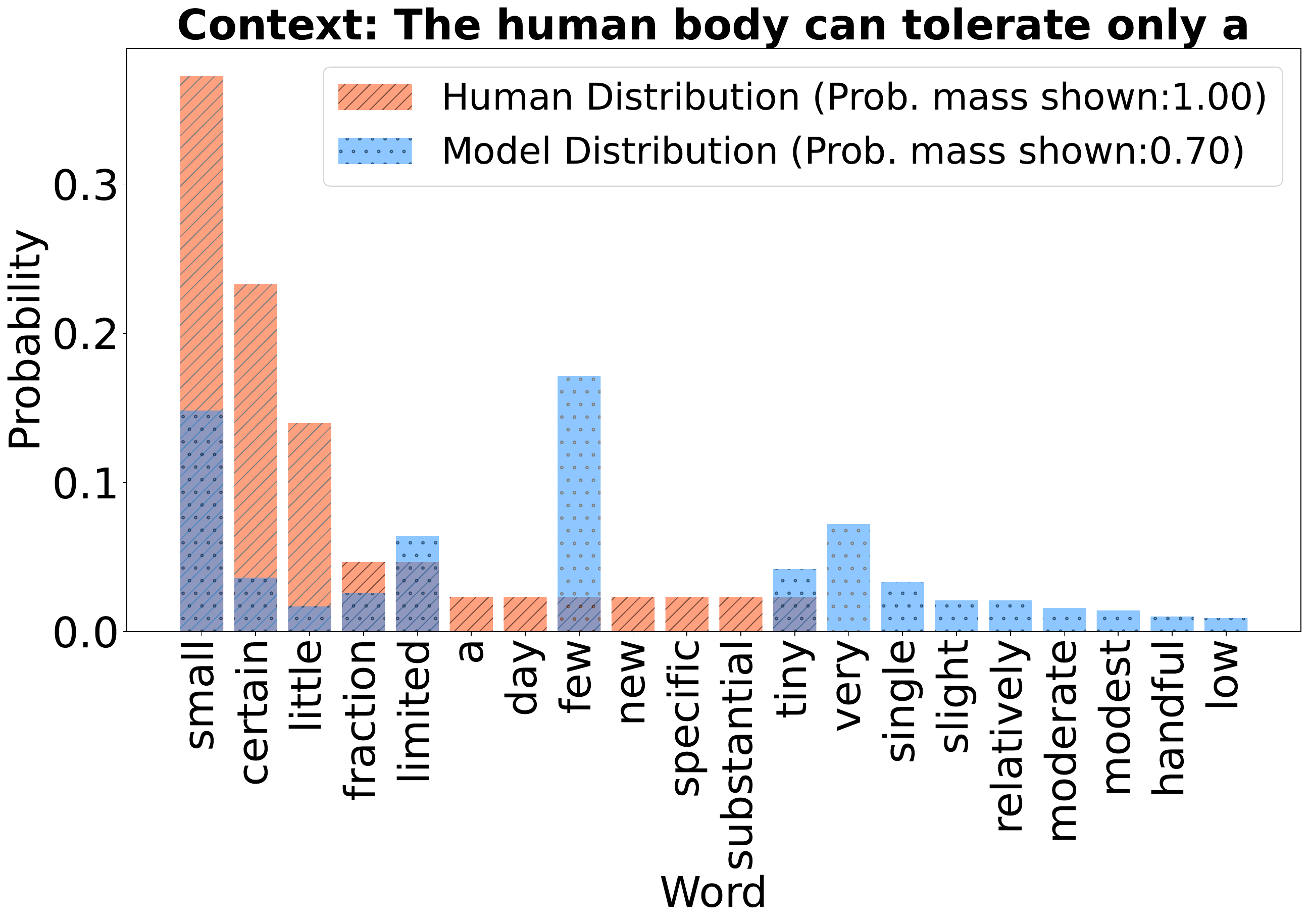}
    \includegraphics[width=7.8 cm]{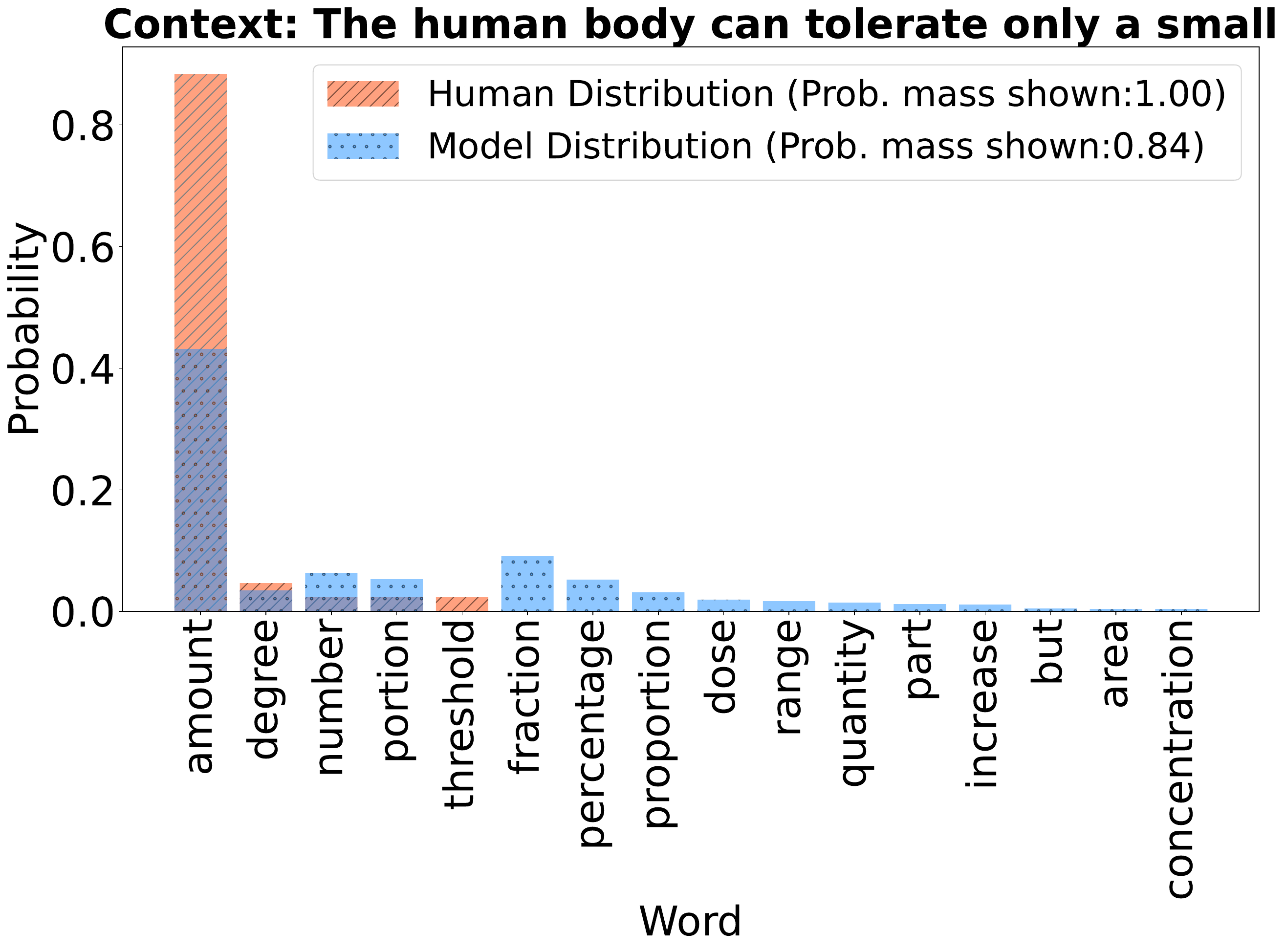}
\end{figure}

\begin{figure}
    \includegraphics[width=7.8 cm]{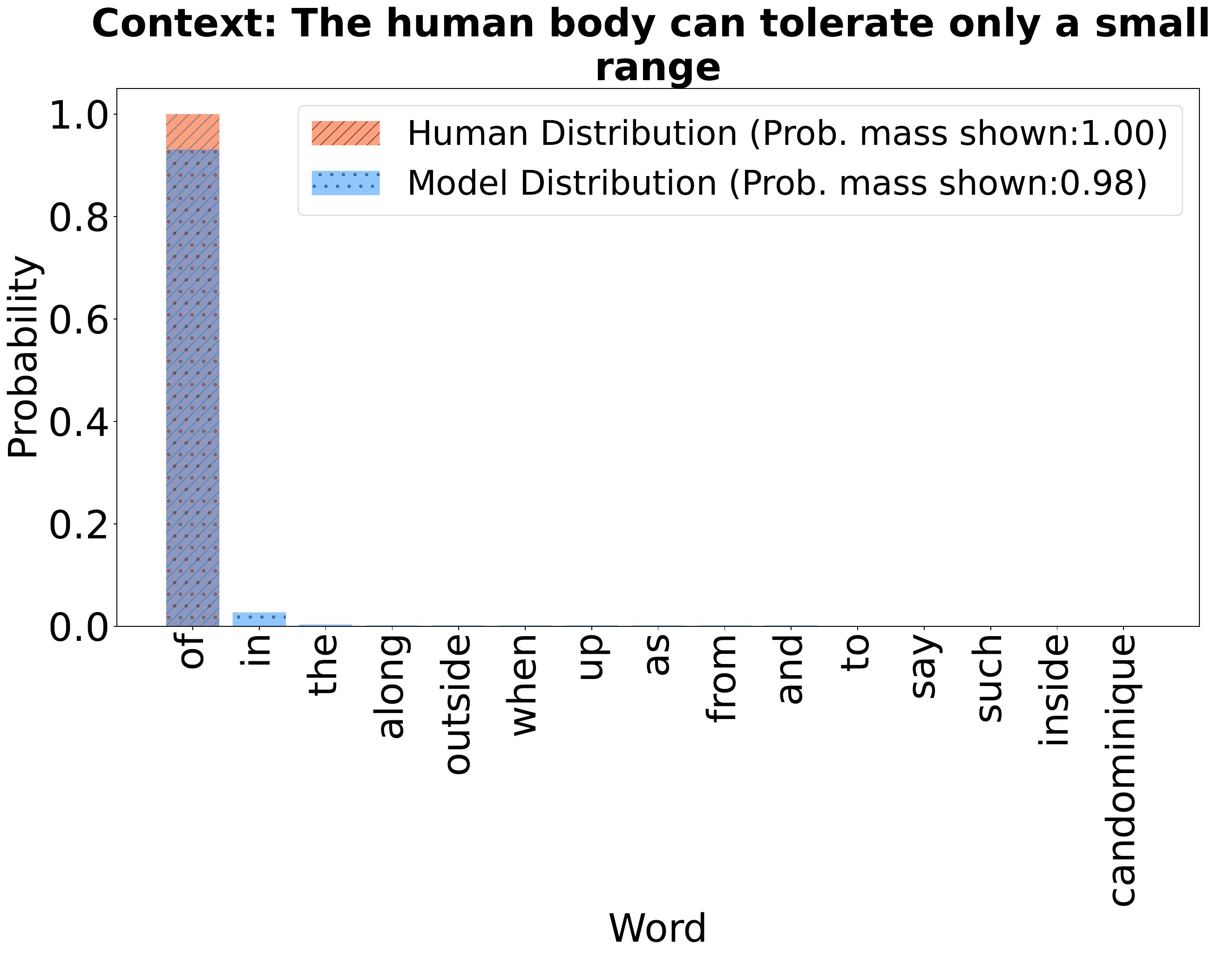} \includegraphics[width=7.8 cm]{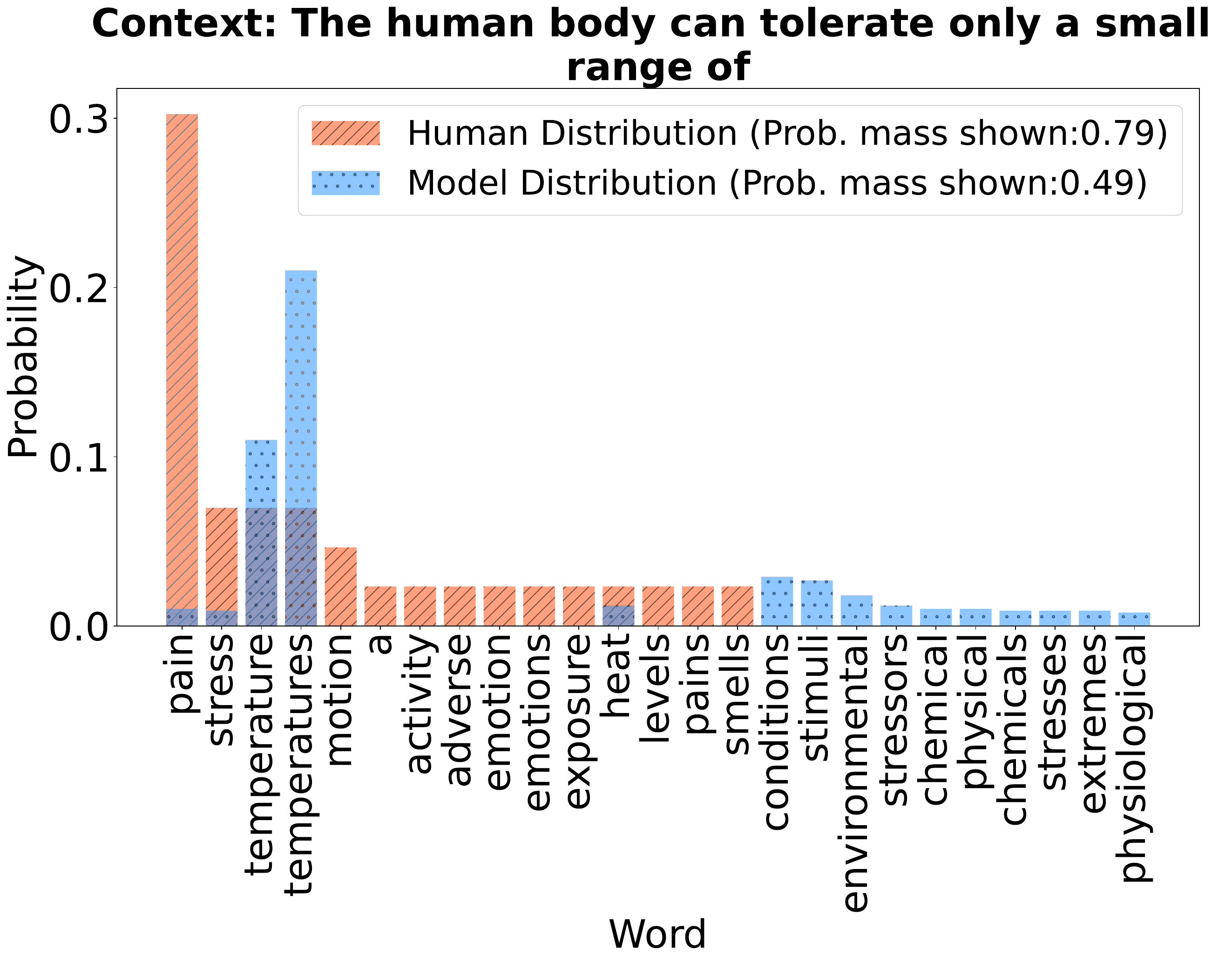}
    \includegraphics[width=7.8 cm]{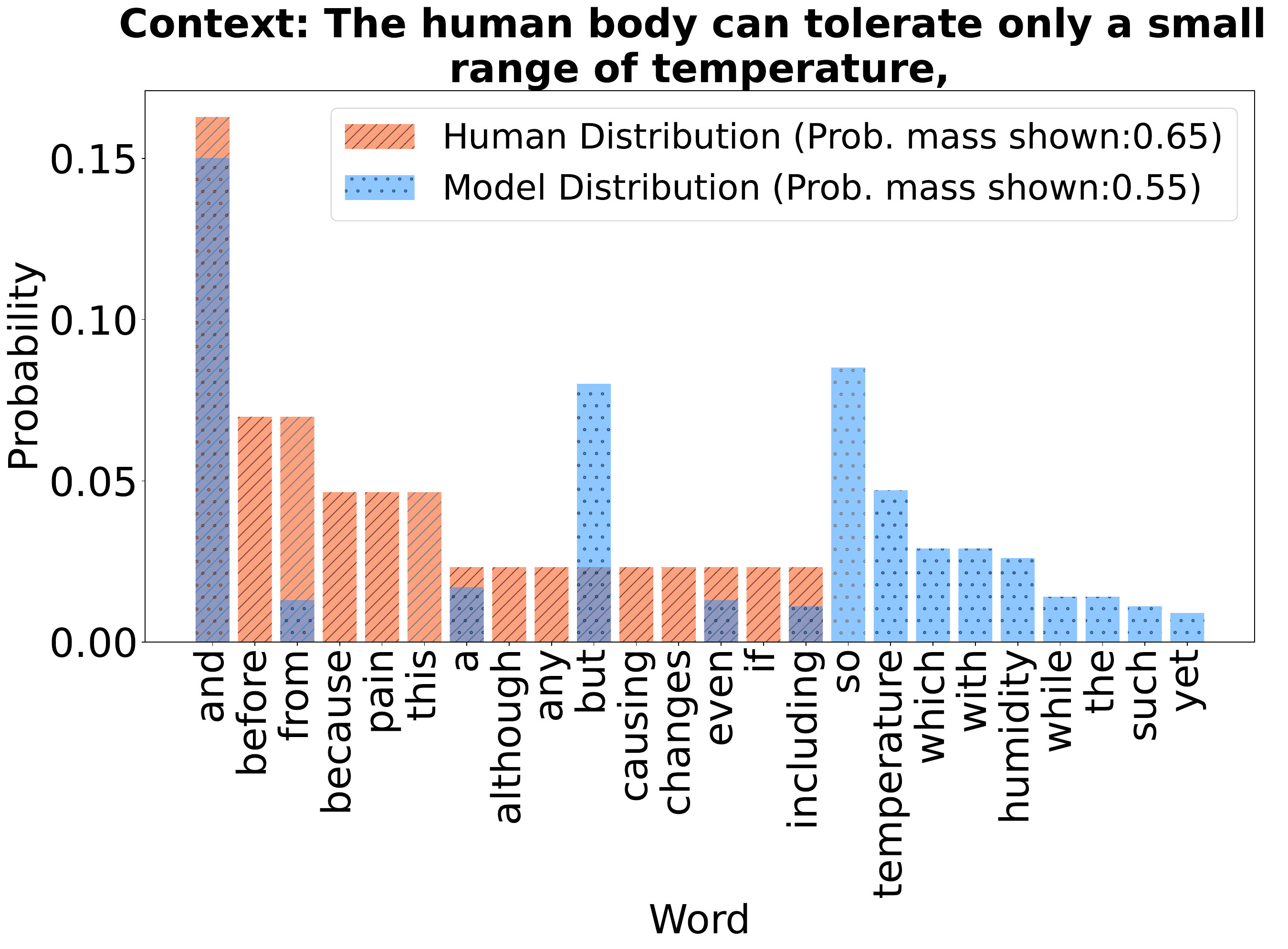}
    \includegraphics[width=7.8 cm]{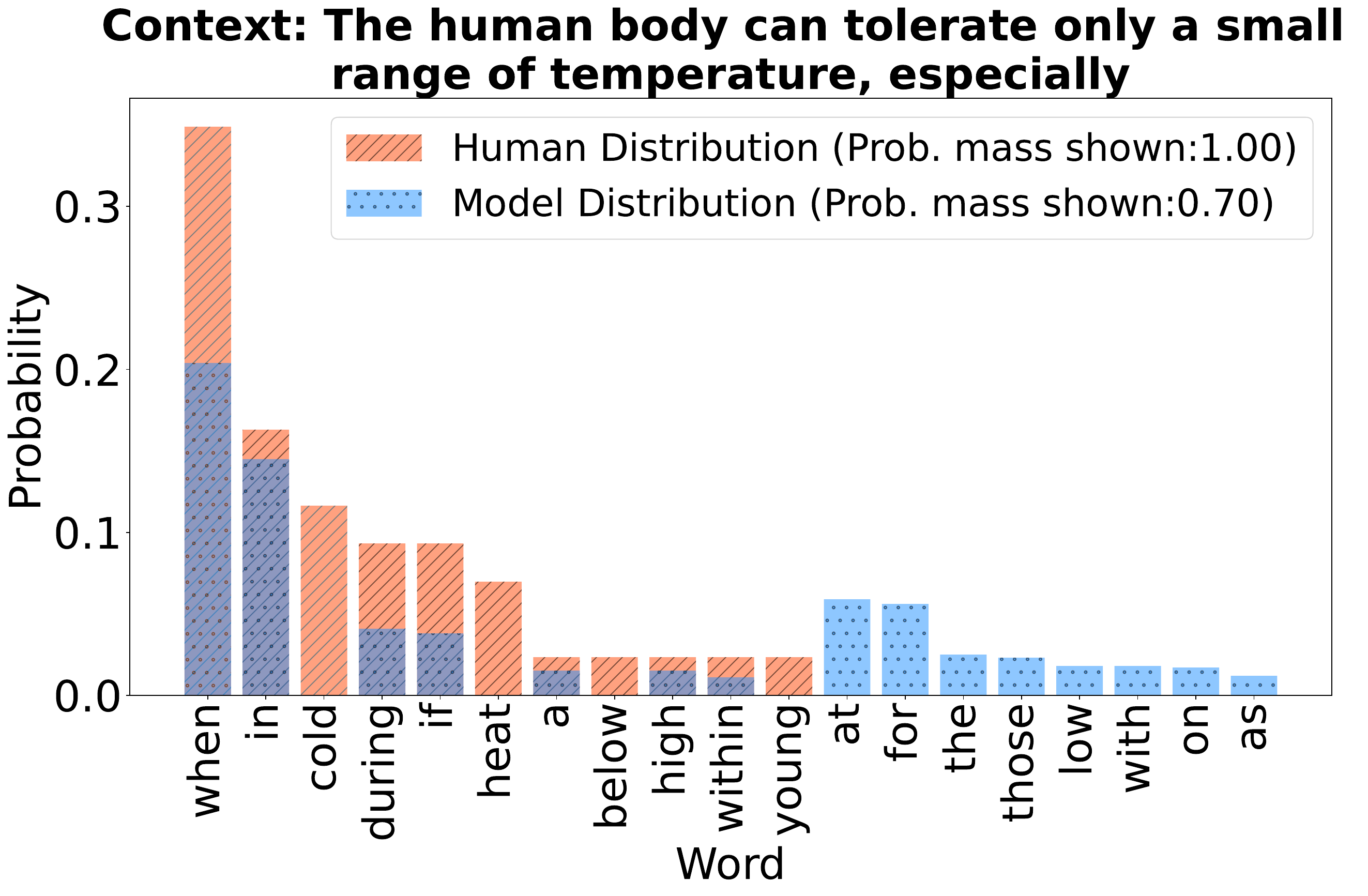}

\end{figure}

\begin{figure}
    \includegraphics[width=7.8 cm]{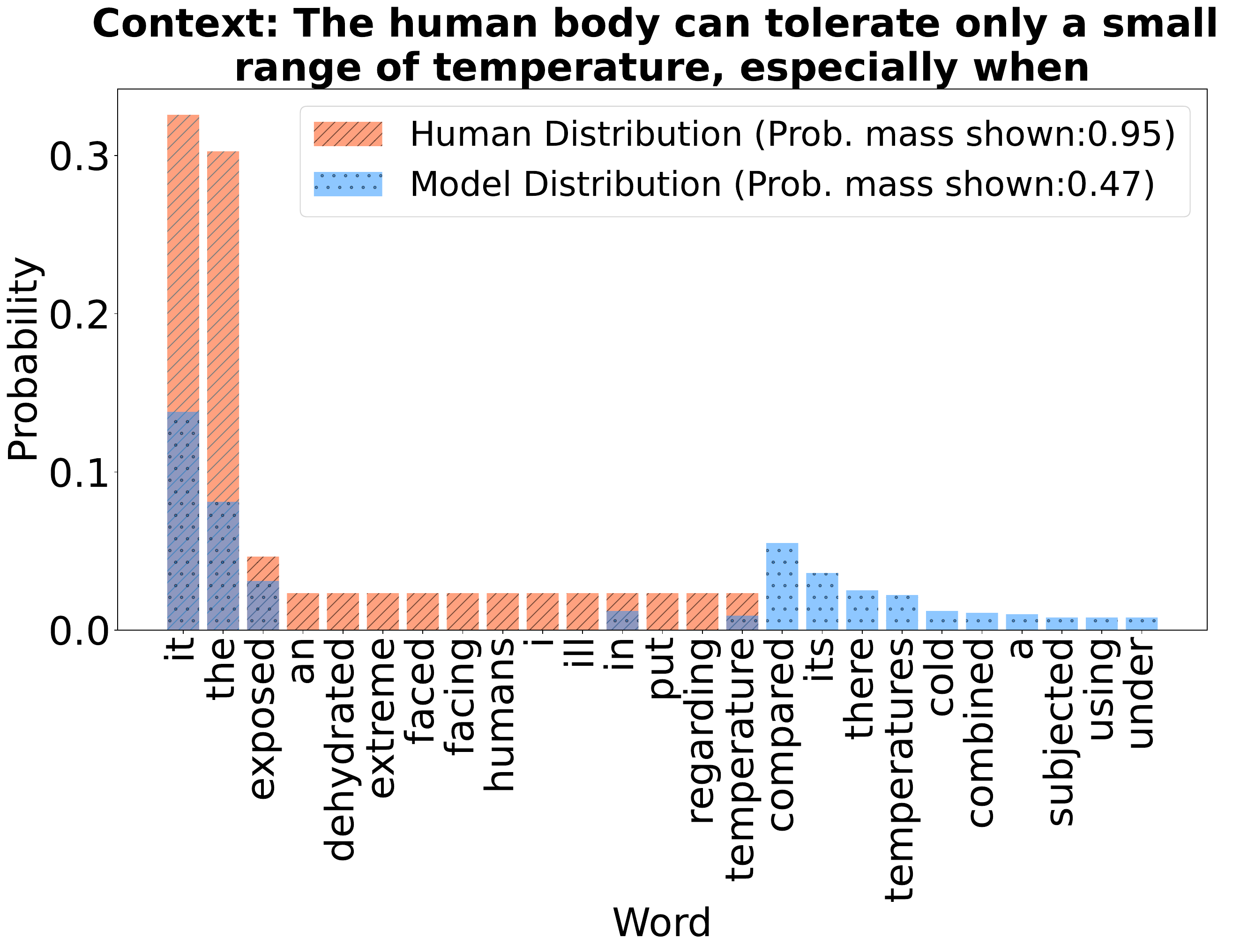} \includegraphics[width=7.8 cm]{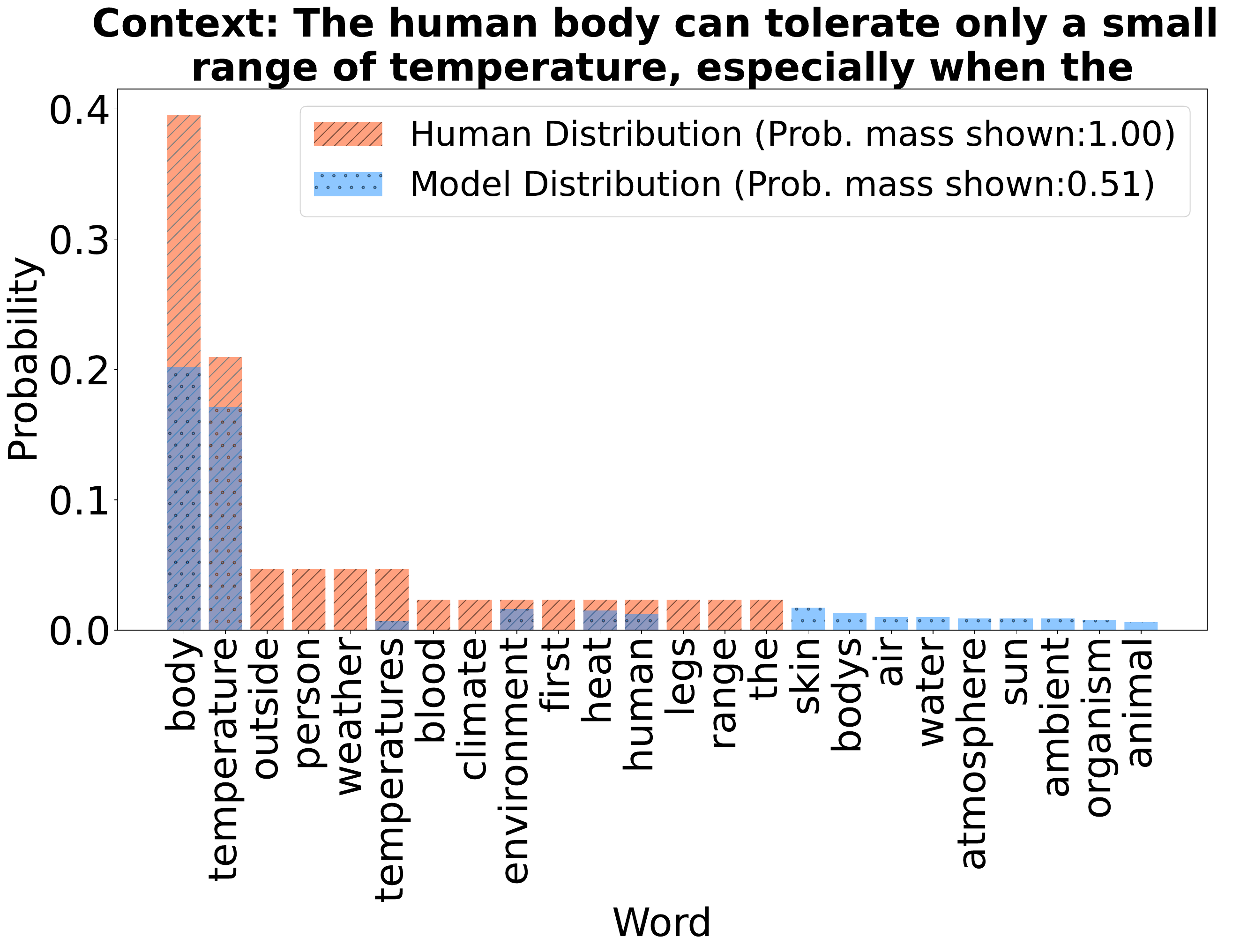}
    \includegraphics[width=7.8 cm]{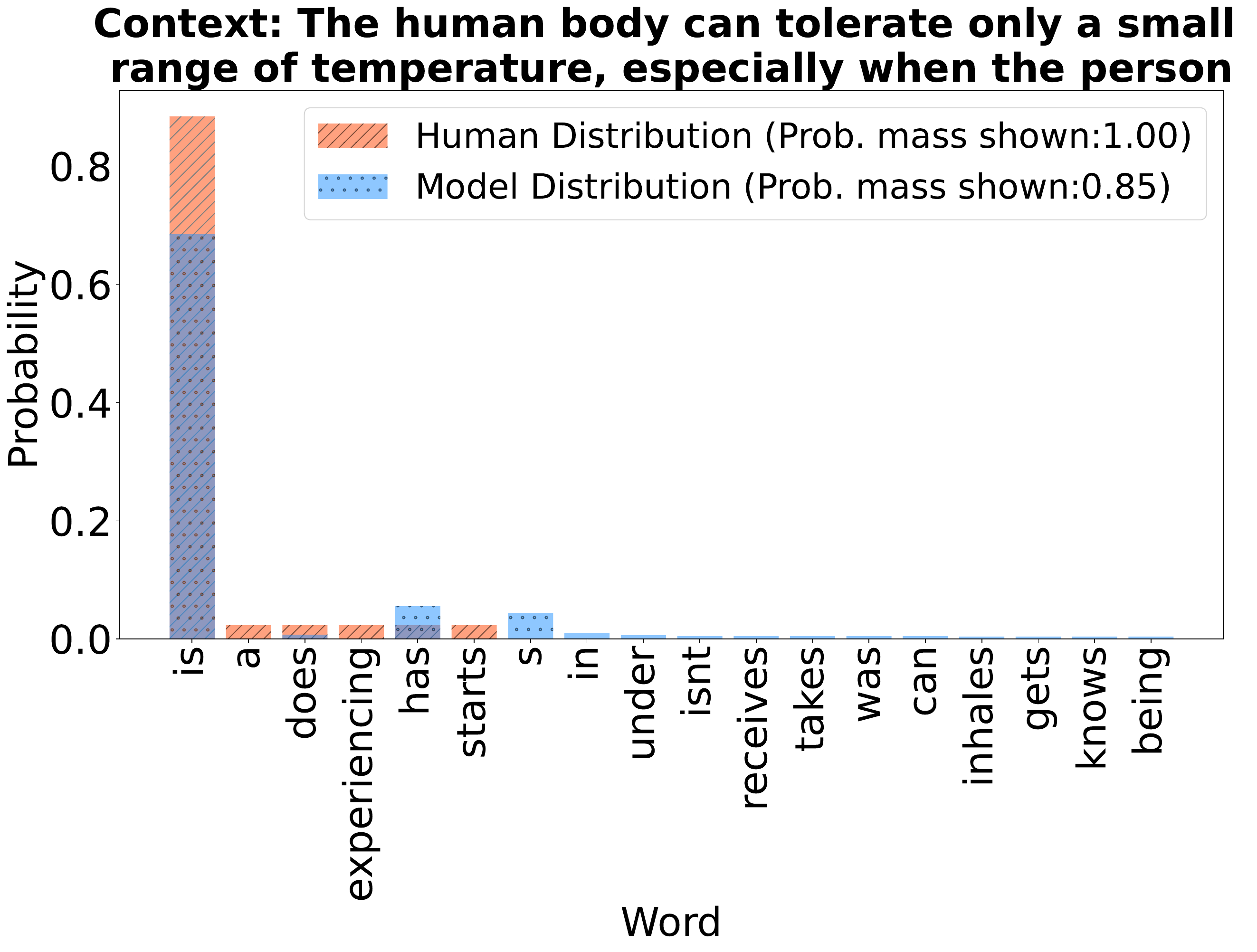}
    \includegraphics[width=7.8 cm]{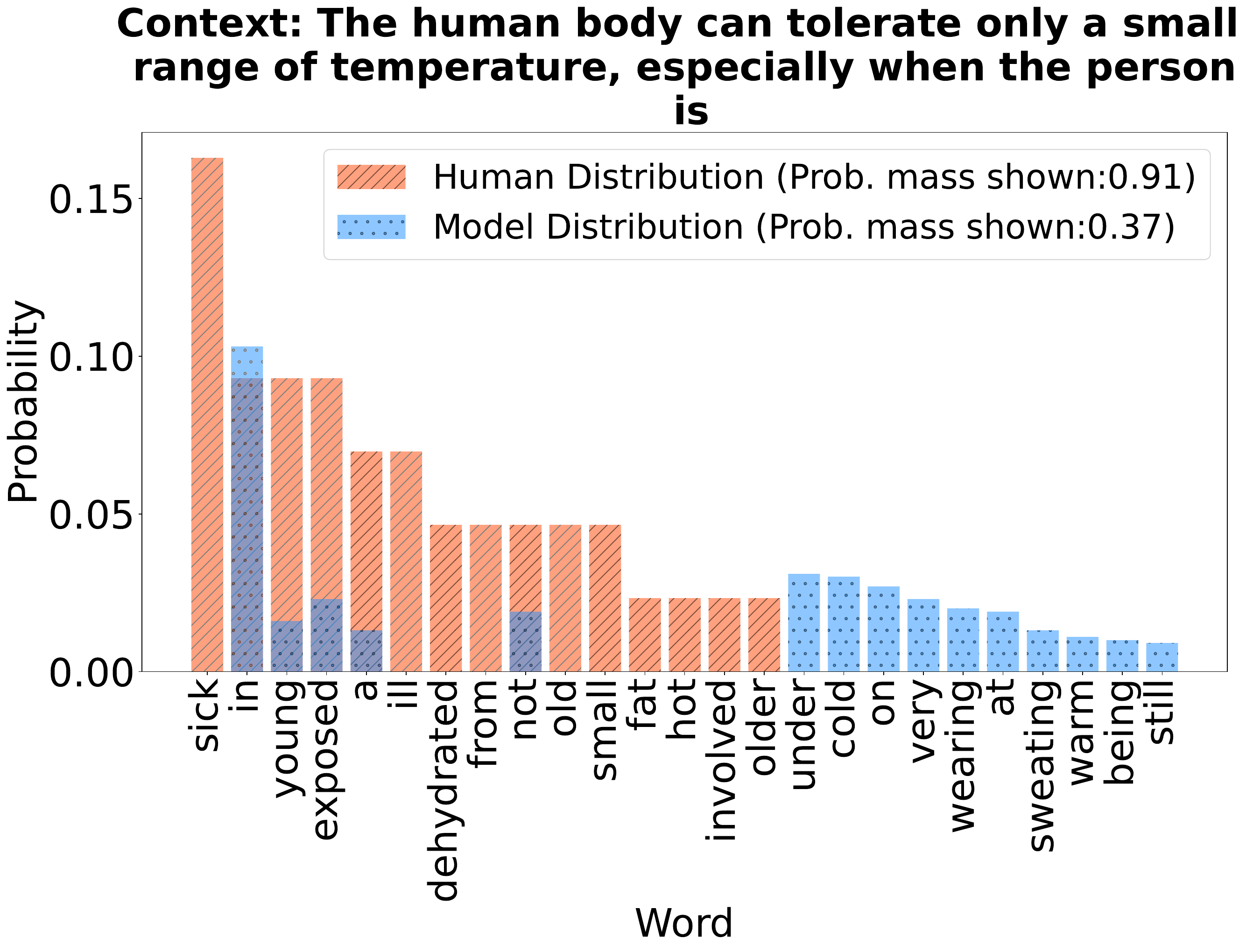}

\end{figure}

\begin{figure}
    \includegraphics[width=7.8 cm]{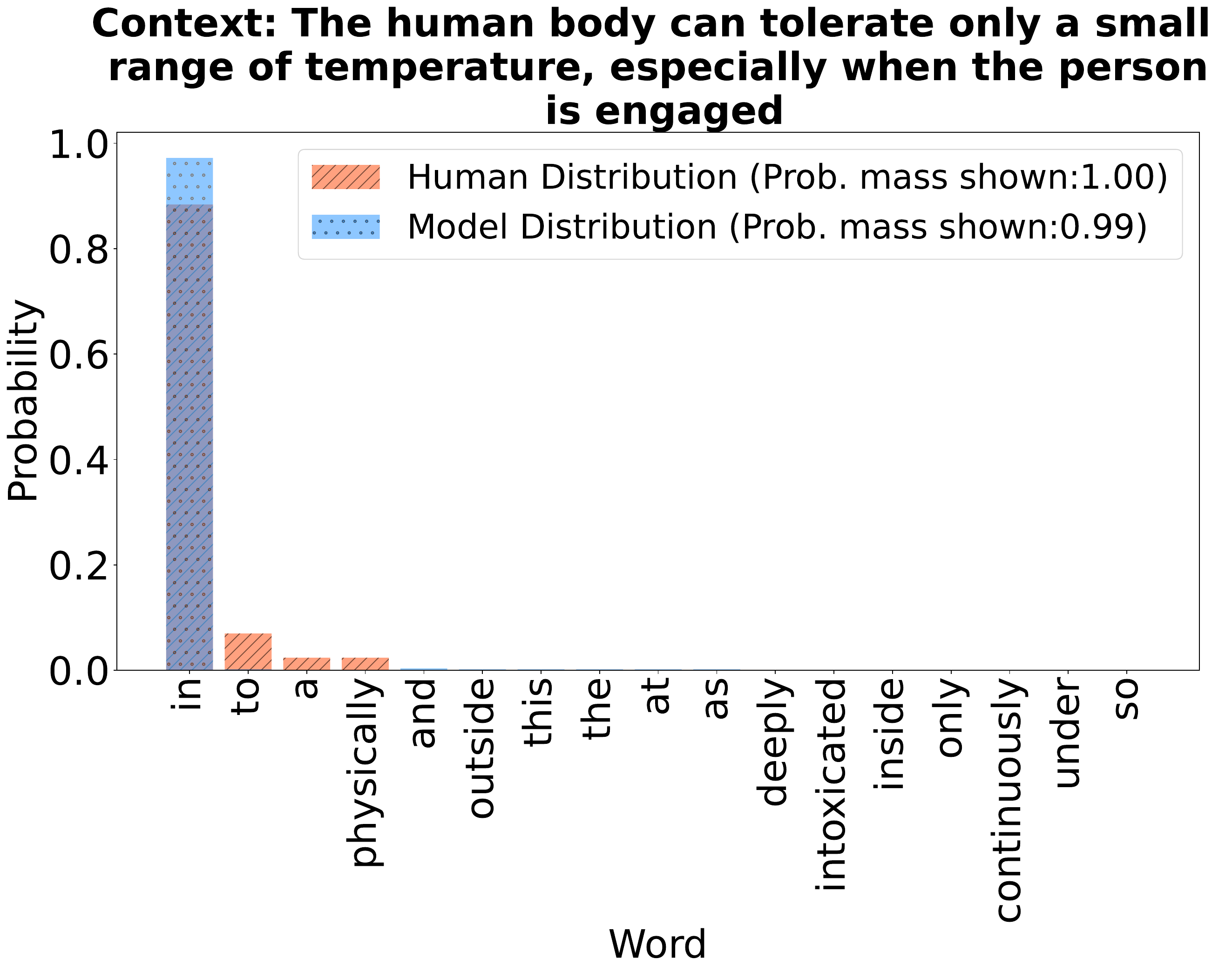} \includegraphics[width=7.8 cm]{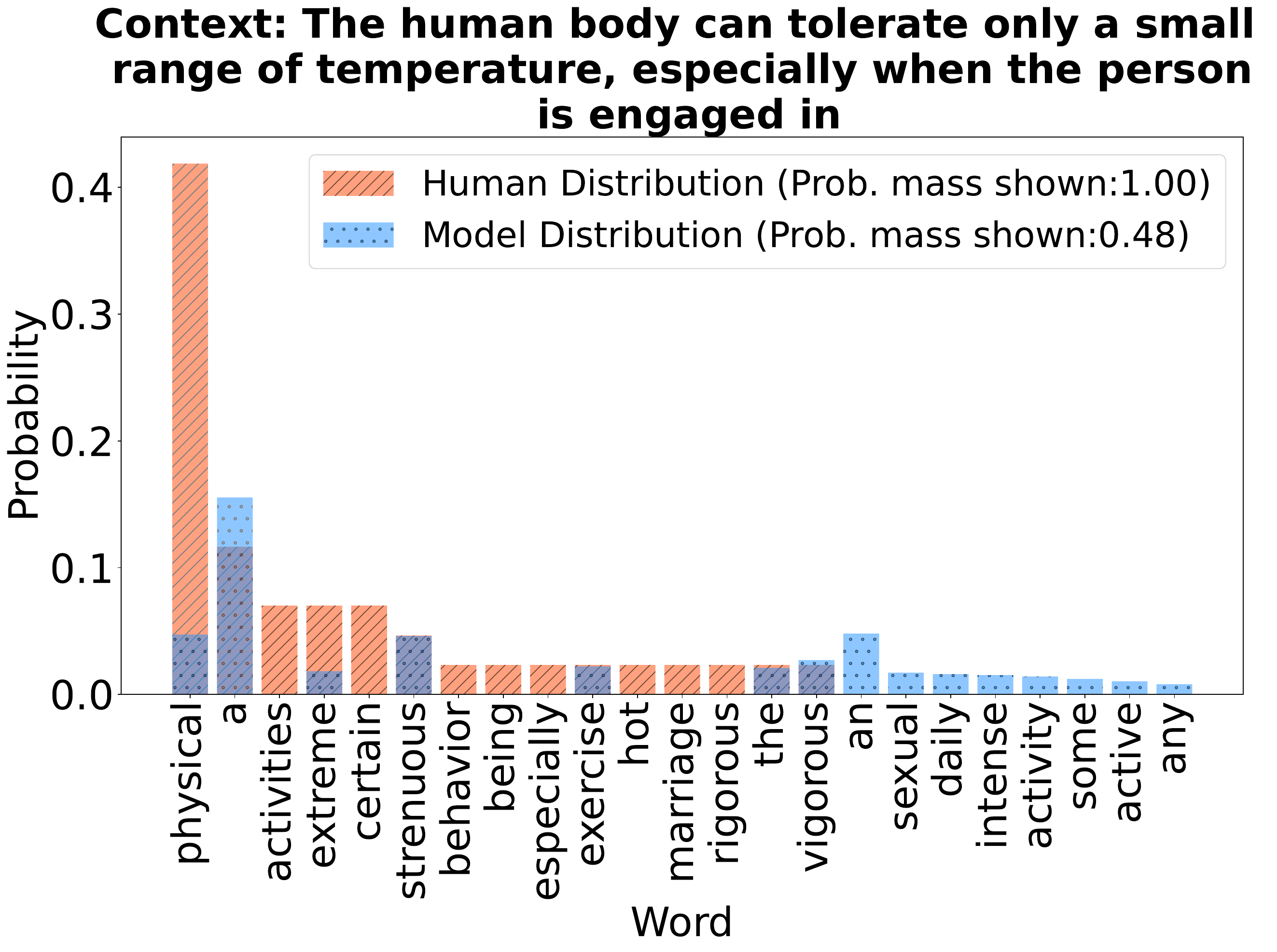}
    \includegraphics[width=7.8 cm]{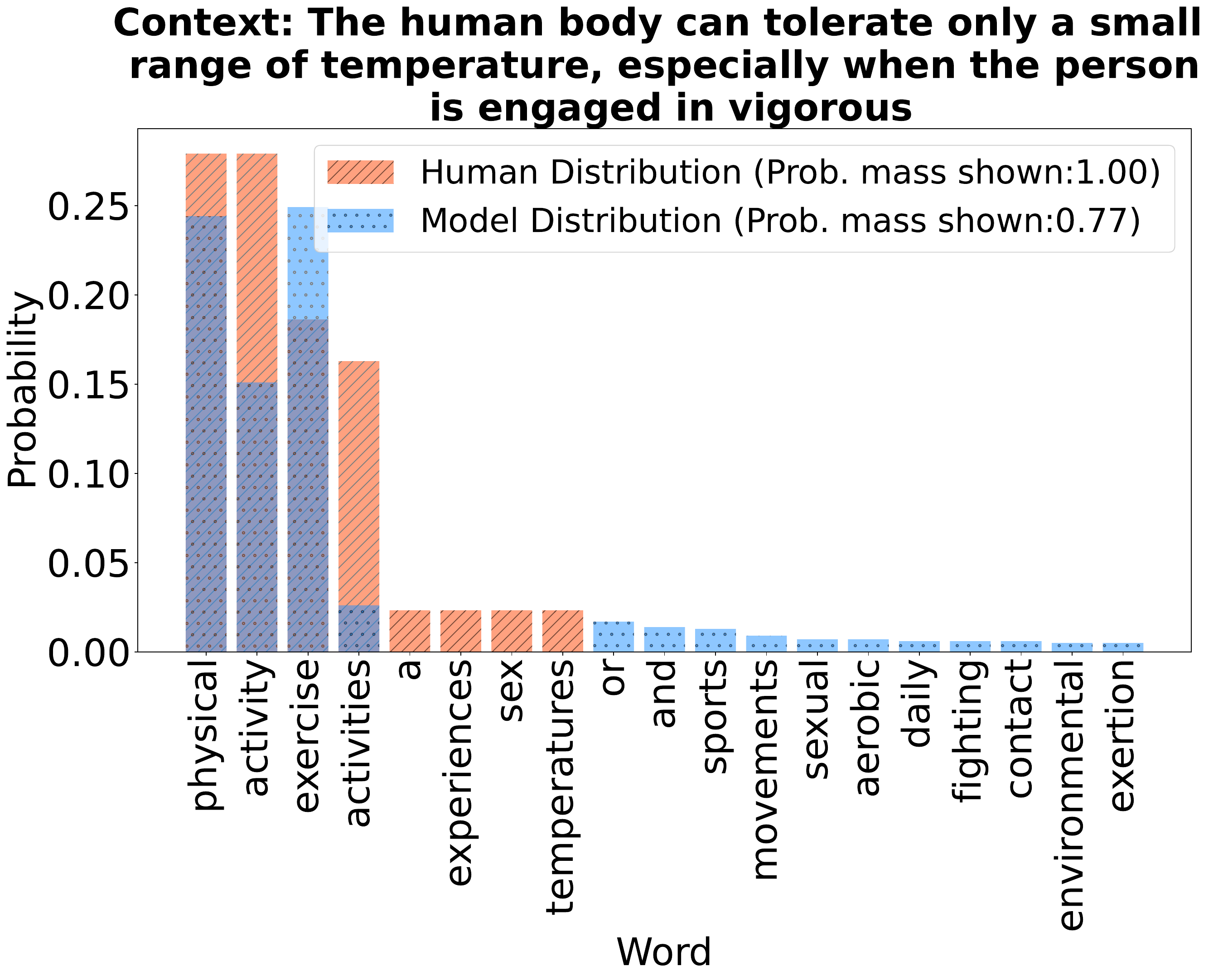}
    \includegraphics[width=7.8 cm]{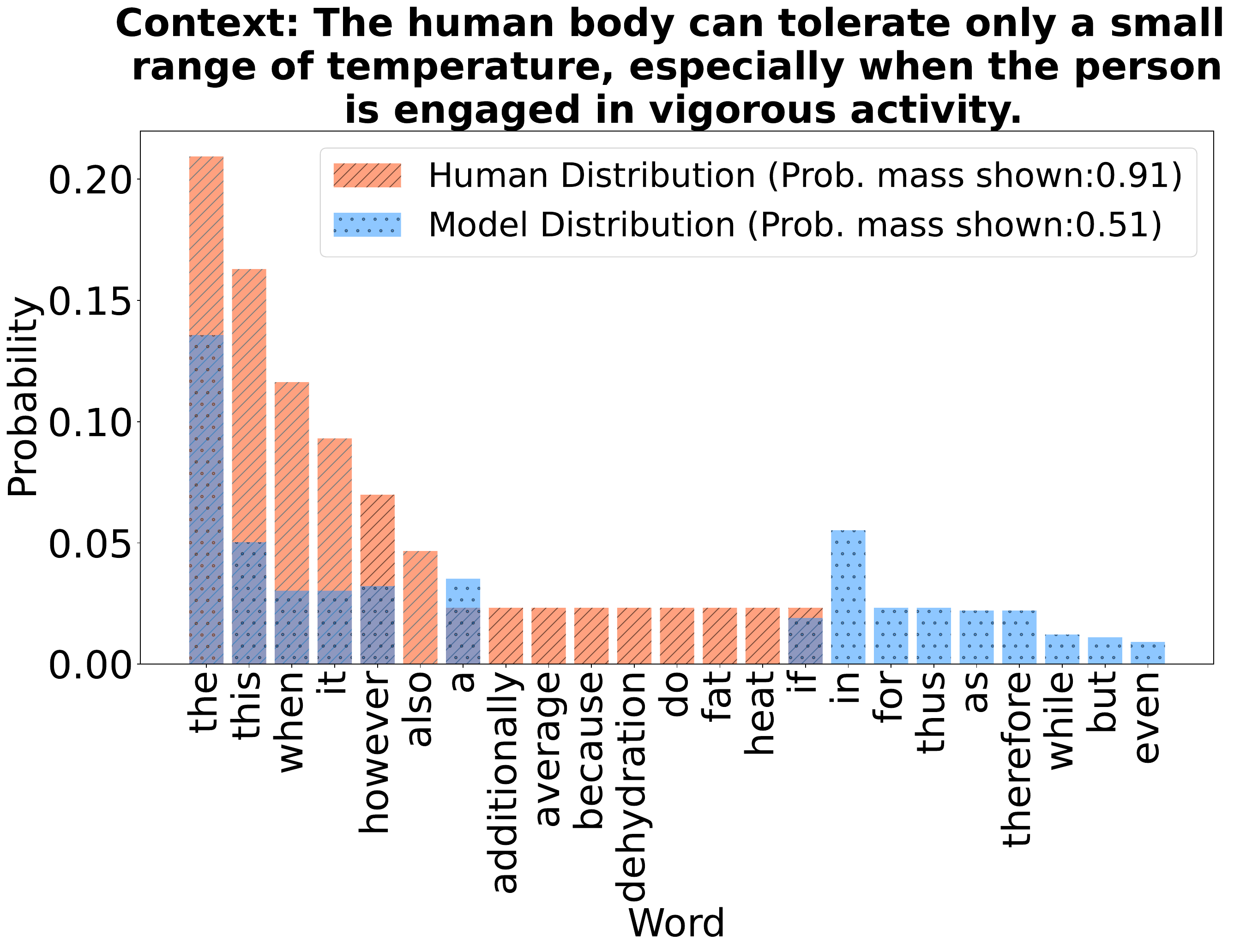}
\end{figure}

\begin{figure}
    \includegraphics[width=7.8 cm]{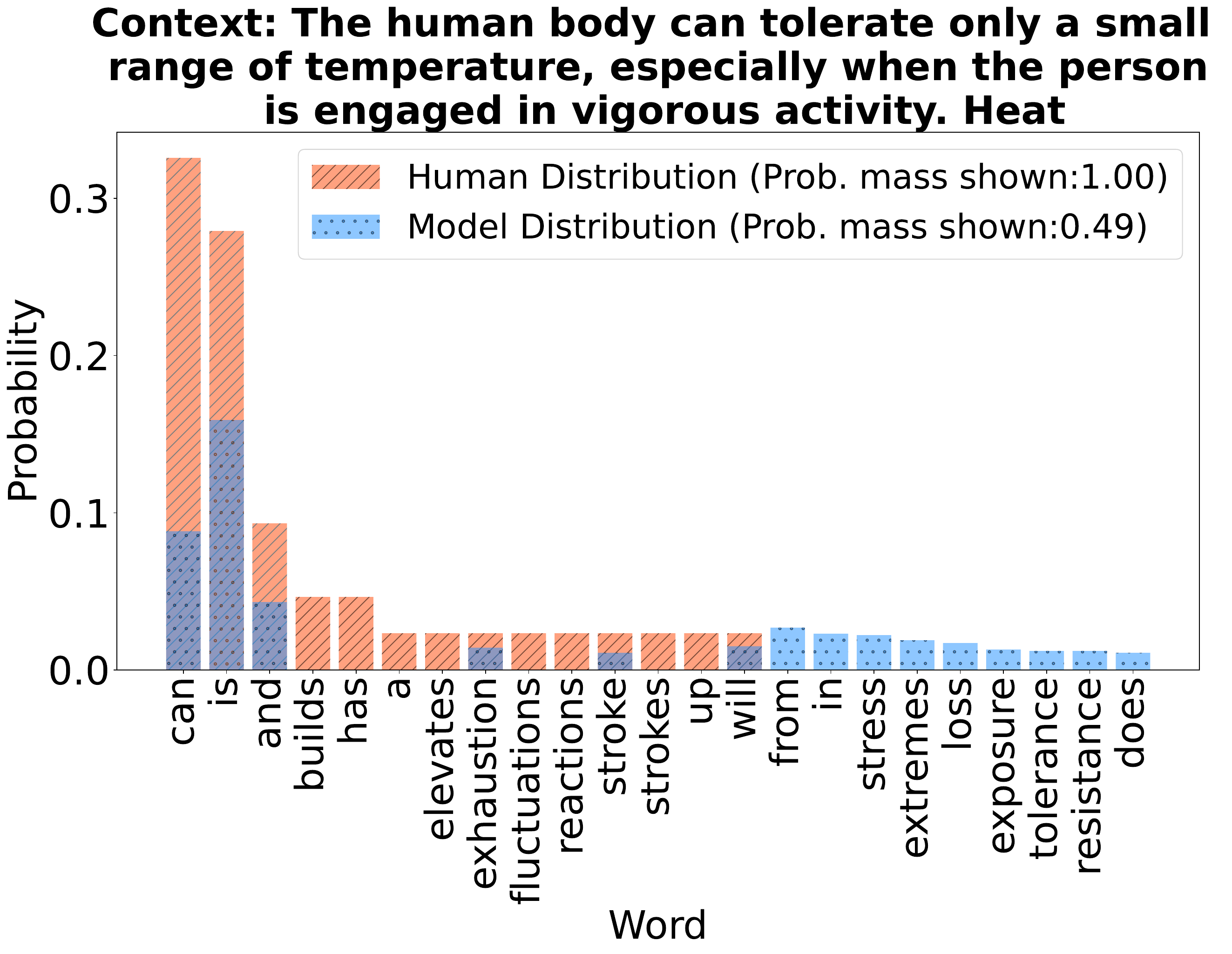} \includegraphics[width=7.8 cm]{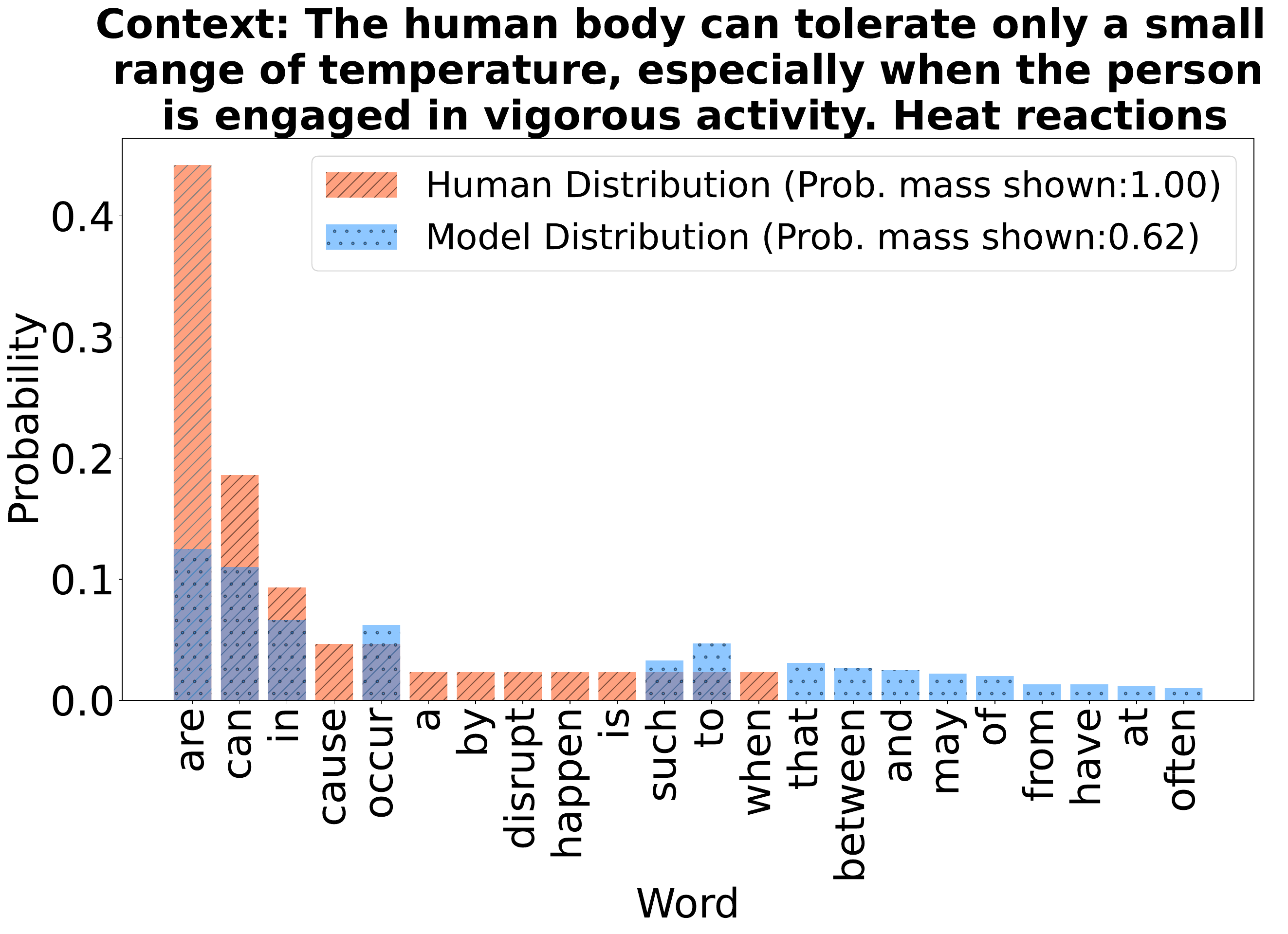}
    \includegraphics[width=7.8 cm]{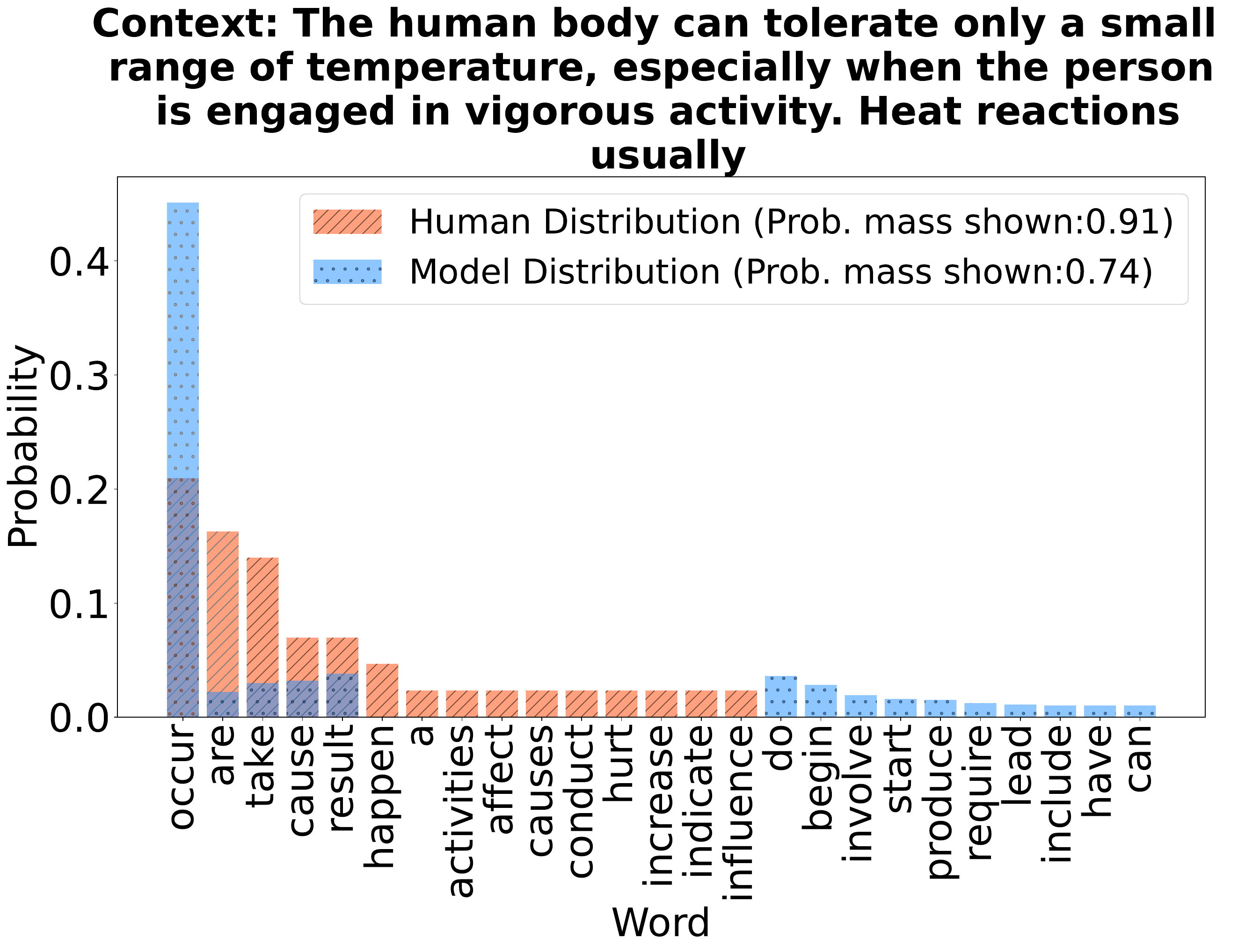}
    \includegraphics[width=7.8 cm]{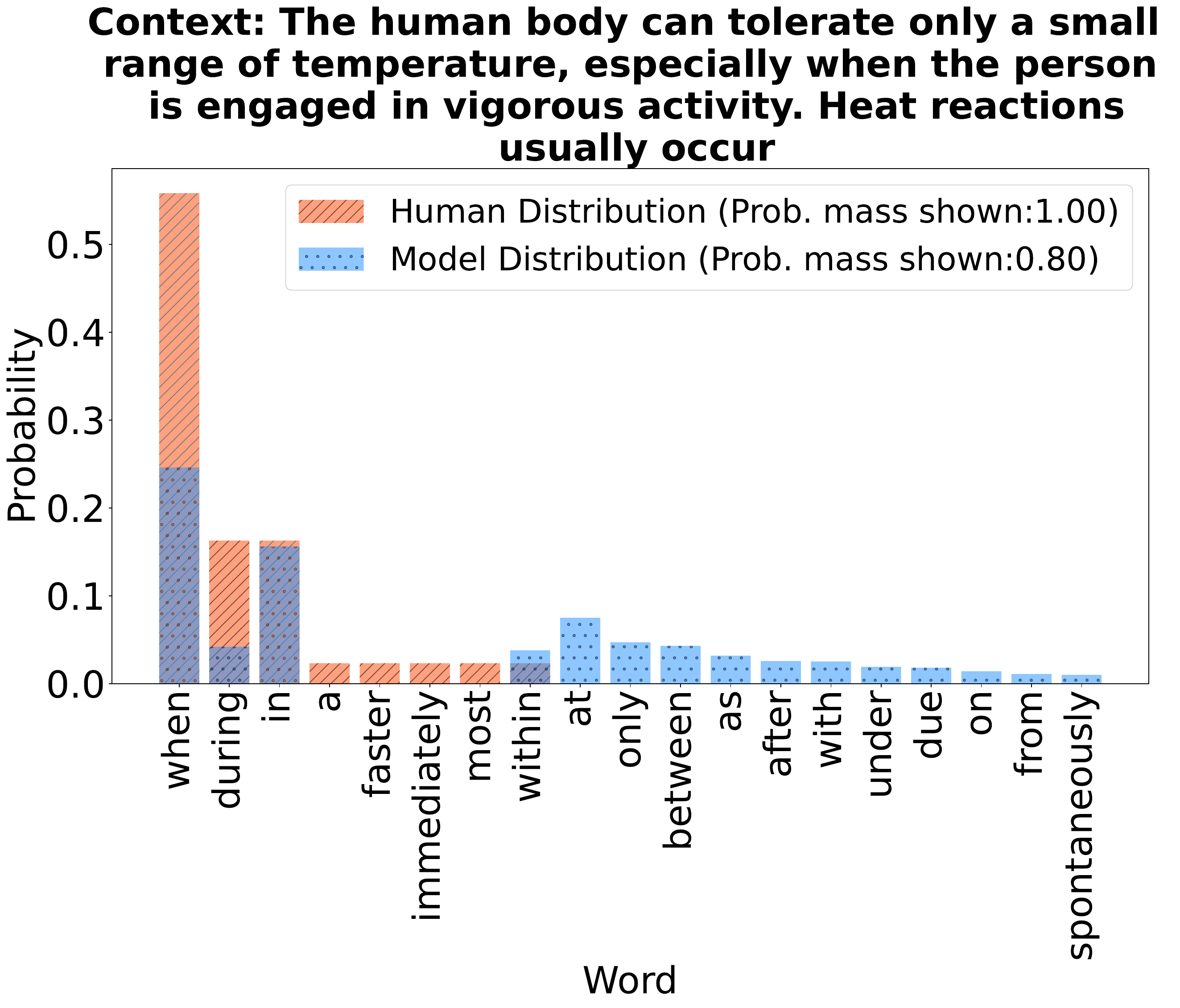}

\end{figure}

\begin{figure}
    \includegraphics[width=7.6 cm]{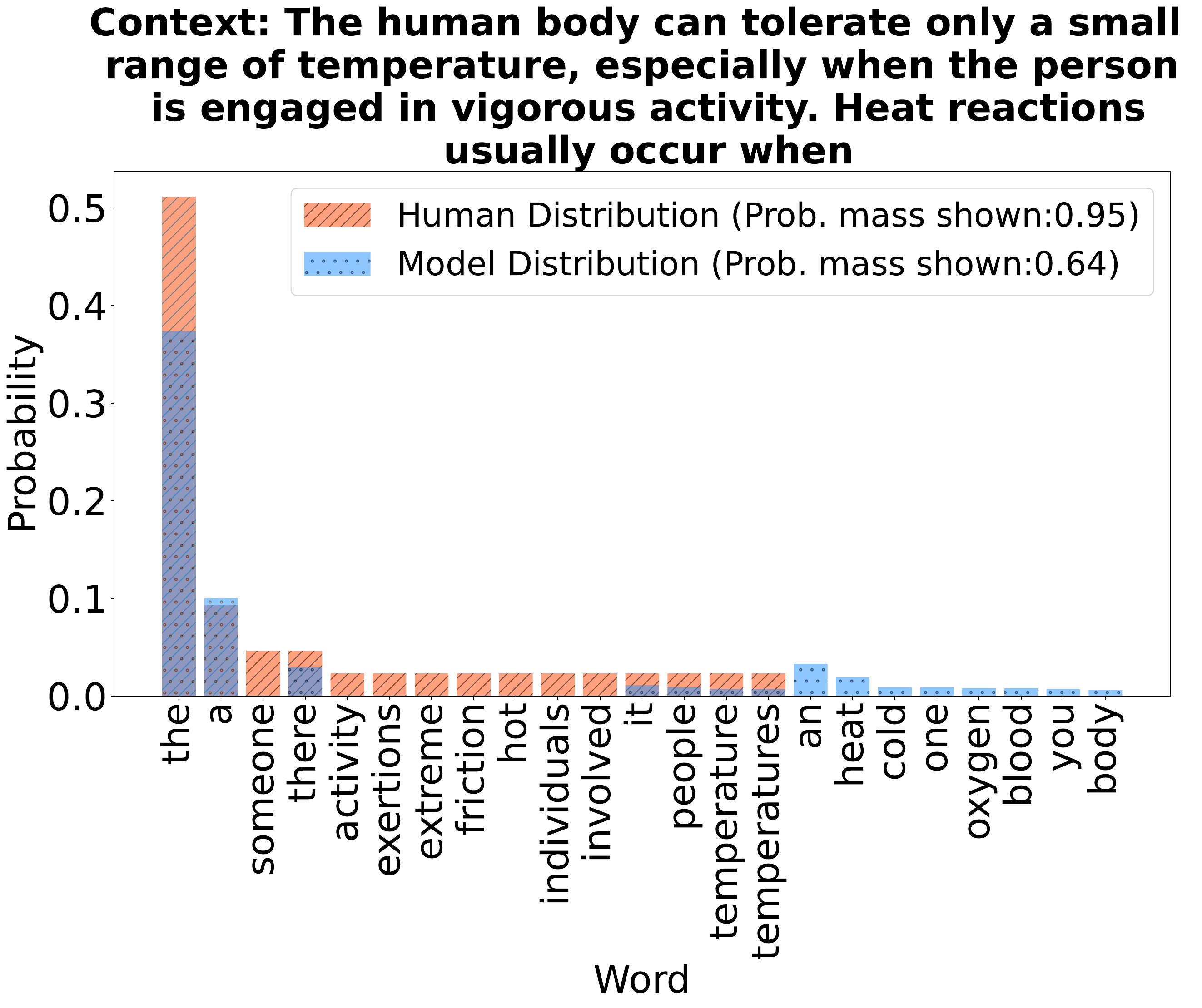} \includegraphics[width=7.6 cm]{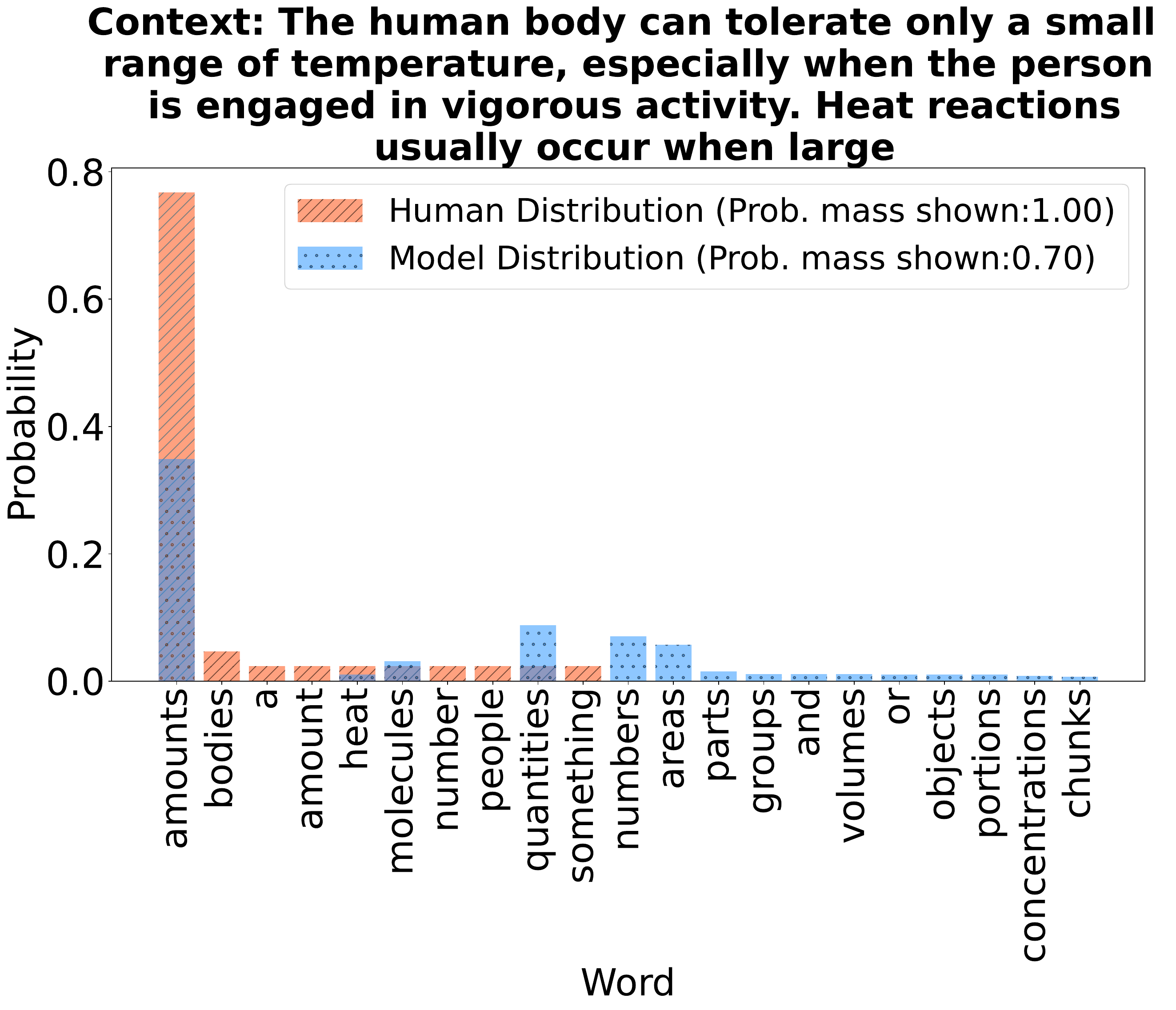}
    \includegraphics[width=7.6 cm]{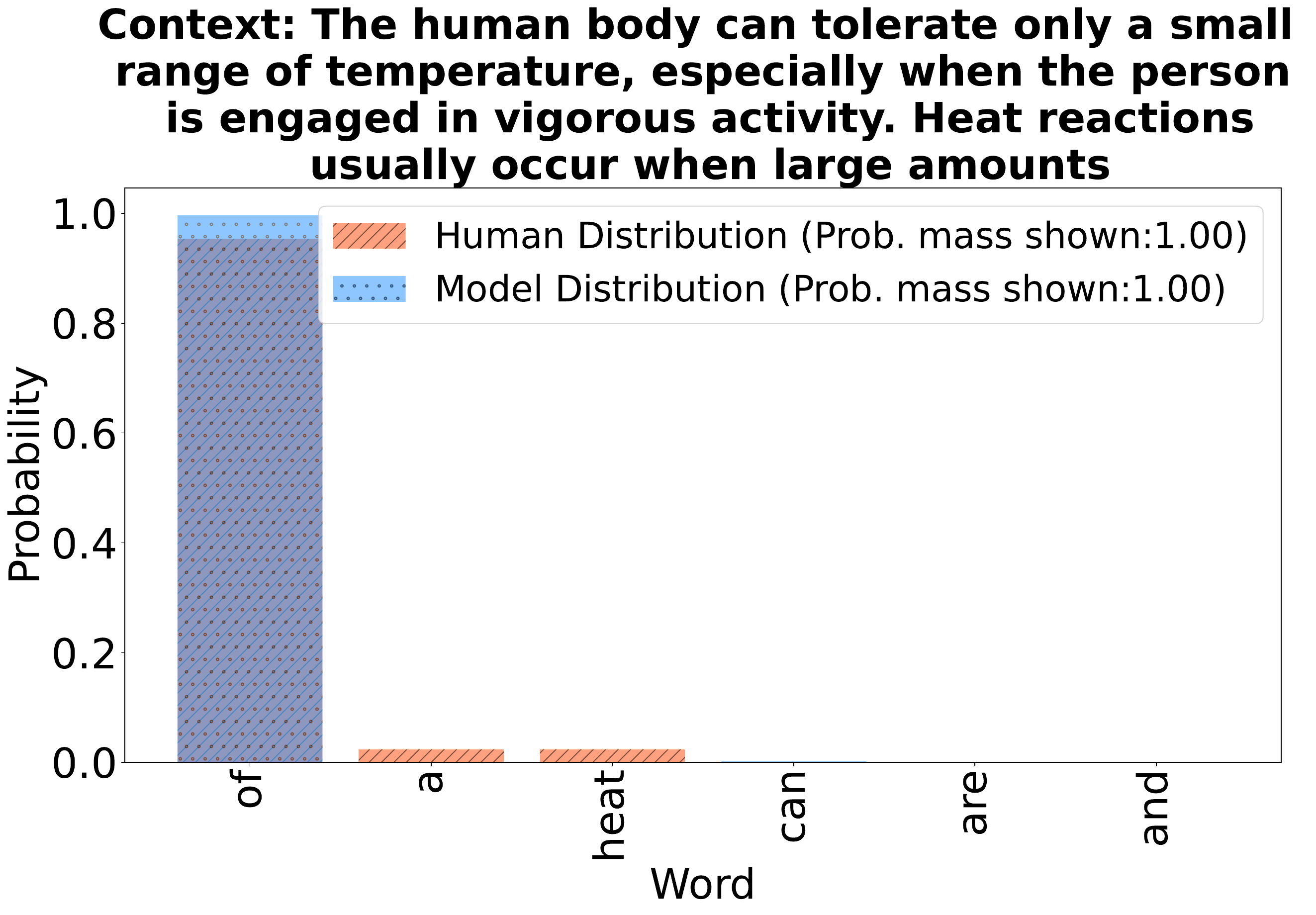}
    \includegraphics[width=7.6 cm]{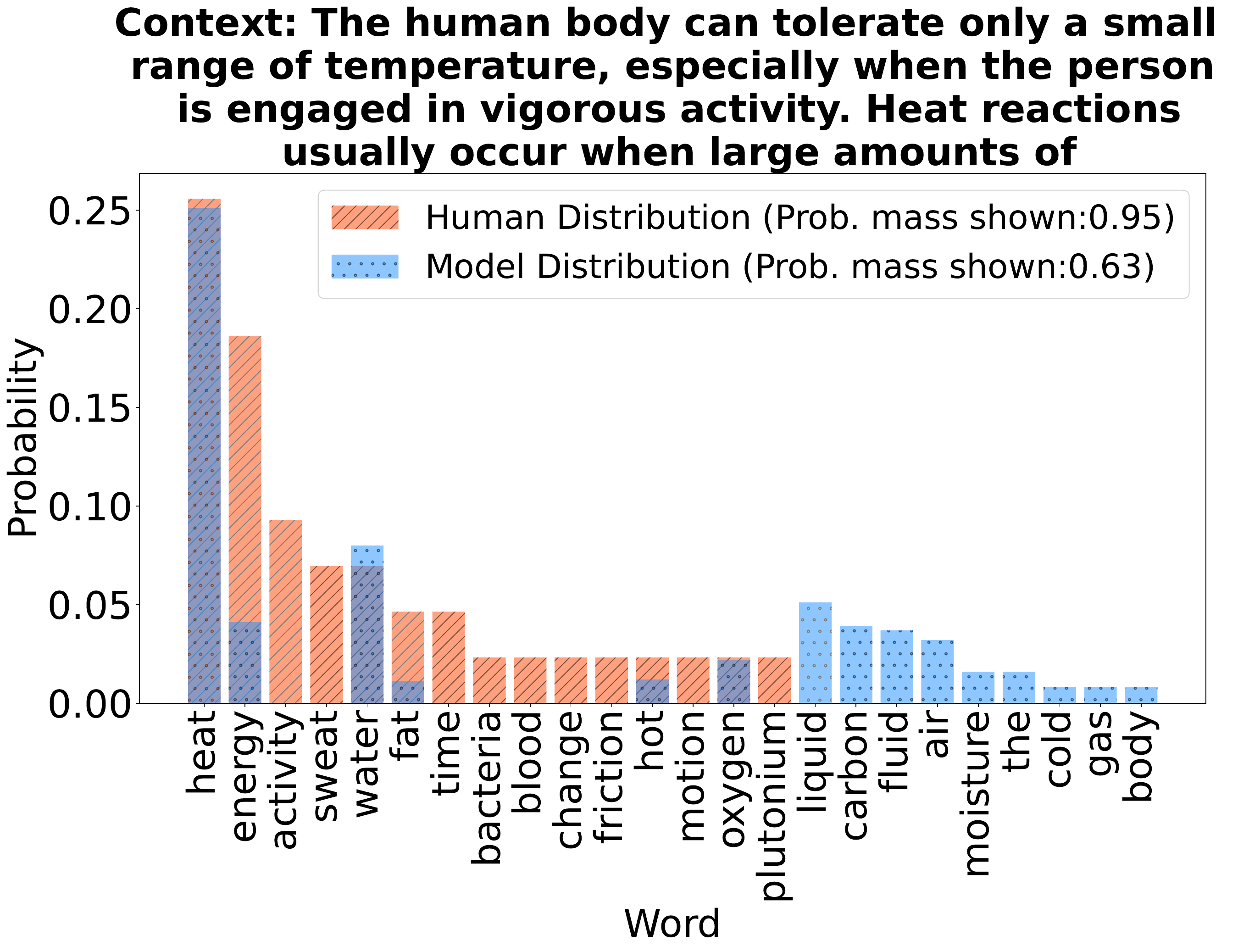}
\end{figure}

\begin{figure}
    \includegraphics[width=7.6 cm]{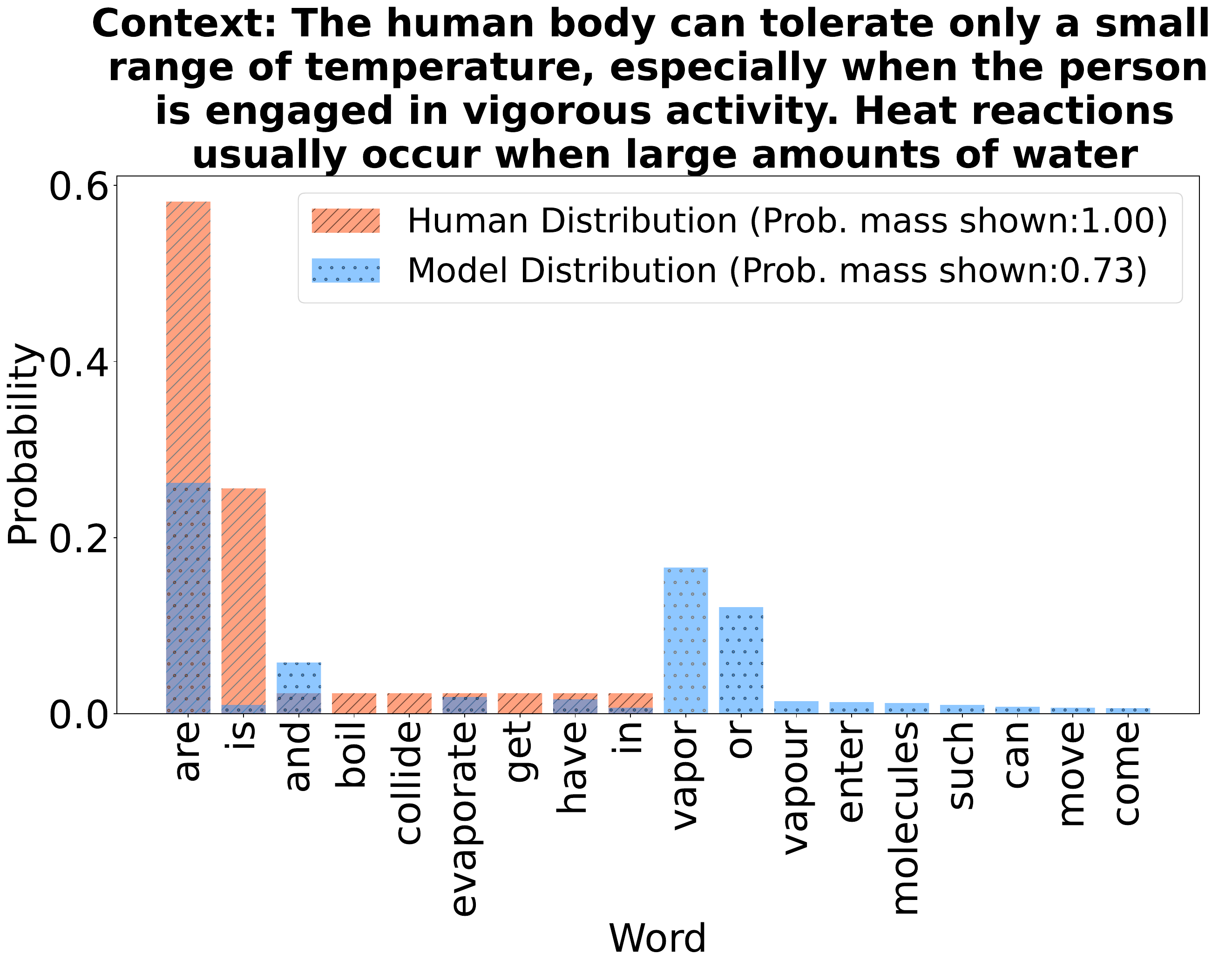} \includegraphics[width=7.6 cm]{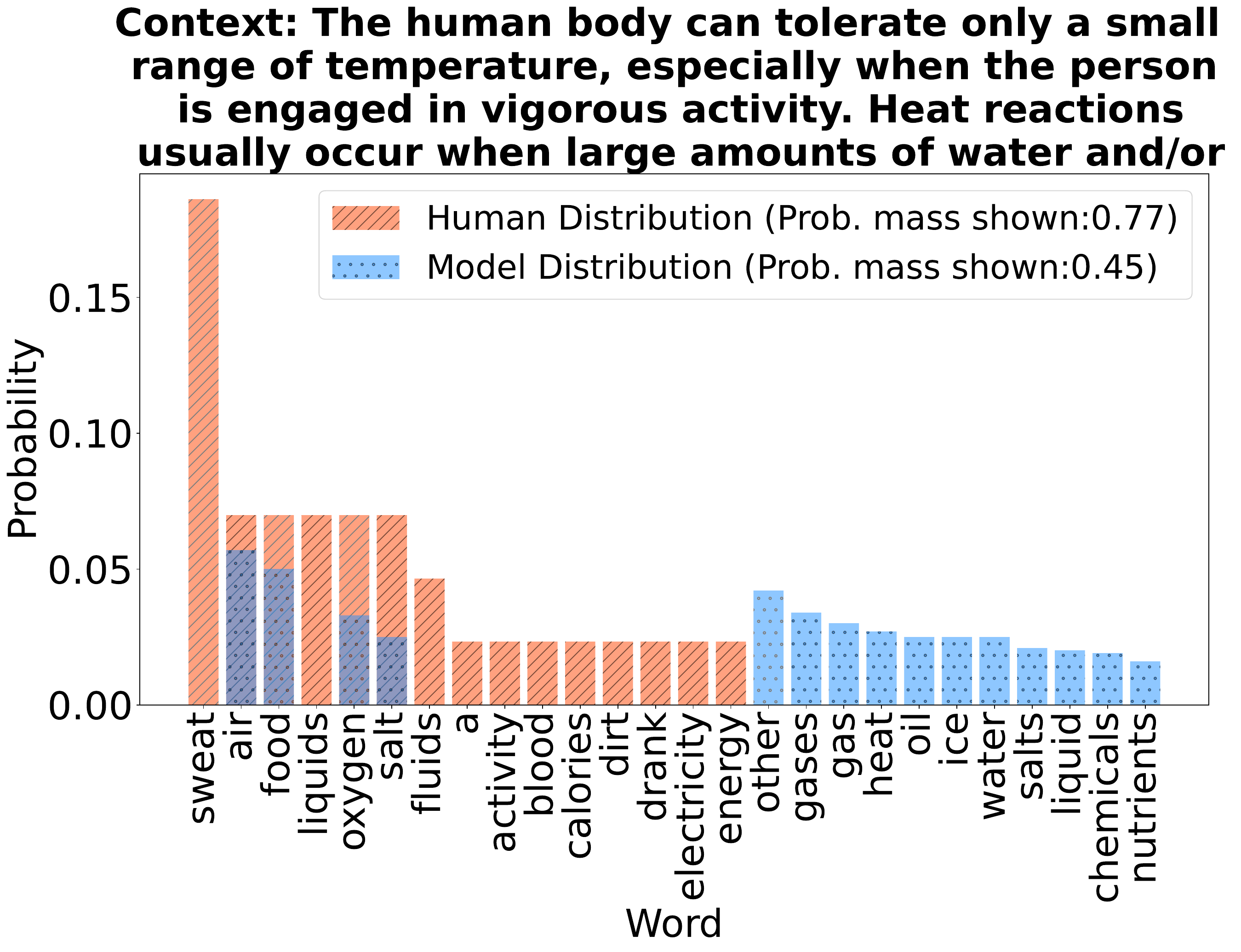}
    \includegraphics[width=7.6 cm]{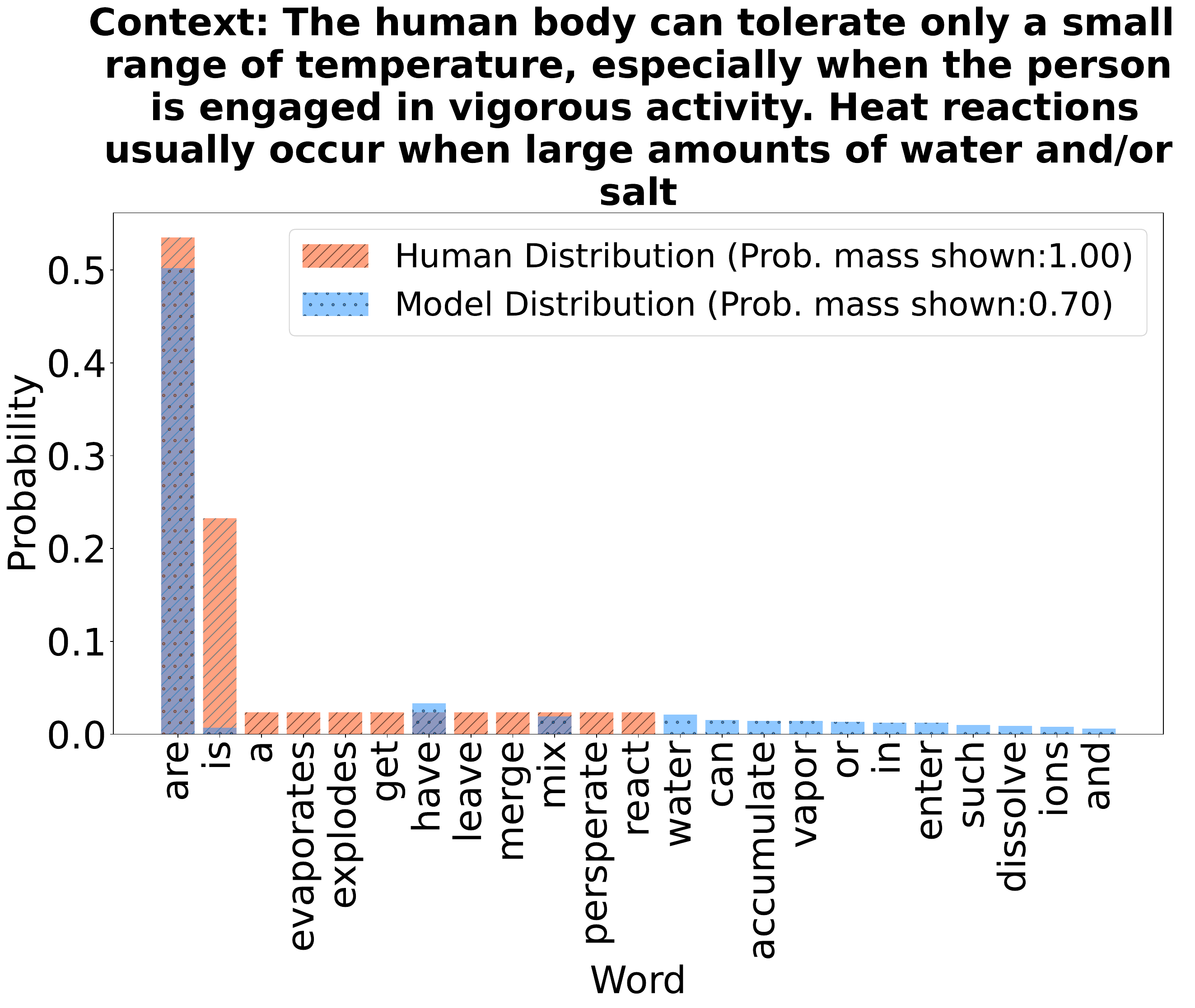}
    \includegraphics[width=7.6 cm]{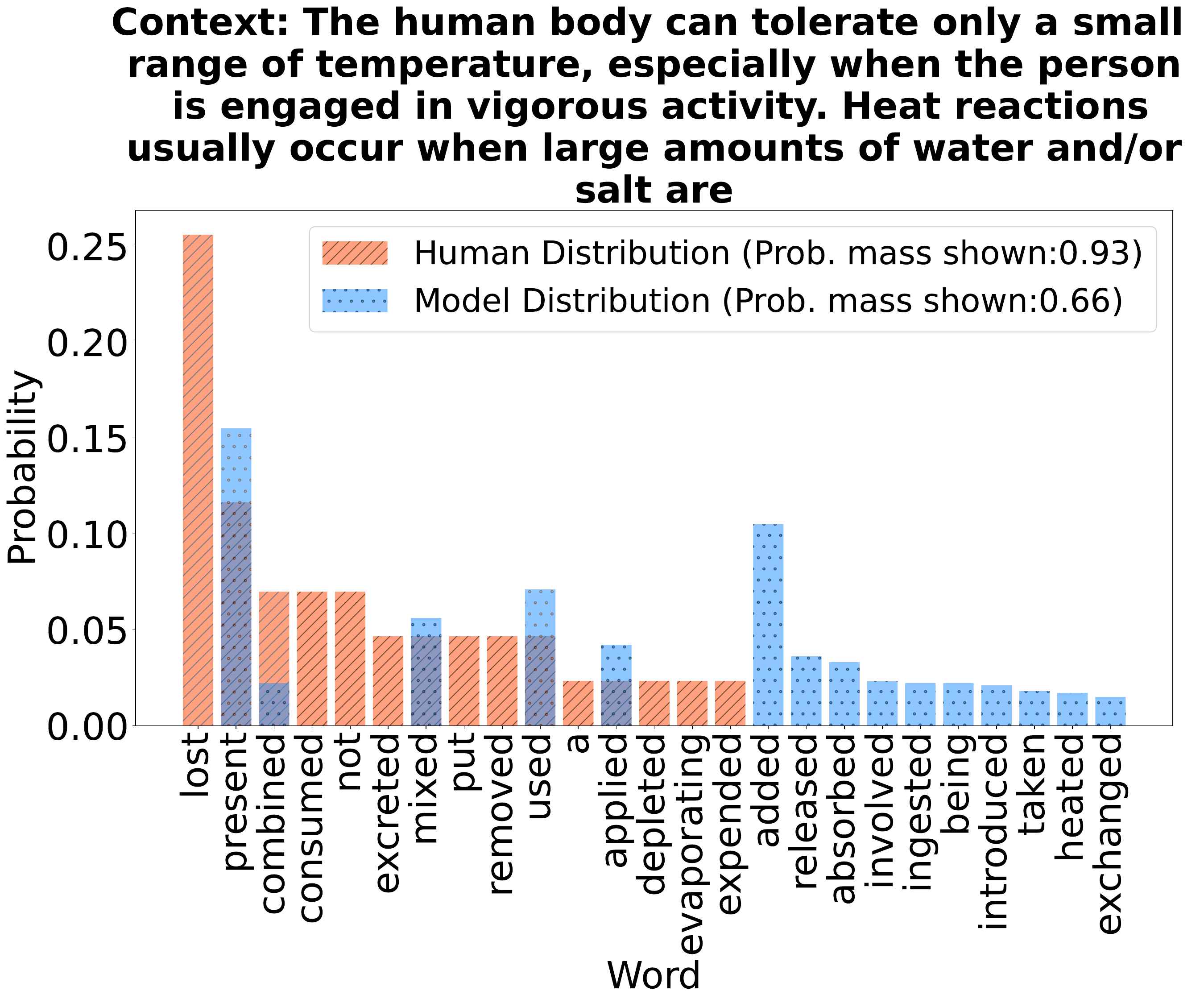}
\end{figure}

\begin{figure}
    \includegraphics[width=7.8 cm]{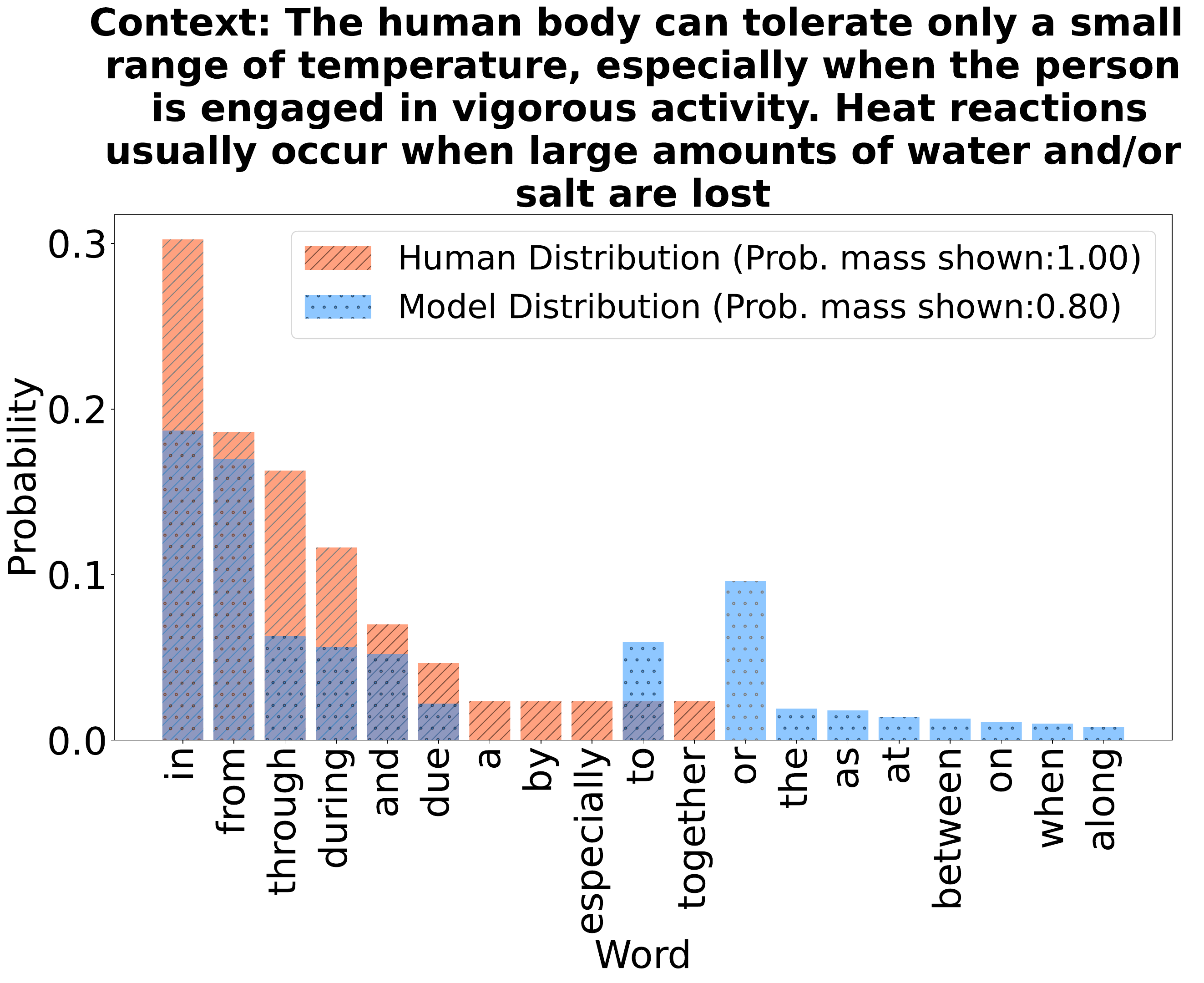} \includegraphics[width=7.8 cm]{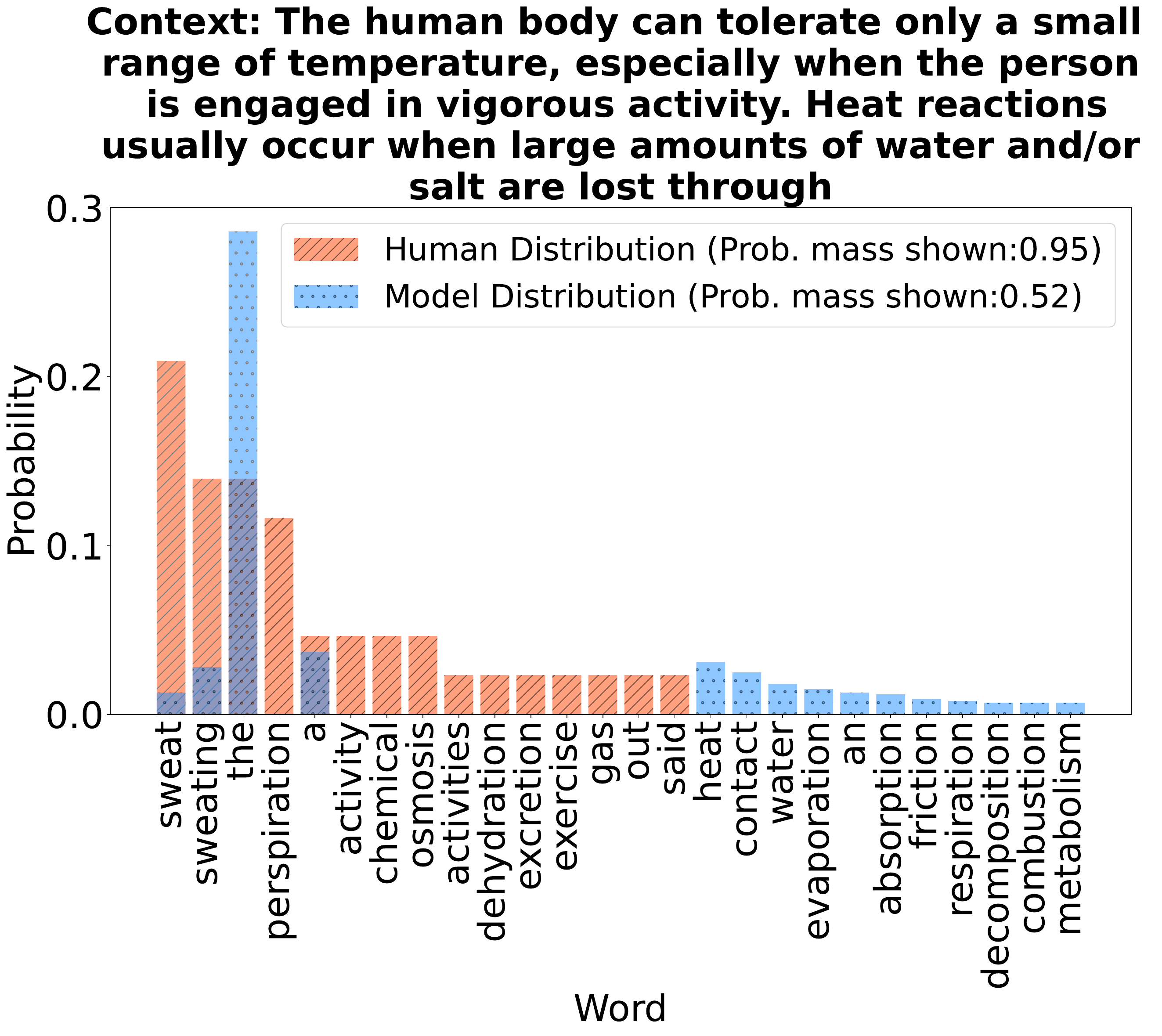}
    \includegraphics[width=7.8 cm]{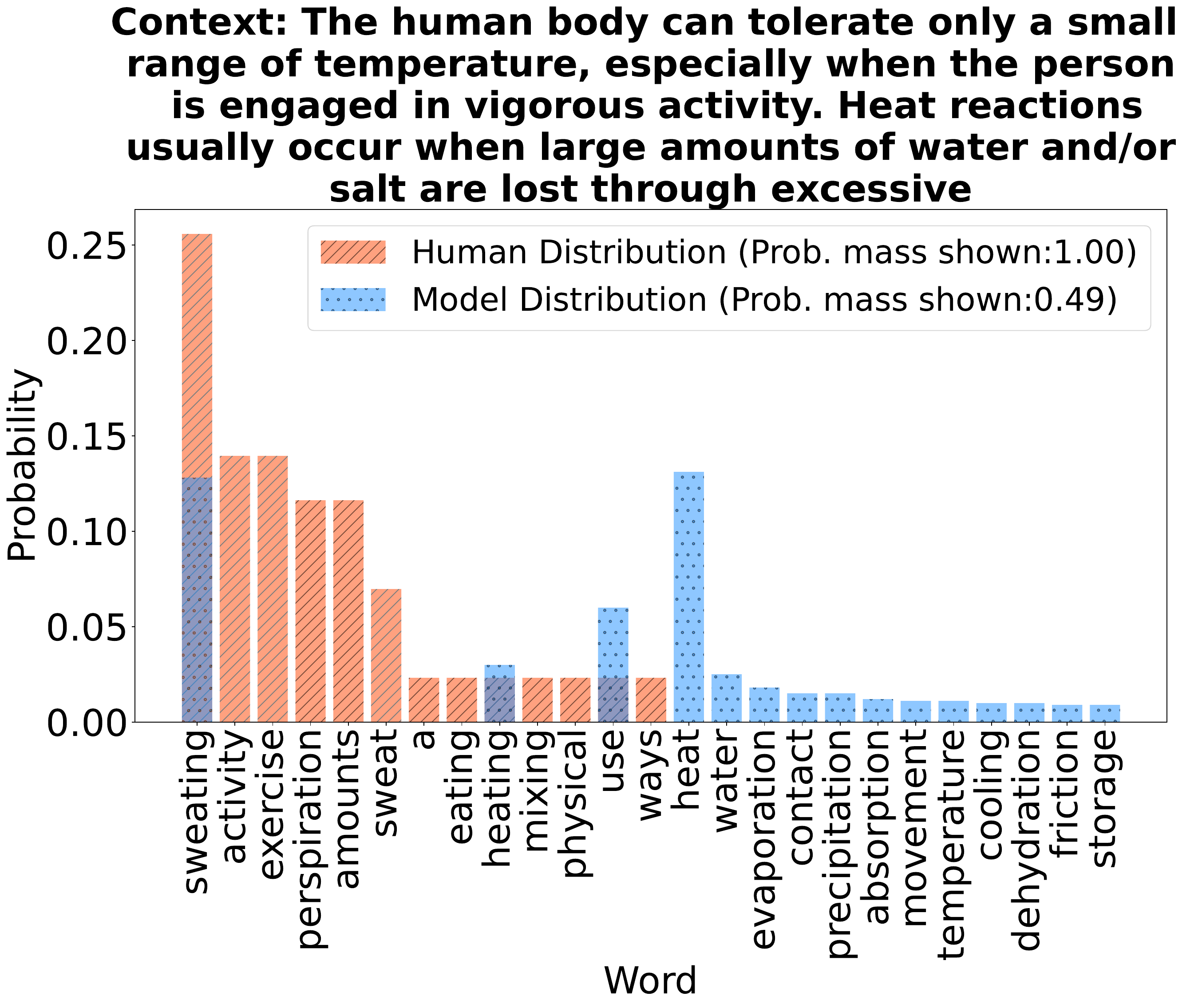}
\end{figure}

\begin{figure}
    \includegraphics[width=7.8 cm]{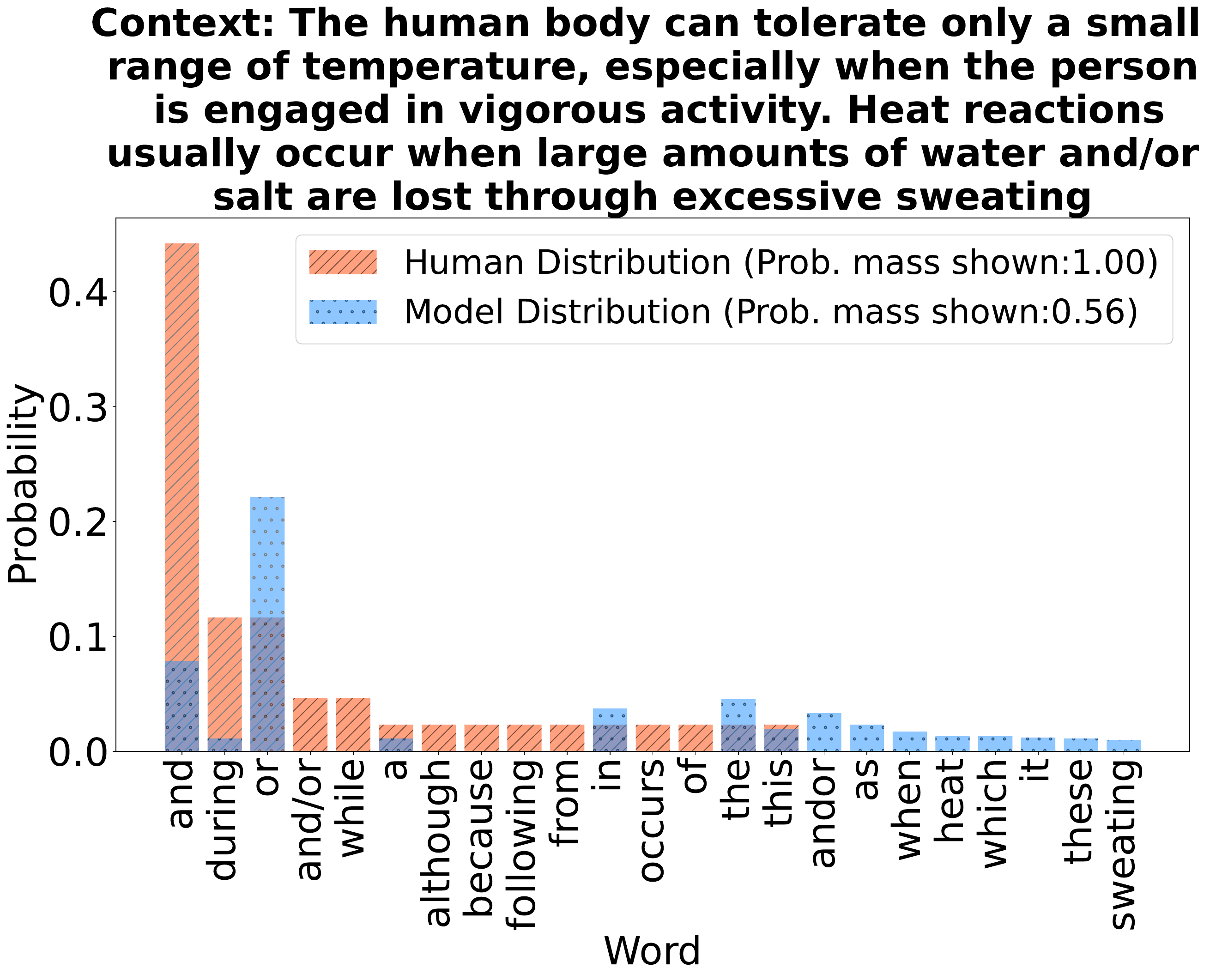}
    \includegraphics[width=7.8 cm]{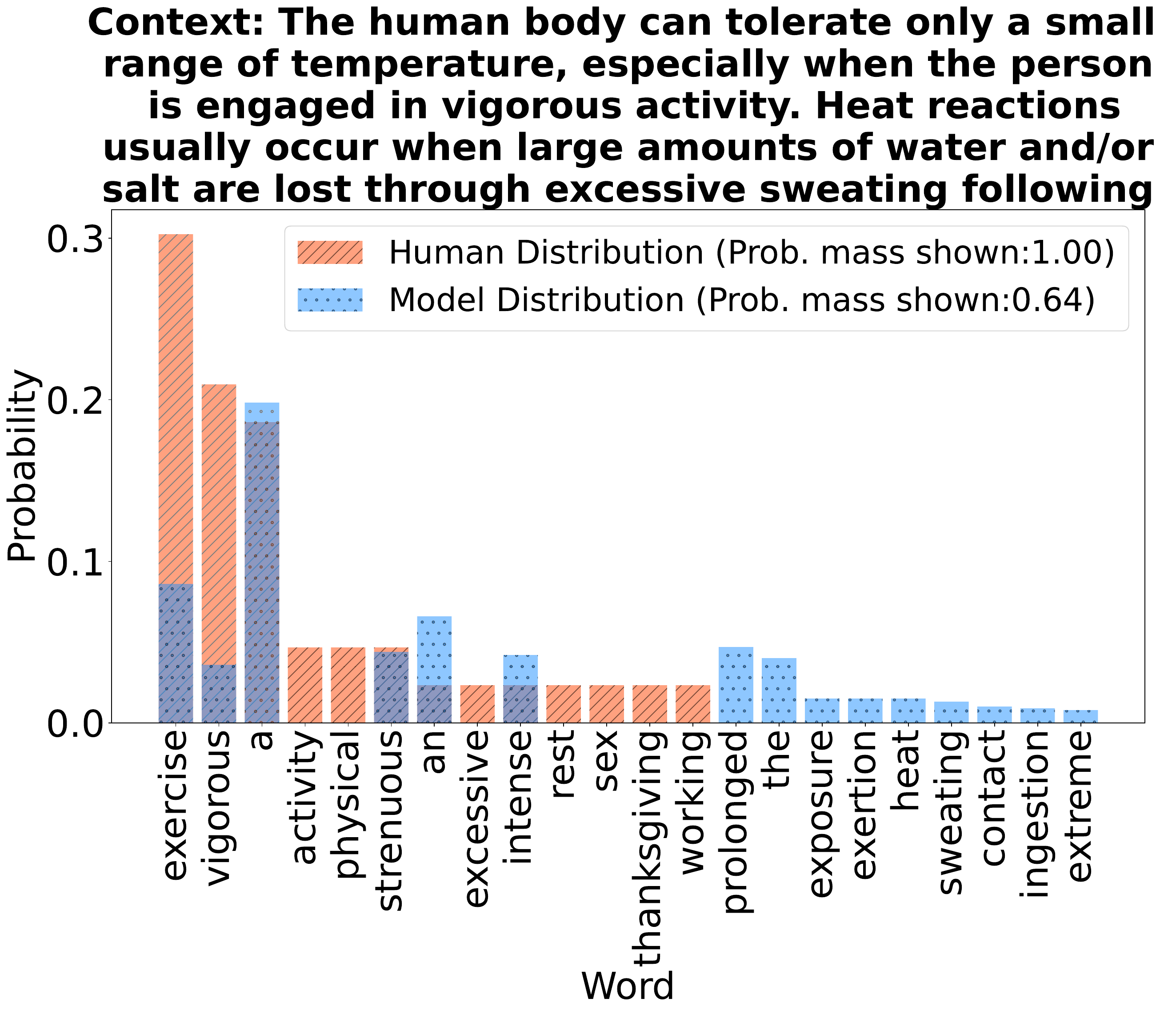} \includegraphics[width=7.8 cm]{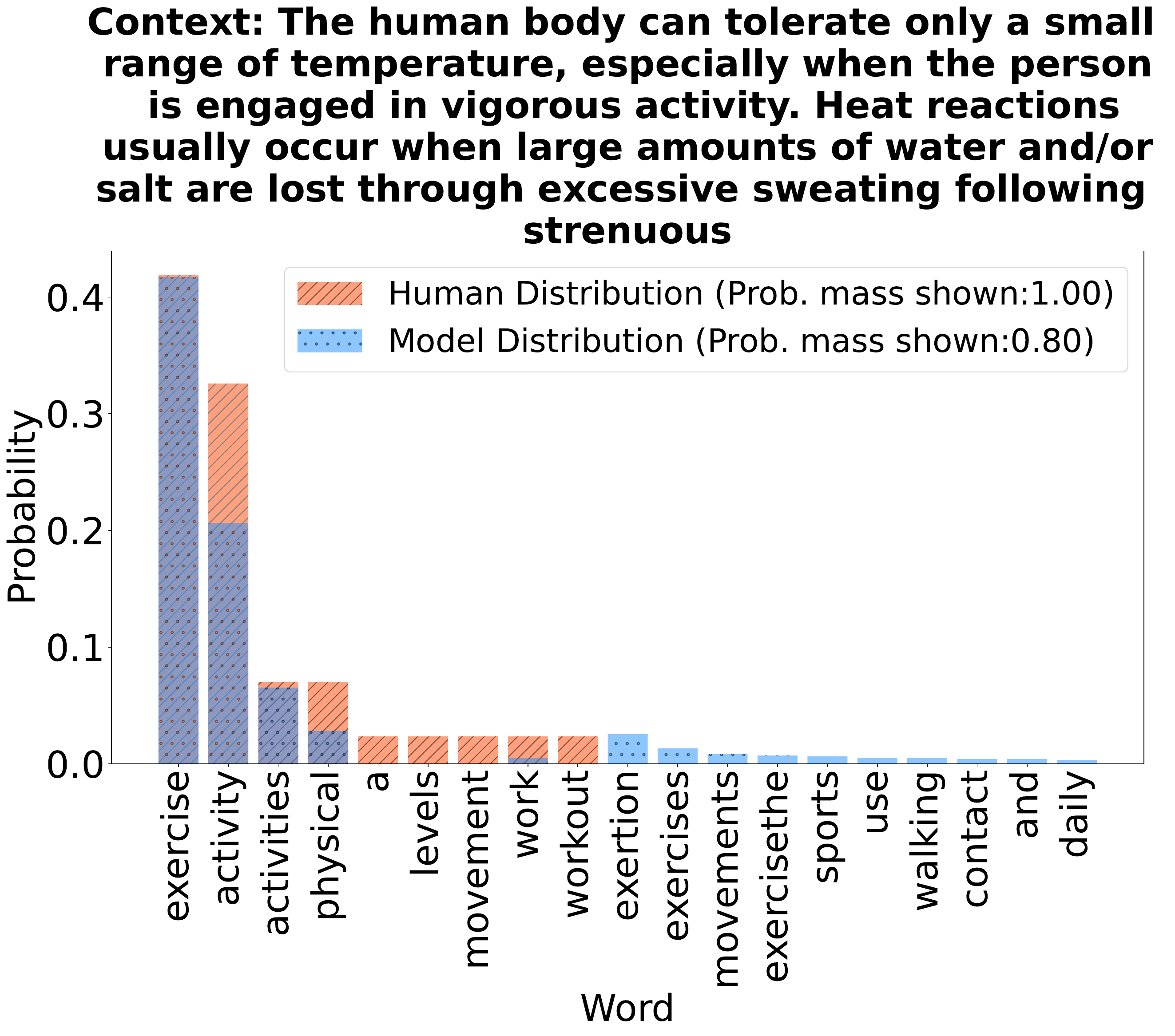}
\end{figure}

\begin{figure}
    \includegraphics[width=7.8 cm]{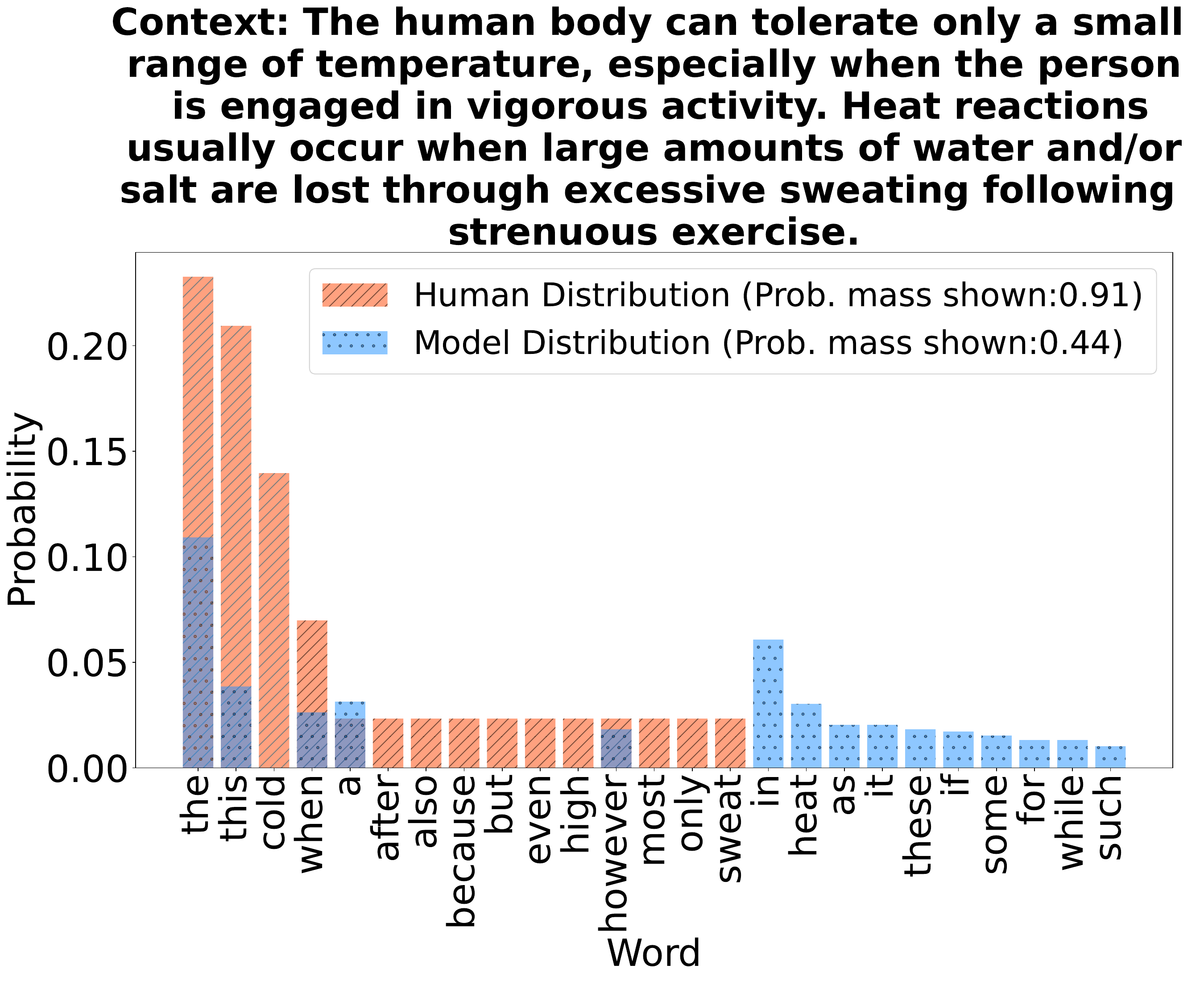}
    \includegraphics[width=7.8 cm]{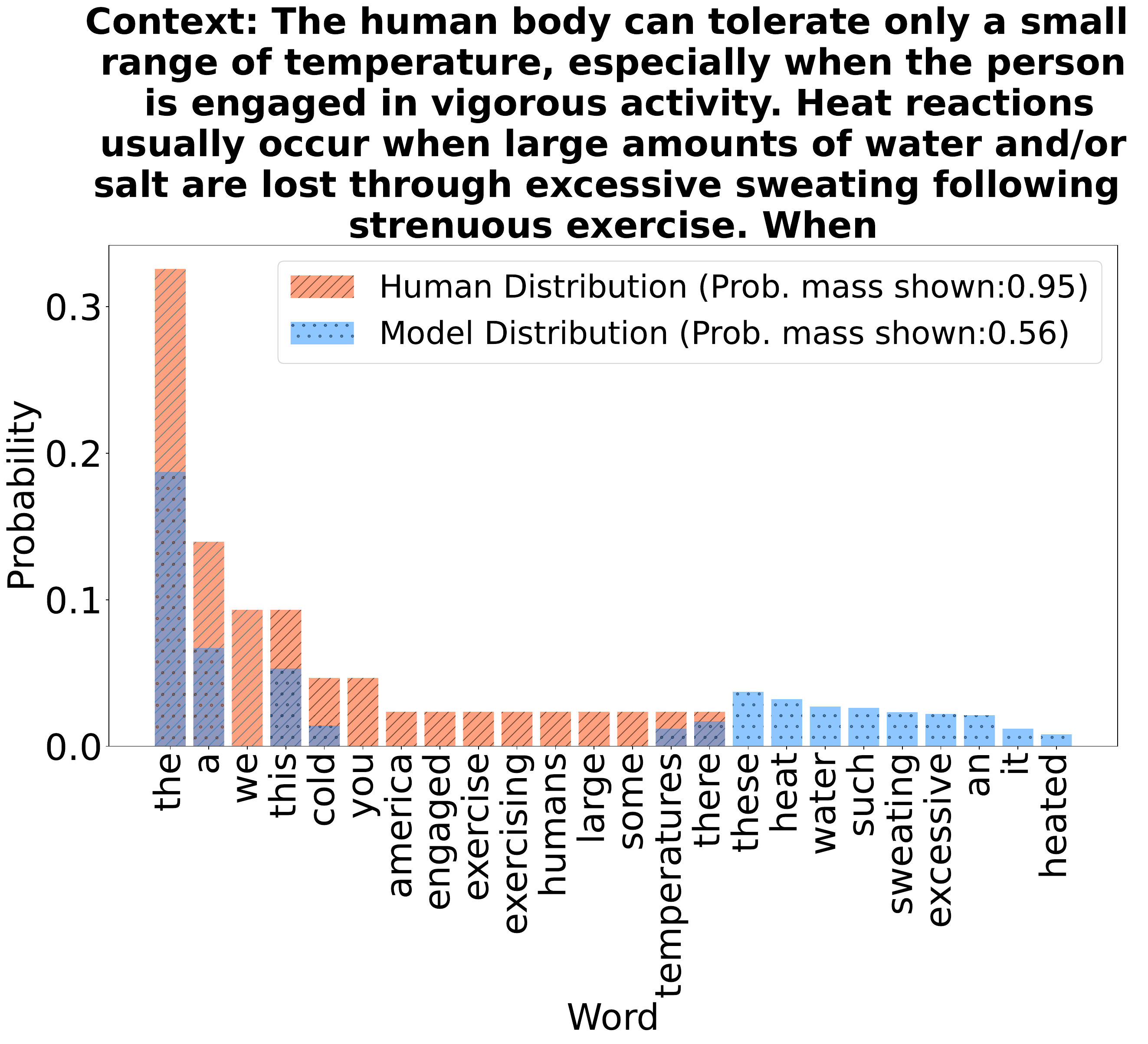}
    \includegraphics[width=7.8 cm]{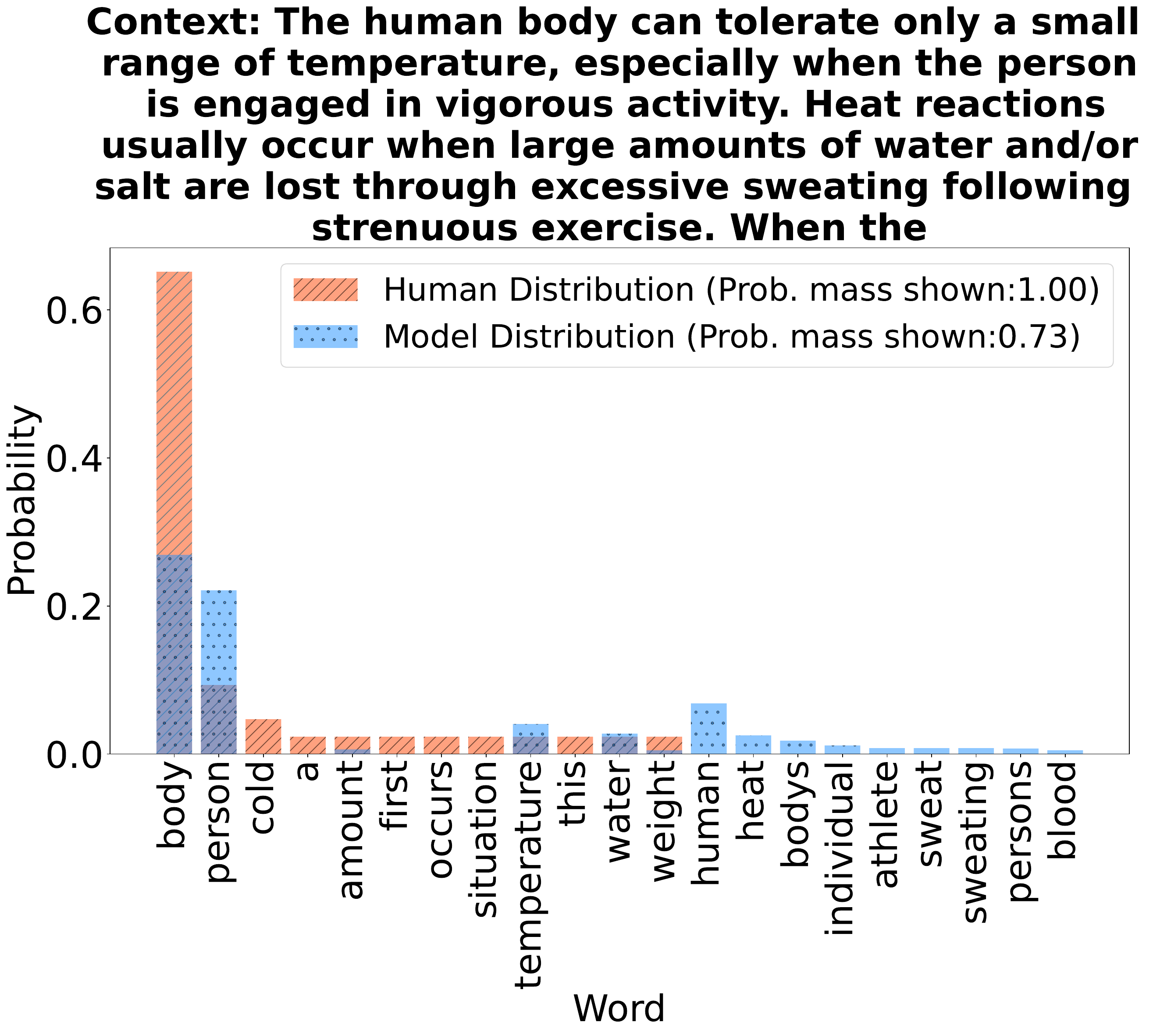} 
\end{figure}

\begin{figure}
    \includegraphics[width=7.8 cm]{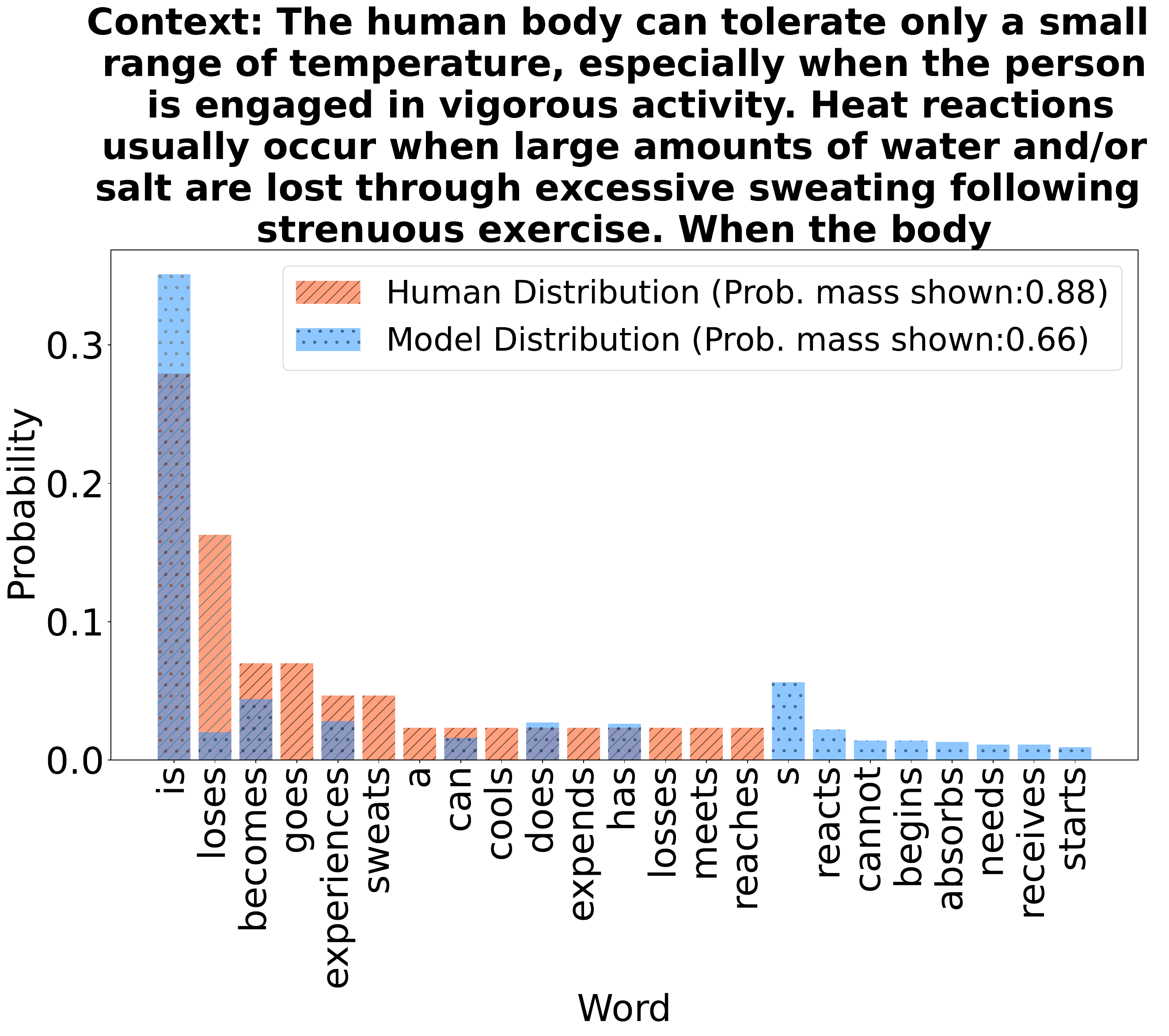}
    \includegraphics[width=7.8 cm]{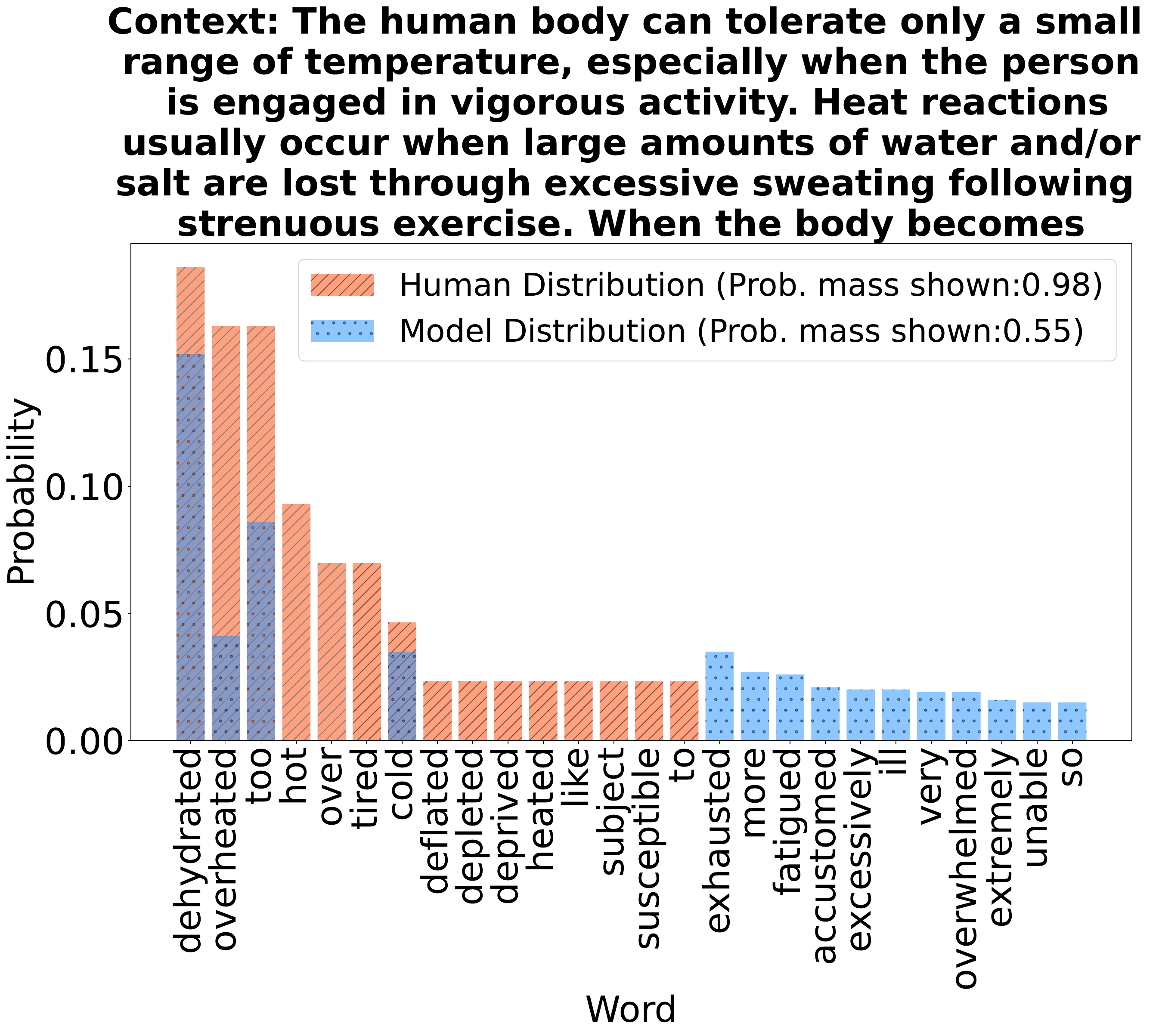}
    \includegraphics[width=7.8 cm]{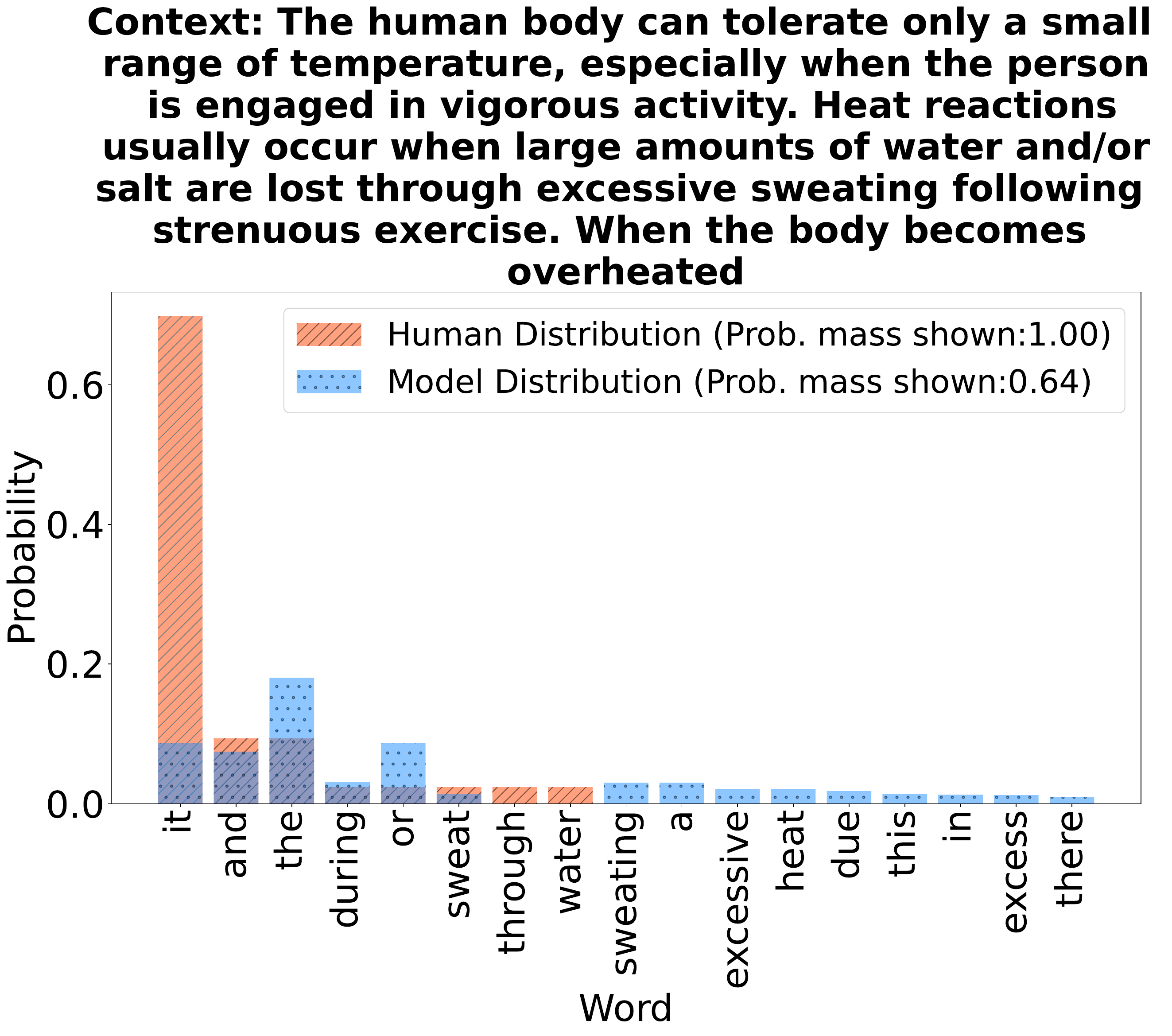}
\end{figure}

\begin{figure}
    \includegraphics[width=7.8 cm]{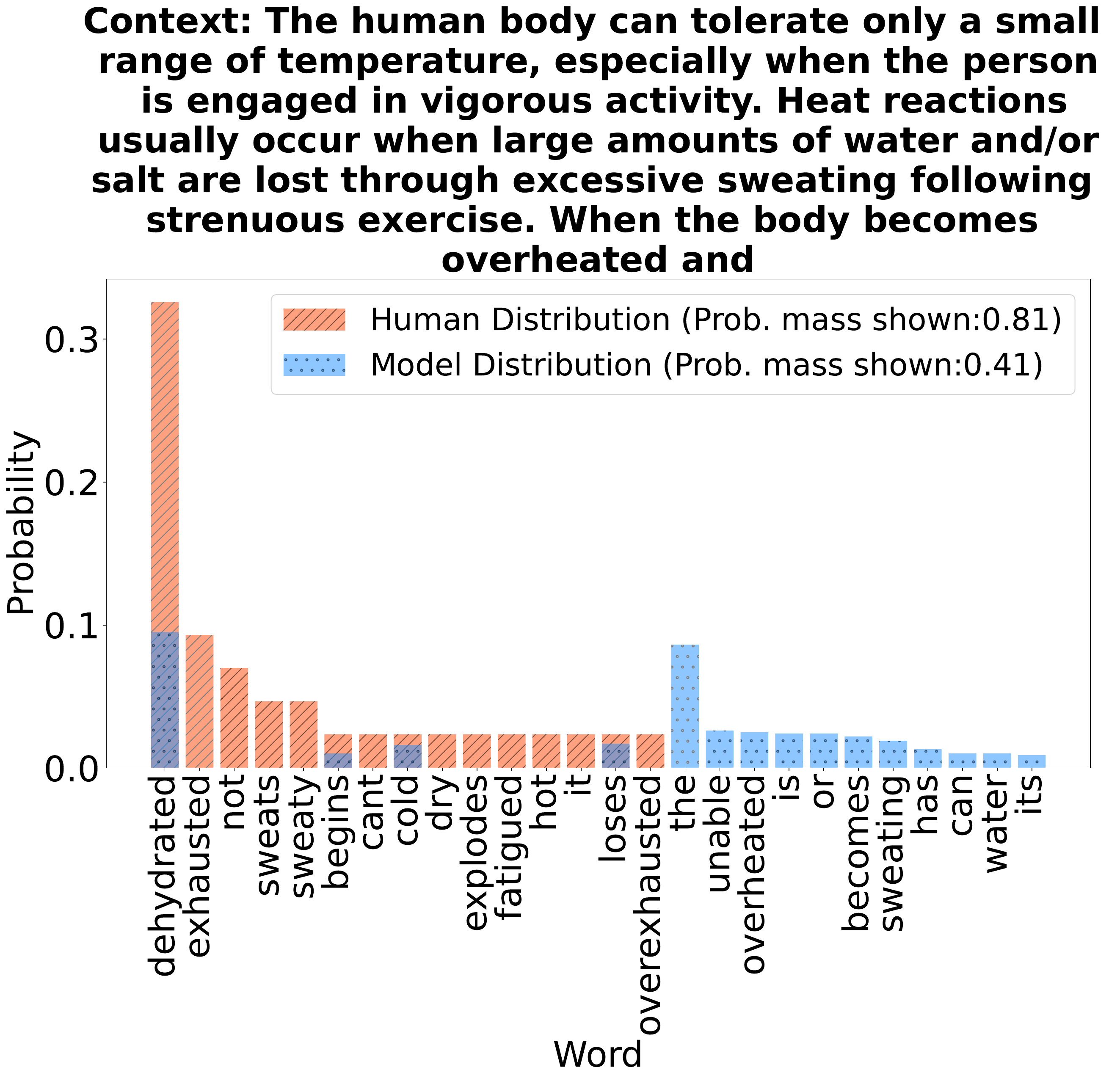} \includegraphics[width=7.8 cm]{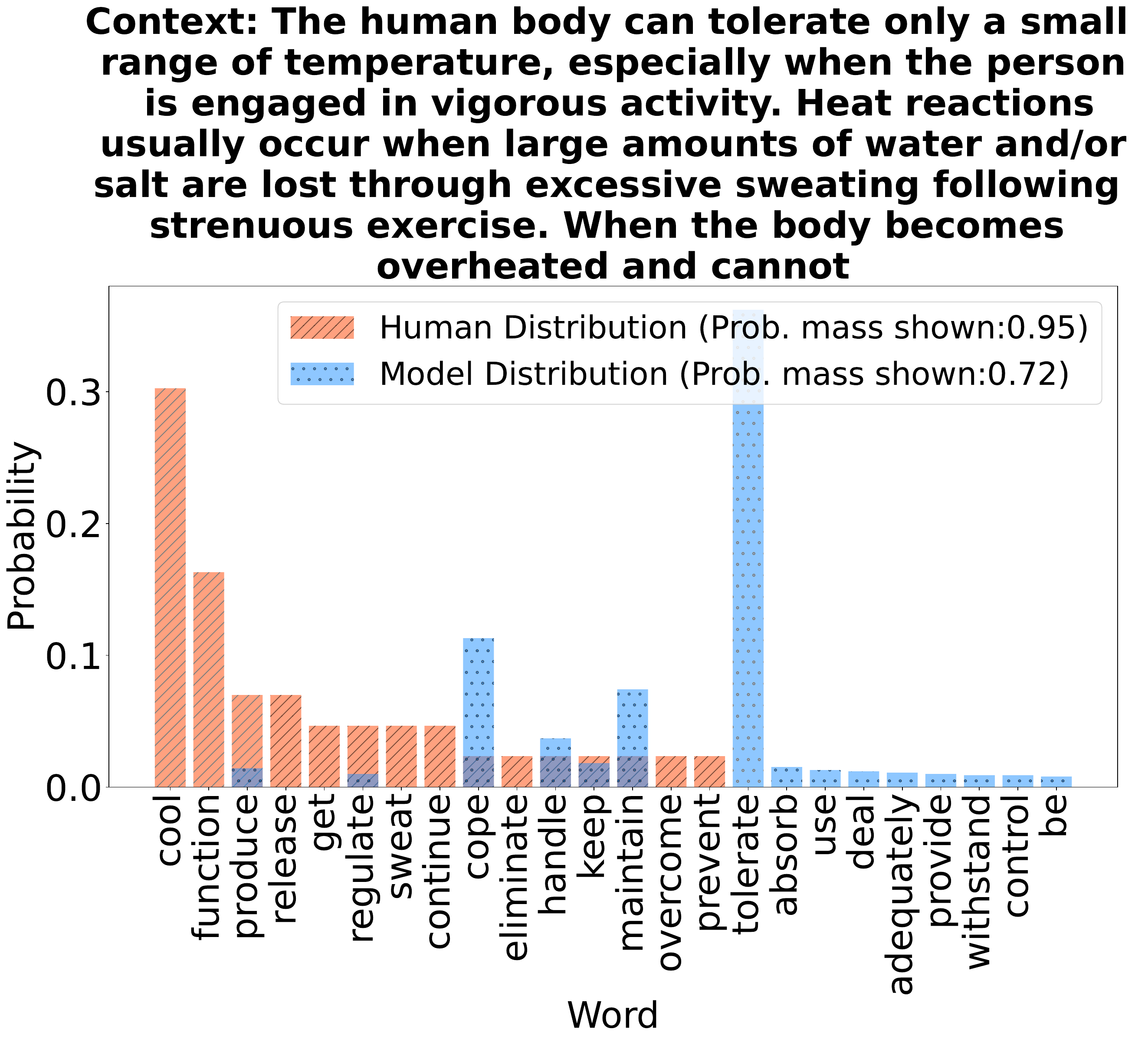}
    \includegraphics[width=7.8 cm]{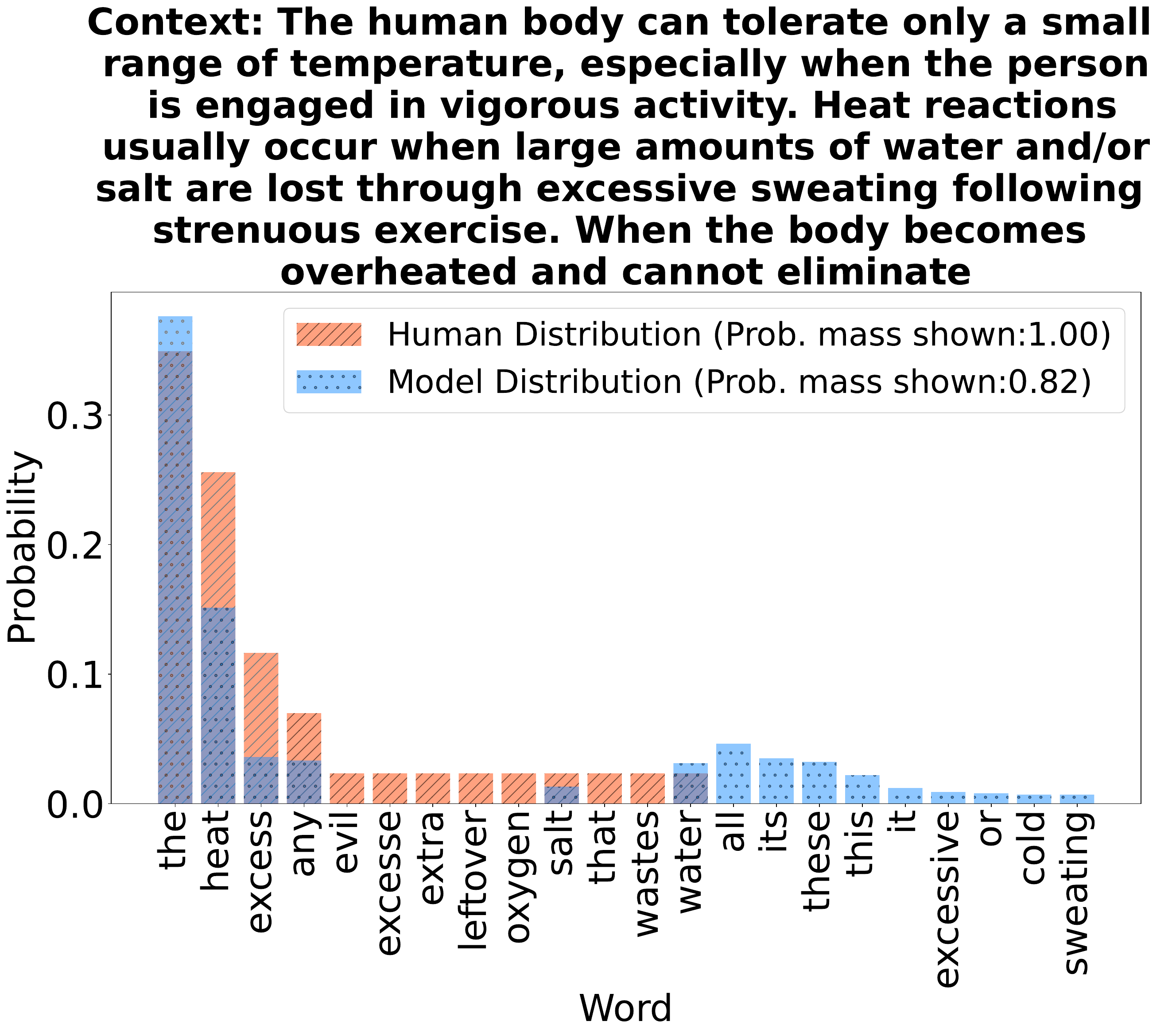}
\end{figure}
    
\begin{figure}    
    \includegraphics[width=7.8 cm]{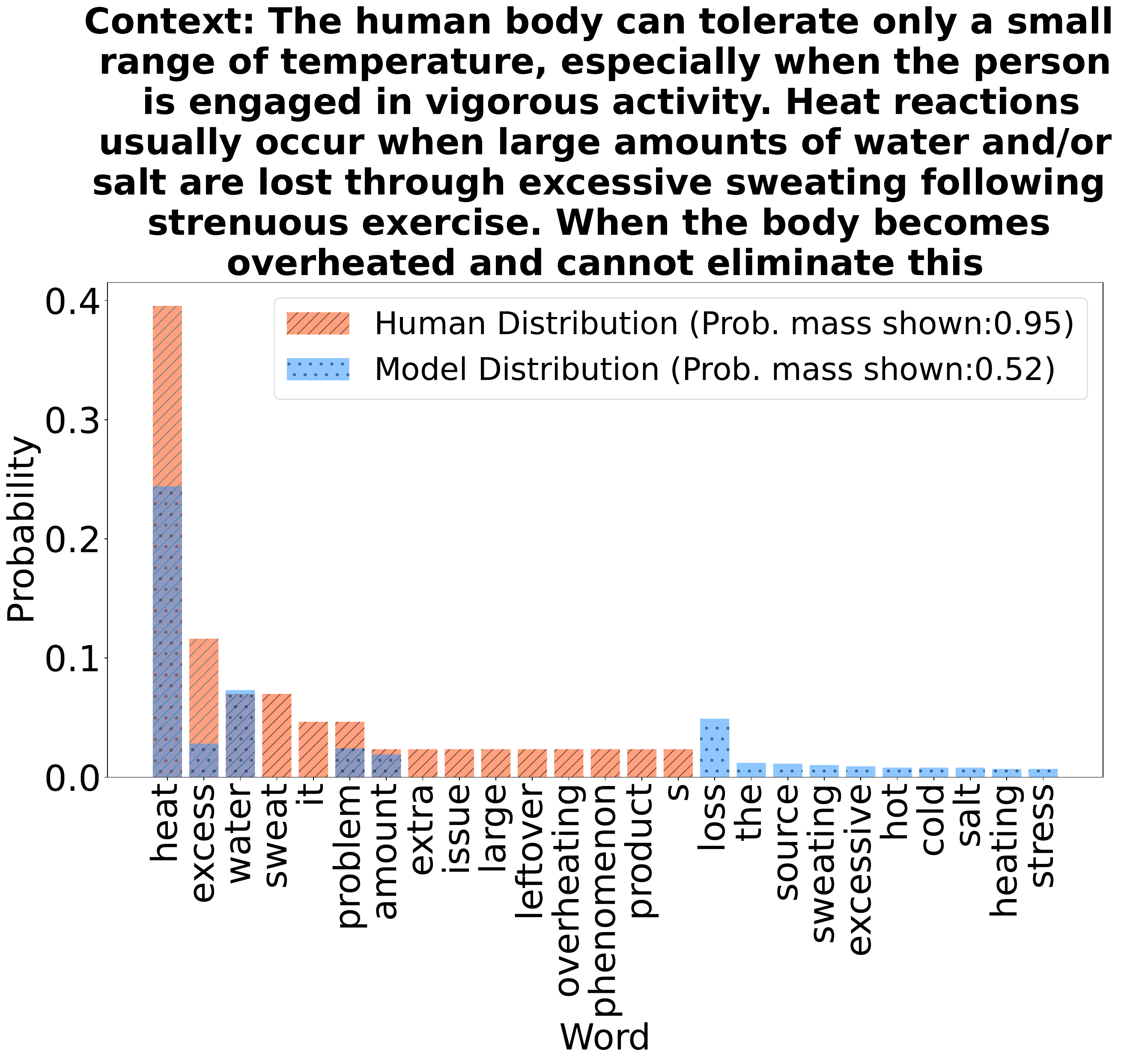}
    \includegraphics[width=7.8 cm]{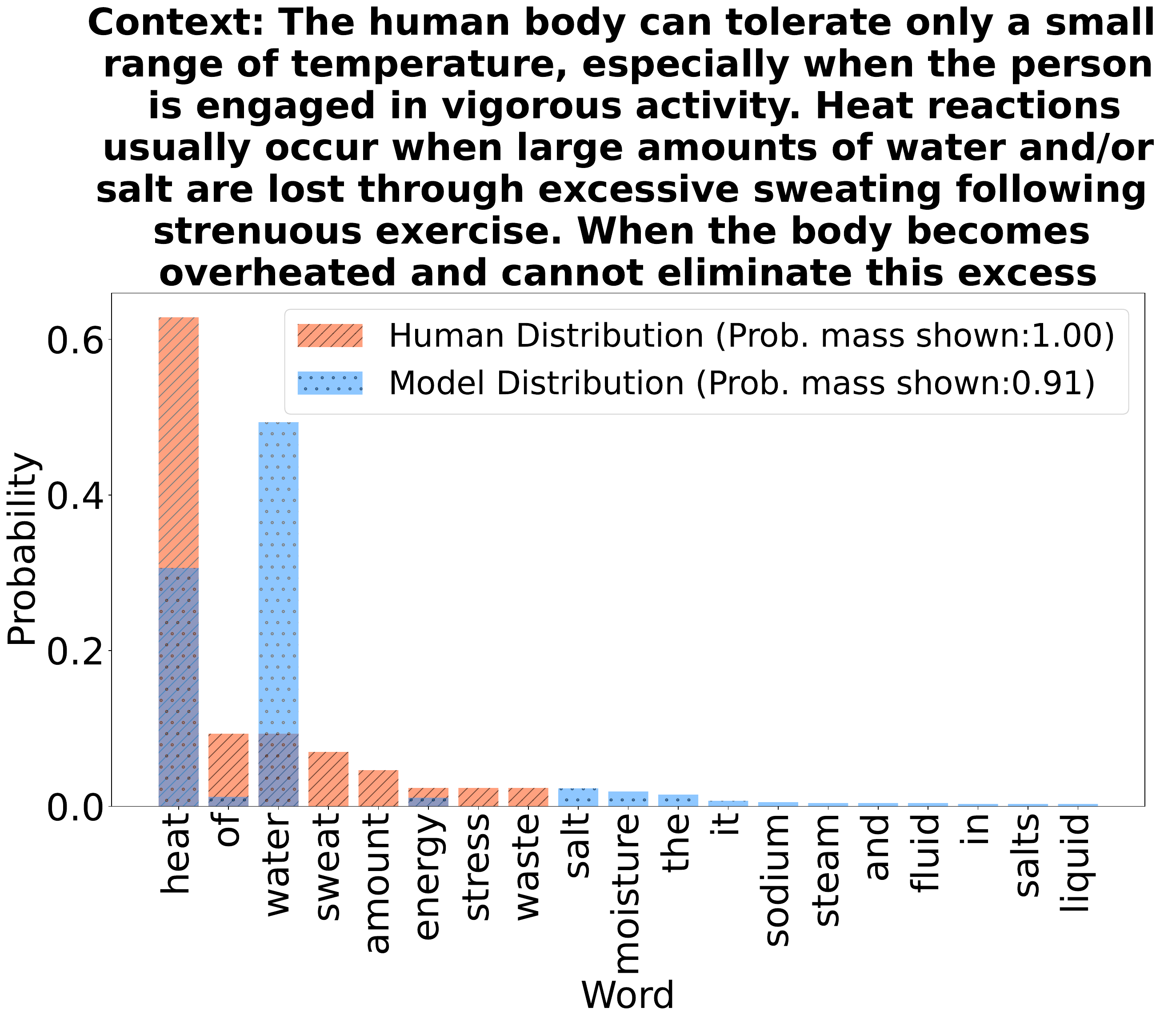} 
    \includegraphics[width=7.8 cm]{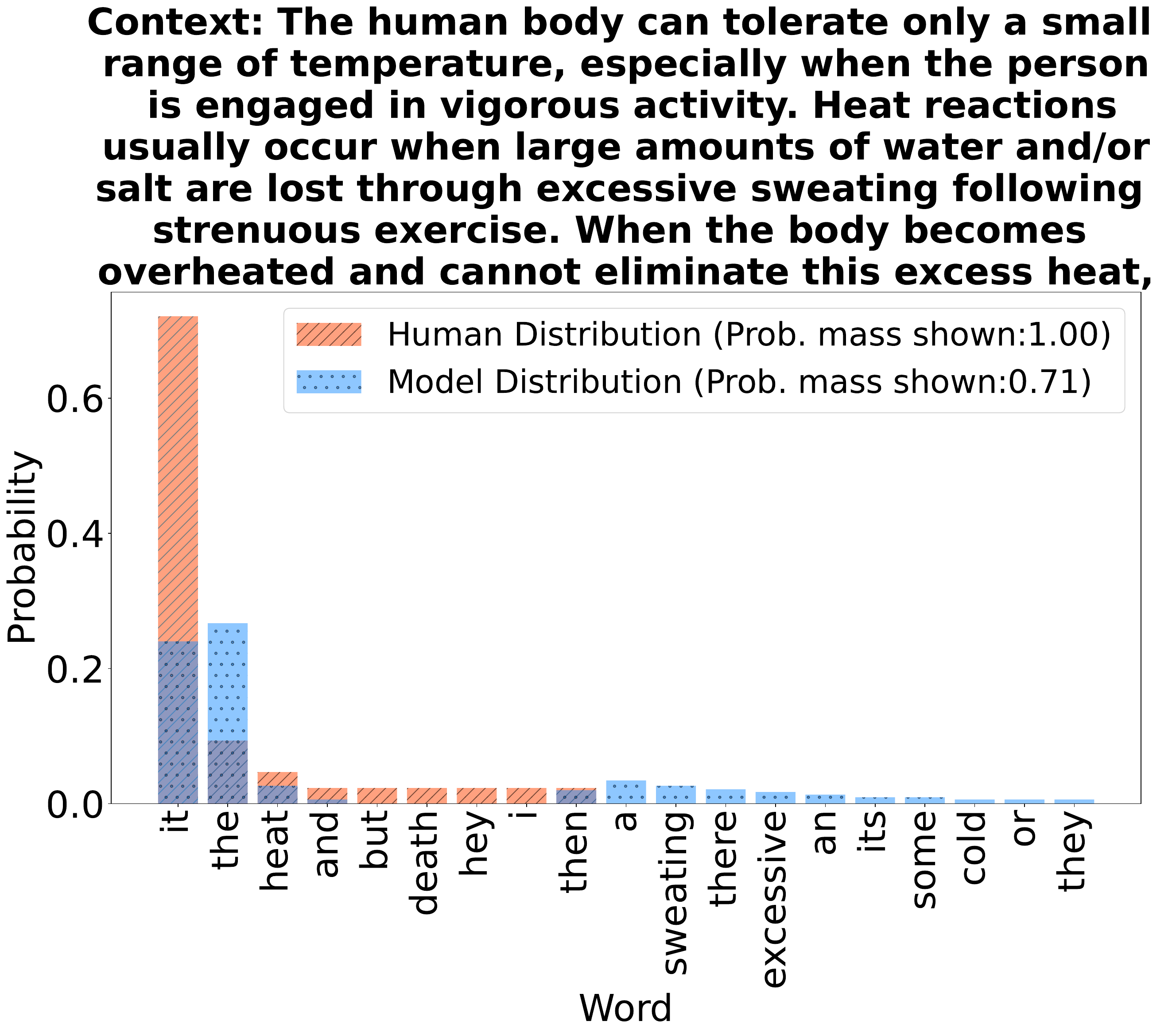}
\end{figure}

\begin{figure}
    \includegraphics[width=7.8 cm]{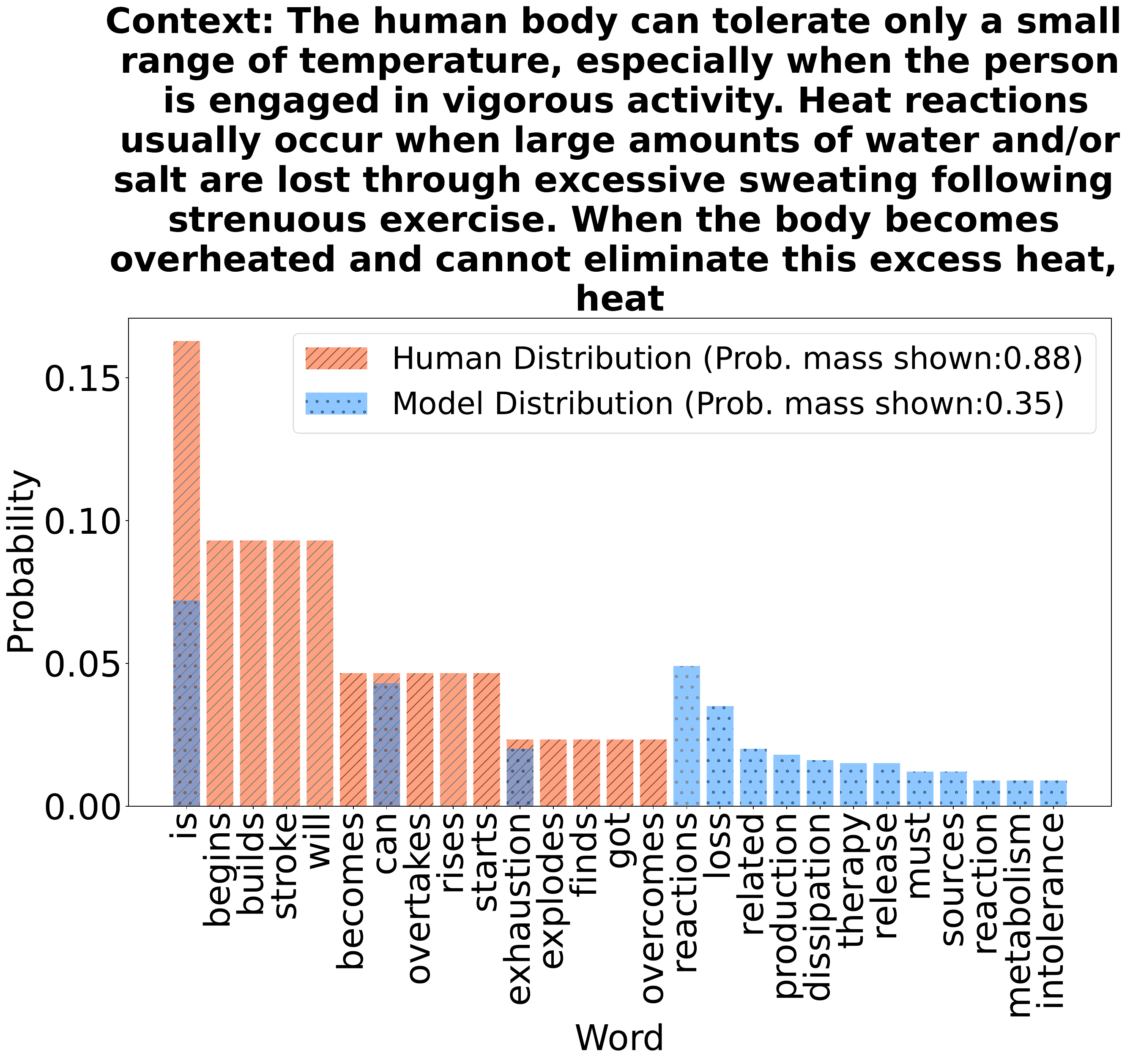}
    \includegraphics[width=7.8 cm]{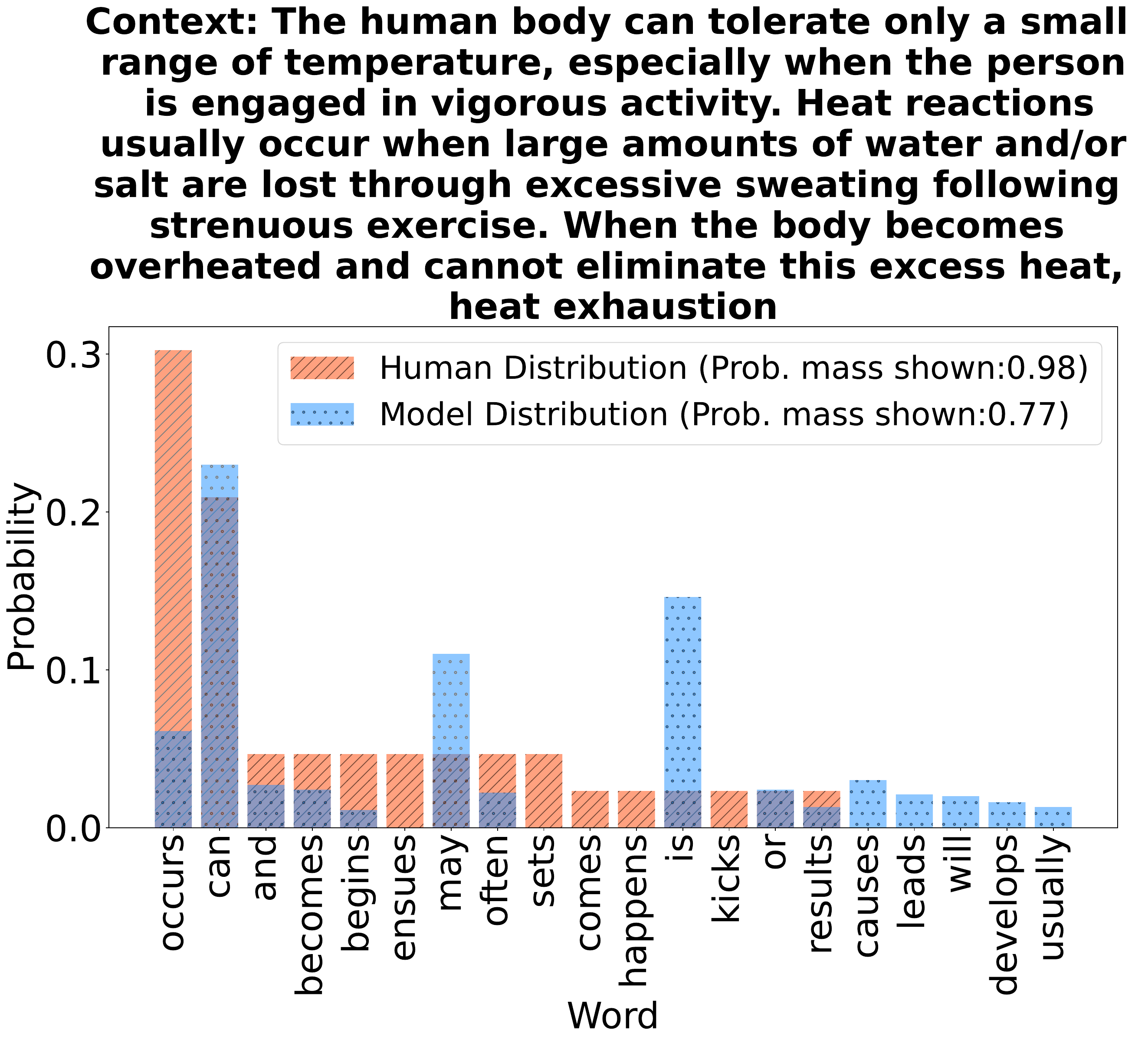}
    \includegraphics[width=7.8 cm]{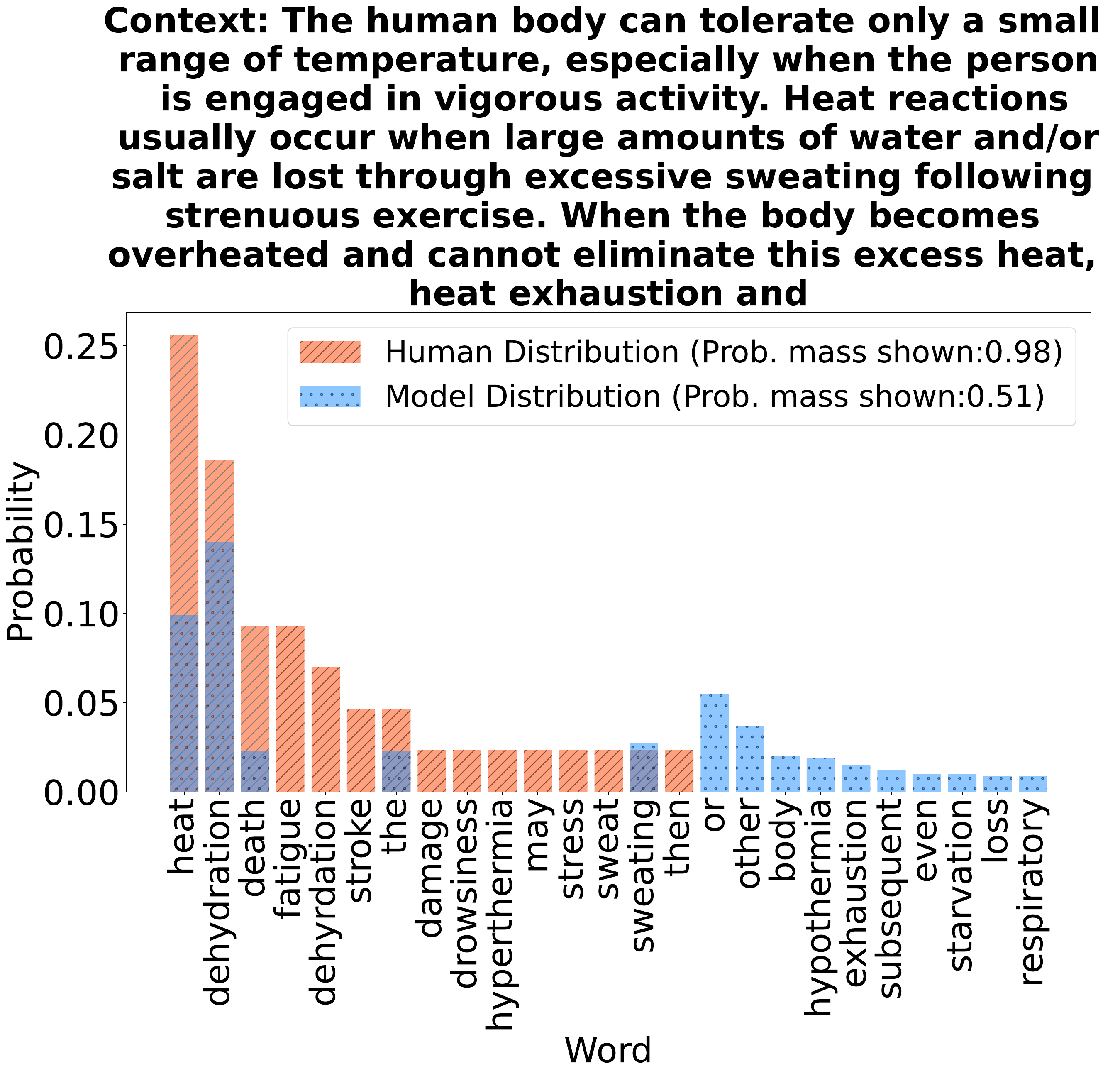} 
\end{figure}

\begin{figure}
    \includegraphics[width=7.8 cm]{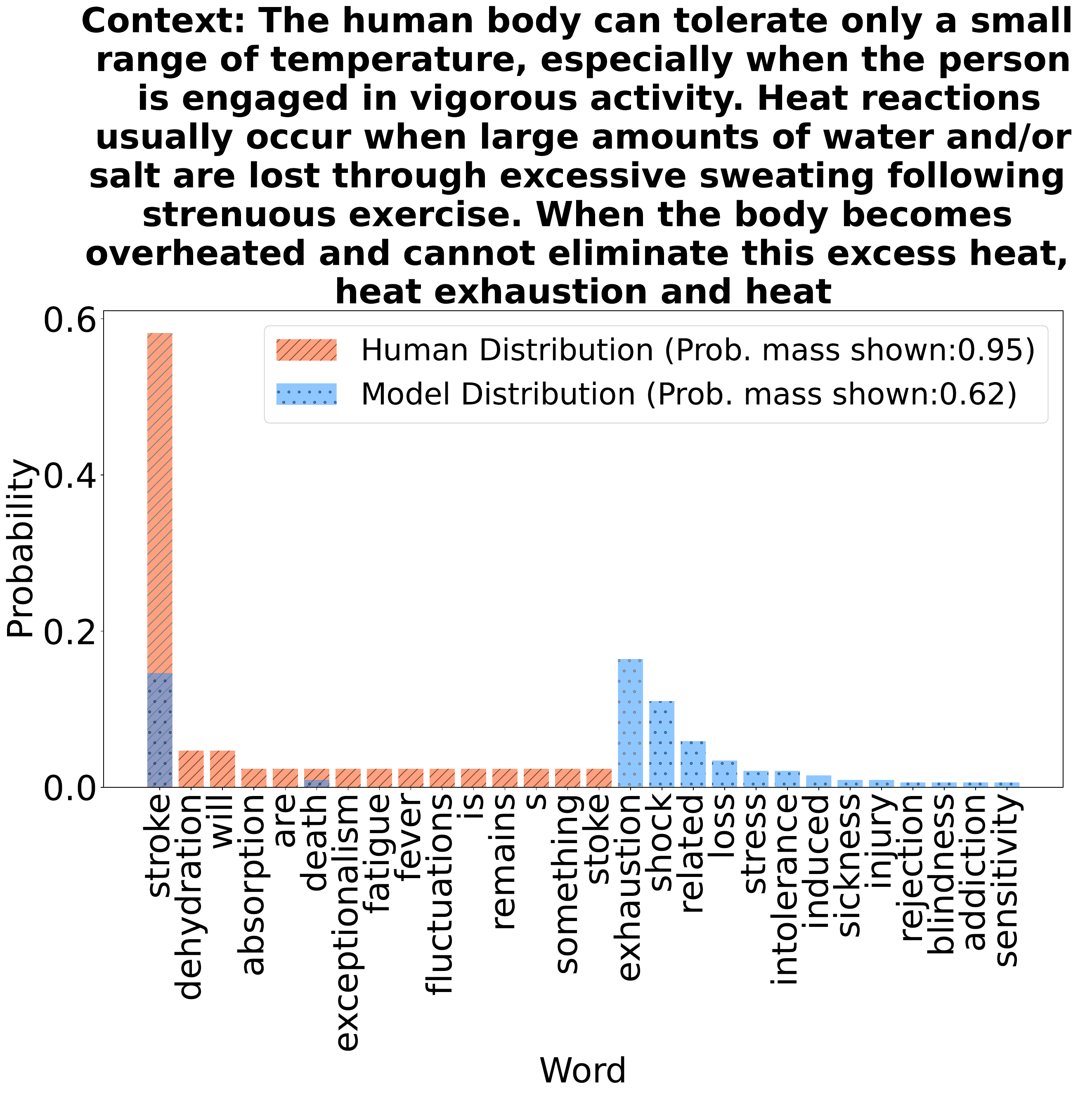}
    \includegraphics[width=7.8 cm]{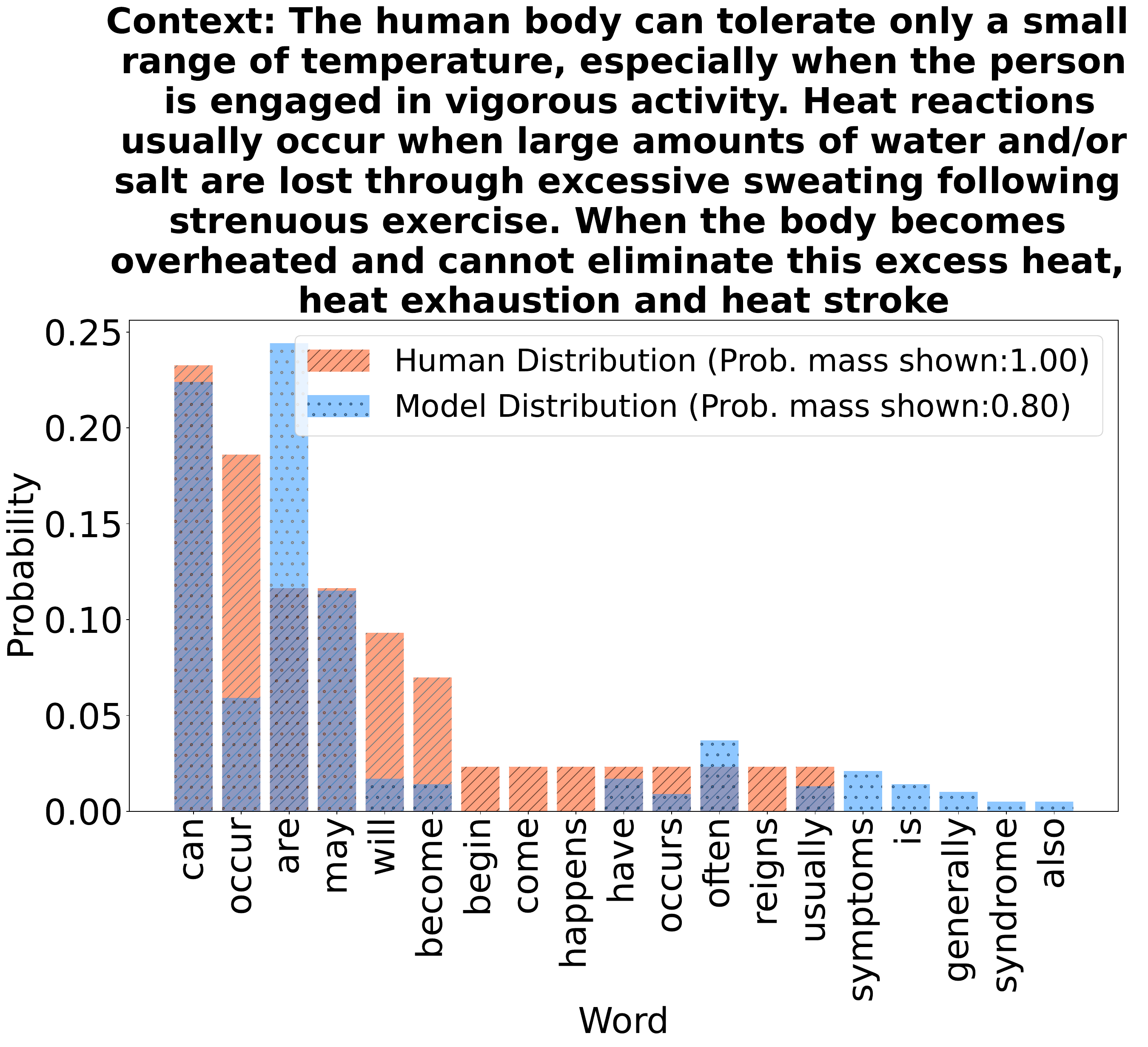}
    \includegraphics[width=7.8 cm]{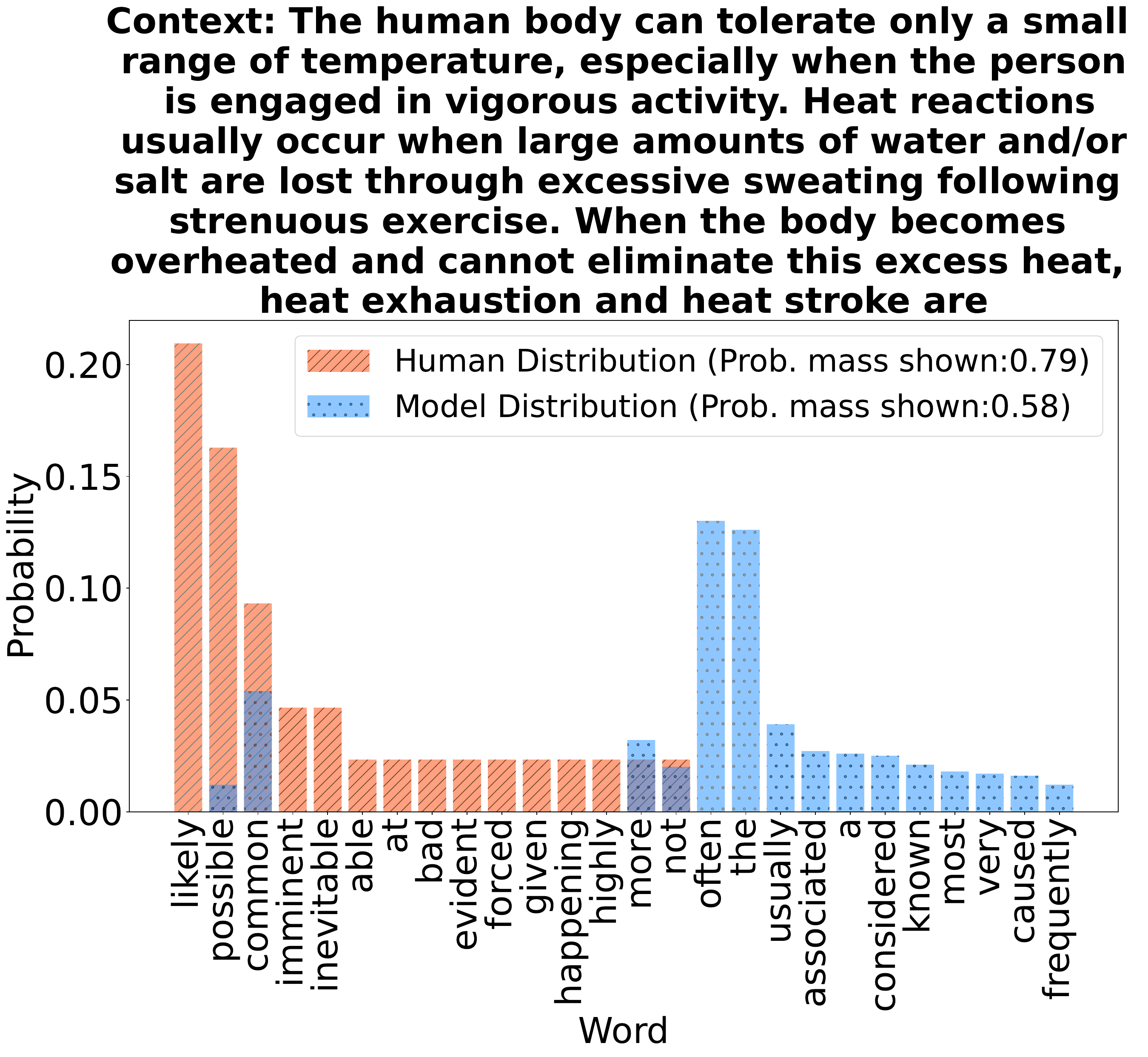}
\end{figure}

\end{document}